\pdfoutput=1

\documentclass[11pt]{article}

\usepackage{acl}

\usepackage{times}
\usepackage{latexsym}

\usepackage[T1]{fontenc}

\usepackage[utf8]{inputenc}

\usepackage{microtype}

\usepackage{inconsolata}

\usepackage{booktabs}
\usepackage{multirow}
\usepackage{adjustbox}
\usepackage{makecell}

\usepackage{subcaption}

\usepackage{graphicx}
\usepackage{dsfont}
\usepackage{amsfonts}
\usepackage{amsmath}
\usepackage{stackrel}
\DeclareMathOperator*{\argmax}{arg\,max}

\usepackage[english]{babel}

\newcommand{\BOS}{\textsc{BOS}}
\newcommand{\EOS}{\textsc{EOS}}
\newcommand{\vx}{\mathbf{x}}
\newcommand{\vy}{\mathbf{y}}
\newcommand{\vh}{\mathbf{h}}

\newcommand{\Pmodel}{P_\mathrm{{model}}} 
\newcommand{\Pm}{P}

\newcommand{\CandH}{\mathcal{H}_{\mathrm{cand}}}
\newcommand{\RefH}{\mathcal{H}_{\mathrm{ref}}}

\usepackage{xcolor}

\title{Generating Diverse and High-Quality Texts by \\ Minimum Bayes Risk Decoding}

\author{Yuu Jinnai, Ukyo Honda, Tetsuro Morimura, Peinan Zhang \\
  CyberAgent \\
  \texttt{\{jinnai\_yu,honda\_ukyo,morimura\_tetsuro,zhang\_peinan\}@cyberagent.co.jp}
}

\begin{document}
\maketitle
\begin{abstract}
One of the most important challenges in text generation systems is to produce outputs that are not only correct but also diverse.
Recently, Minimum Bayes-Risk (MBR) decoding has gained prominence for generating sentences of the highest quality among the decoding algorithms. 
However, existing algorithms proposed to generate diverse outputs are predominantly based on beam search or random sampling, thus their output quality is capped by these underlying decoding algorithms. 
In this paper, we investigate an alternative approach -- we develop diversity-promoting decoding algorithms by enforcing diversity objectives to MBR decoding. 
We propose two variants of MBR; (i) Diverse MBR (DMBR) that adds a diversity penalty to the decoding objective and (ii) $k$-medoids MBR (KMBR) that reformulates the decoding task as a clustering problem.
We evaluate DMBR and KMBR on a variety of directed text generation tasks using encoder-decoder models and a language model with prompting. The experimental results show that the proposed methods achieve a better trade-off than the diverse beam search and sampling algorithms overall.
Our code is available at \url{https://github.com/CyberAgentAILab/diverse-mbr/}.
\end{abstract}

\section{Introduction}

There are many reasons why natural language generation systems want to produce outputs that are not only correct but also diverse.
For example, in systems involving reranking candidate outputs, the reranking algorithms are more effective when the candidates are diverse \cite{gimpel-etal-2013-systematic,li2016mutual,li-etal-2016-diversity,choudhary2017domain}.
In image captioning, an image may contain many concepts with multiple levels of detail. To achieve human-level image captioning, it is important for models to be able to output a variety of captions covering such diverse information \cite{WangC19}. As an application, \citet{KrauseJKF17} shows that a diverse set of image captions can be transformed into an entire descriptive paragraph explaining the image.
For question generation tasks, a diversity-promoting question generation system can improve question answering systems \cite{sultan-etal-2020-importance} and enhance the engagement in chatbots \cite{laban-etal-2020-whats}.





This importance of diversity in text generation has brought many studies aimed at producing diverse outputs instead of the most probable ones. 
The majority of the approaches are based on either random sampling \cite{fan-etal-2018-hierarchical,ippolito-etal-2019-comparison,Holtzman2020The,hewitt-etal-2022-truncation} or beam search \cite{cho2016noisy,li2016mutual,Vijayakumar2016,vijayakumar2018diverse,kulikov-etal-2019-importance,tam2020cluster}, thus the output quality is bounded by the quality of either random sampling or beam search.


In this paper, we develop diverse decoding methods by extending Minimum Bayes Risk (MBR) decoding \cite{goel2000minimum,kumar-byrne-2002-minimum,kumar-byrne-2004-minimum,eikema-aziz-2022-sampling}.
MBR decoding is shown to generate higher quality sentences than random sampling and beam search in directed text generation tasks including machine translation, text summarization, and data-to-text \cite{freitag-etal-2022-high,suzgun-etal-2023-follow}. The procedure of the MBR decoding consists of the following steps. First, it samples a set of candidate outputs from the probability model. 
Then, it computes the similarity of each sequence to the others according to a utility function.
Finally, it selects the sequence that maximizes the expected utility over the sequences.
A naive approach to generate $k$ outputs using MBR is to select the top-$k$ outputs with the highest expected utility. However, it tends to select a set of similar sentences with a large overlap (see Appendix~\ref{sec:sample} for examples).

We propose two approaches to promote diversity in MBR: Diverse MBR (DMBR) and $k$-Medoids MBR (KMBR).
DMBR extends MBR by introducing a diversity penalty to the objective so that it maximizes the weighted sum of the expected utility and diversity. DMBR can tune the quality-diversity trade-off by the weight hyperparameter. 
KMBR selects a set of sentences by solving the $k$-medoids problem \cite{rdusseeun1987clustering,kaufman2009finding}. $k$-medoids is a variant of $k$-means where the center points are restricted to be one of the data points instead of anywhere in the space. We pick a center point from each cluster to generate diverse and high quality outputs. 

We evaluate DMBR and KMBR on machine translation, image captioning, question generation, generative common sense reasoning, and text summarization.
The experimental results show that DMBR and KMBR achieve better trade-offs than diverse beam search and sampling algorithms.
We also observe that DMBR and KMBR achieve higher Oracle quality scores in all the tasks.

\section{Background}
Sequence-to-sequence generation is the task of generating an output sequence $\vy$ given an input sequence $\vx$.
Probabilistic text generators define a probability distribution $p_\theta (\mathbf{y} | \mathbf{x})$ over an output space of hypotheses $\mathcal{Y}$ conditioned on an input $\mathbf{x}$.
The set of complete hypotheses $\mathcal{Y}$ is:
\begin{equation}
    \mathcal{Y} := \{\BOS \circ \mathbf{v} \circ \EOS | \mathbf{v} \in \mathcal{V}^*\},
\end{equation}
where $\circ$ is a string concatenation and $\mathcal{V}^*$ is the Kleene closure of a set of vocabulary $\mathcal{V}$. 
Typically, the goal of decoding is to find the hypothesis that best matches the human preference $P_{\mathrm{human}}$:
\begin{equation}
    \vh^* = \argmax_{\vh \in \mathcal{Y}} P_{\mathrm{human}}(\vh | \vx).
\label{eq:set}
\end{equation}
In this paper, we consider a variant of the set decoding problem \cite{meister-etal-2021-determinantal} where the task is to generate a set of $k$ sentences $H$ that maximizes the sum of $P_{\mathrm{human}}$ and the diversity according to the preference of human $d_{\mathrm{human}}$: 
\begin{equation}
    H^* = \argmax_{H \subseteq \mathcal{Y}} \sum_{\vh \in H} P_{\mathrm{human}}(\vh | \vx) + d_{\mathrm{human}}(H).
\label{eq:setdiv}
\end{equation}
In this paper, we use automated evaluation metrics to approximate $d_{\mathrm{human}}$. In particular, we consider P-BLEU, distinct-n, and P-SentBERT as measures of diversity \cite{shen2019mixture,li-etal-2016-diversity,reimers-gurevych-2019-sentence}. 

\subsection{Decoding Algorithms for Diversity}
There have been two major approaches to diversity-aware text decoding: random sampling and diversity-aware beam search.

\paragraph{Random sampling.} Random sampling is commonly used to generate diverse outputs in both directed and open-ended text generation tasks. 
A simple solution is to use an ancestral sampling with a temperature parameter to control the stochasticity of the sampling. There have been a lot of studies on biasing the ancestral sampling to generate higher quality outputs while maintaining the diversity of the randomized algorithm.
Prior work shows that top-$k$ sampling that restricts the sampling to the $k_{\mathrm{top}}$ most likely tokens at each step is a better alternative to controlling the temperature in a story generation task \cite{fan-etal-2018-hierarchical,ippolito-etal-2019-comparison}.
Nucleus sampling \cite{Holtzman2020The} is similar to top-$k$ sampling and has shown to be more effective than top-$k$ sampling in WebText dataset \cite{radford2019language}. Nucleus sampling truncates all tokens except those in the {\it nucleus}, the smallest possible set of tokens that covers a fraction $p$ of the model probability.
Epsilon sampling \cite{hewitt-etal-2022-truncation} is also a variant of ancestral sampling that truncates tokens whose probability is less than a threshold. They are shown to be more effective than nucleus sampling in the WebText dataset.
However, these random samplings improve the diversity of outputs at the expense of the quality of outputs \citep{ippolito-etal-2019-comparison}.

\paragraph{Diversity-aware beam search.}
Another line of work is to generate a diverse set of outputs by introducing diversity objectives to the beam search. 
Beam search is known to produce higher quality sequences than random sampling in a wide range of tasks \cite{graves2012sequence,Sutskever2014}. 
Diversity has been induced in the beam search procedure in various forms. 
\citet{li2016mutual} propose to add a diversity constraint to standard beam search so that the number of descendants from the same parent hypothesis is capped at some upper bound.
Noisy Parallel Approximate Decoding induces noise to the hidden state of the decoder to generate randomized outputs \cite{cho2016noisy}. 
Diverse beam search (DBS) adds a diversity term to the reranking objective, penalizing sequences with a small Hamming distance to other groups of the sequences \cite{Vijayakumar2016,vijayakumar2018diverse}.
Iterative beam search runs beam search multiple times with a constraint that any partial hypothesis that has been generated in previous iterations is prohibited \cite{kulikov-etal-2019-importance}.
Clustered beam search prunes similar sequences by clustering the candidates using a word embedding at each decoding step \cite{tam2020cluster}.
Post-Decoding Clustering (PDC) encourages diversity by running the clustering after generating a large number of outputs, selecting good candidates from each cluster \cite{kriz-etal-2019-complexity,ippolito-etal-2019-comparison}.
Best-$k$ search maintains best-first queue in a course of beam search, resulting in higher quality and diversity outputs \cite{xu-etal-2023-best}.

\subsection{Minimum Bayes Risk (MBR) Decoding}

One of the most common decision rules to solve the decoding problem is maximum-a-posteriori (MAP) decoding such as beam search. MAP decoding finds the most probable translation under the model.
\begin{equation}
    \vh^{\mathrm{MAP}} = \argmax_{\vh \in \mathcal{Y}} P(\vh | \vx).
\end{equation}
In this paper, we denote $P(\vh | \vx)$ as $P(\vh)$ for simplicity.
Although it seems intuitive to compute this MAP objective, previous work has pointed out two critical problems with this strategy. First, because the size of the hypothesis set $|\mathcal{Y}|$ is extremely large, it is intractable to solve optimally. Second, the MAP objective often leads to low quality output \cite{stahlberg-byrne-2019-nmt,Holtzman2020The,meister-etal-2020-beam}. Indeed, \citet{stahlberg-byrne-2019-nmt} show that $\vh^{\mathrm{MAP}}$ is often found to be the empty sequence in their experimental setting.

Unlike MAP decoding, which searches for the most probable output, MBR decoding searches for the output that maximizes expected utility, which is equivalent to minimizing risk \cite{goel2000minimum,kumar-byrne-2002-minimum,kumar-byrne-2004-minimum}.
The procedure consists of two components: a text generation model and a utility metric. The model $\Pmodel(\vy)$ estimates the probability of an output $\vy$ given an input sentence $\vx$. The utility metric $u(\vh, \vy)$ estimates the quality of a candidate output $\vh$ given a reference output $\vy$.
Given a set of candidate hypotheses $\CandH \subseteq \mathcal{Y}$, MBR decoding selects the best hypothesis according to its expected utility:
\begin{equation}
    \vh^{\mathrm{human}} = \argmax_{\vh \in \CandH} \sum_{\vy \in \mathcal{Y}} u(\vh, \vy) \cdot P_{\mathrm{human}}(\vy).
\end{equation}
Since $P_\mathrm{human}$ is unknown, MBR instead uses the model probability $\Pmodel$ to approximate $P_\mathrm{human}$.
\begin{equation}
    \vh^{\mathrm{model}} =  \argmax_{\vh \in \CandH} \sum_{\vy \in \mathcal{Y}} u(\vh, \vy) \cdot \Pmodel(\vy).    
\label{eq:mbr}
\end{equation}
For the rest of the paper, we will denote $\Pmodel$ as $\Pm$ for simplicity, unless otherwise noted.
Since integration over $\mathcal{Y}$ is computationally intractable, Eq.~\eqref{eq:mbr} is approximated by a Monte Carlo estimate \cite{eikema-aziz-2022-sampling,farinhas2023empirical} using a set of reference hypotheses $\RefH \subseteq \mathcal{Y}$ sampled from the model $\Pm$:
\begin{align}
    \vh^{\mathrm{MBR}} = \argmax_{\vh \in \CandH} \frac{1}{N} \sum_{\vy \in \RefH} u(\vh, \vy),
\label{eq:empirical-p}
\end{align}
where $N = |\RefH|$.
Standard practice is to use the same set of hypotheses for the candidate pool ($\mathcal{H}$) and the reference pool ($\RefH$), therefore $\mathcal{H} = \RefH$.

\section{Minimum Bayes Risk Decoding with Diversity}
We now introduce MBR decoding to the set decoding problem with diversity objective (Eq.~\ref{eq:setdiv}).
A naive way to generate $k$ sentences by MBR decoding is to select the top-$k$ hypotheses by Eq.~\eqref{eq:mbr}:
\begin{align}
    H^{\mathrm{MBR}} &= \argmax_{\stackrel{H \subseteq \CandH}{|H| = k}} \sum_{\vh \in H}\frac{1}{N} \sum_{\vy \in \RefH} u(\vh, \vy)
\label{eq:mbrset}
\end{align}
However, it results in a set of similar sentences with a large overlap. In fact, it often finds almost duplicated sentences in machine translation tasks (Tables~\ref{tab:deen-output} and \ref{tab:ruen-output} in Appendix~\ref{sec:sample}). 

\subsection{Diverse MBR (DMBR)}
We propose Diversity MBR (DMBR) decoding, a variant of MBR decoding with a diversity penalty $d: 2^\mathcal{Y} \rightarrow \mathbb{R}$ added to the decoding objective. The objective of the DMBR is the following:
\begin{align}
    &H^{\mathrm{DMBR}} = \nonumber\\
    &\;\;\argmax_{\stackrel{H \subseteq \CandH}{|H| = k}} \sum_{\vh \in H}\left( \frac{1}{N} \sum_{\vy \in \RefH} u(\vh, \vy) \right) + d(H) \label{eq:dmbr} 
\end{align}
The objective for $H^{\mathrm{DMBR}}$ consists of two terms: quality objective and diversity objective. The quality objective is the expected utility, the same as the original MBR $H^{\mathrm{MBR}}$. 

$d(H)$ is the diversity objective that penalizes according to a user-defined diversity objective. In this paper, we aim to promote diversity by minimizing the pairwise similarity among the outputs. Since the utility function typically computes the similarity between each pair of texts, we minimize the following pairwise utility:
\begin{equation}
    d(H) = -\sum_{\vh \in H} \sum_{\vh' \in H \setminus \{h\}} \frac{\lambda}{|H|} u(\vh, \vh').
\label{eq:pairwise}
\end{equation}
Assuming $u>0$, Eq.~\eqref{eq:dmbr} with the pairwise similarity results in a non-monotonic submodular function maximization problem \cite{DBLP:books/tf/18/BuchbinderF18} (proof in Appendix~\ref{sec:proof}).
Because solving a non-monotonic submodular function maximization problem is NP-hard \cite{feige1998threshold}, we deploy a greedy heuristic algorithm. We greedily select a hypothesis that maximizes the objective until we have $k$ hypotheses.
This procedure is guaranteed to find a solution with an approximation factor of $(1 - \frac{1}{e})$, provided that $\lambda$ is small enough to ensure the function is non-decreasing; otherwise, the approximation factor is slightly worse than $(1 - \frac{1}{e})$ \cite{nemhauser1978analysis}.
Note that DMBR still needs to compute $u(h, r)$ for every pair of candidates and references. Thus, even with the approximation, DMBR is at best as slow as MBR.\footnote{In our experiments, DMBR with 128 samples was roughly 2-4 times slower than diverse beam search with $k=4$ on \texttt{g4dn.xlarge} instances on AWS EC2 (4 vCPU cores, 16 GB memory, and an NVIDIA T4 GPU).}





\subsection{$k$-Medoids MBR (KMBR)}
As an alternative approach to promote diversity, we propose $k$-Medoids MBR (KMBR) decoding.
$k$-Medoids is a clustering problem similar to $k$-means in that it chooses centers to minimize the total distance of data points to the closest center points. The difference is that $k$-Medoids needs to choose actual data points as centers \cite{rdusseeun1987clustering,kaufman2009finding}. 
Intuitively, $k$-Medoids center points are supposed to be representative of different clusters of hypotheses. Thus, picking the center points for clusters is likely to result in a set of diverse and high-quality hypotheses.
KMBR can be understood as a generalization of the vanilla MBR decoding, which is solving the $1$-Medoid problem to find the single most centered hypothesis out of the sampled hypotheses \cite{jinnai2024hyperparameterfree}.
We use the negative utility as a distance and consider the problem of picking a set of $k$ center points. 
The total distance from the reference set (= data points) to the picked center points is minimized as follows:
\begin{align}
    H^{\mathrm{KMBR}} &= \argmax_{\stackrel{H \subseteq \CandH}{|H| = k}} \sum_{\vy \in \RefH} \min_{\vh \in H} -u(\vh, \vy).
\label{eq:KMBR}
\end{align}
$k$-medoids can be used with arbitrary dissimilarity measures. 
Because the $k$-medoids problem is NP-hard to solve exactly, we deploy Partition Around Medoids (PAM) to compute $H^{\mathrm{KMBR}}$ approximately \cite{park2009simple}. Similarly to DMBR, KMBR needs to compute the utility function for every pair of hypotheses so it is still slower than MBR even with the approximation algorithm.


\begin{figure*}[t]
    \begin{minipage}{.65\textwidth}
    \centering
    \begin{subfigure}[b]{0.48\textwidth}
        \includegraphics[width=\textwidth]{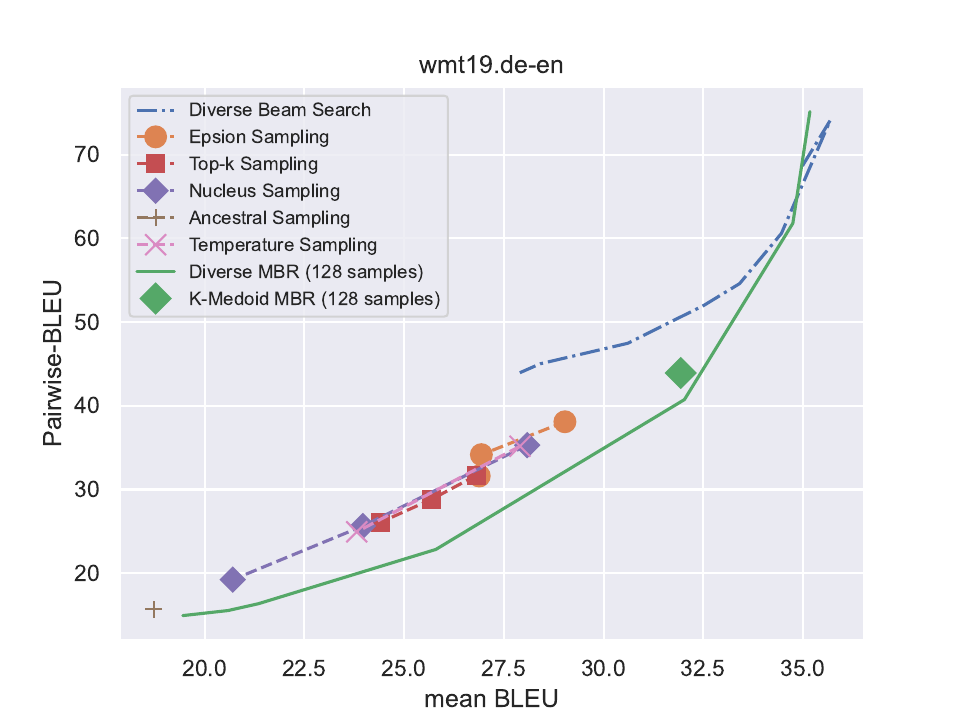}
        \caption{P-BLEU $\downarrow$ (De-En)}
    \end{subfigure}
    \begin{subfigure}[b]{0.48\textwidth}
        \includegraphics[width=\textwidth]{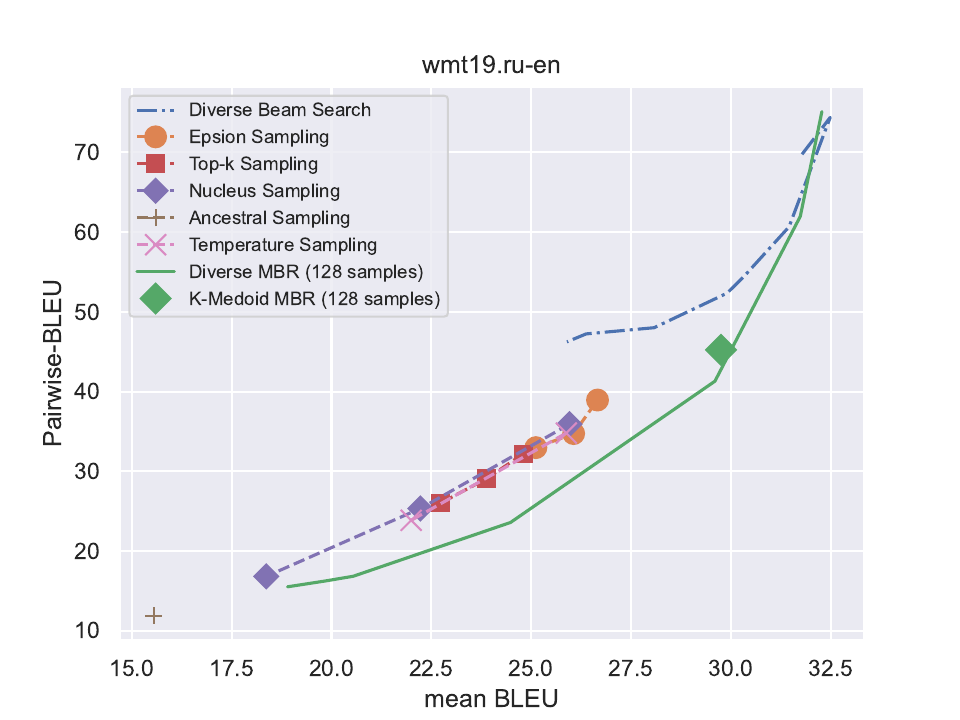}
        \caption{P-BLEU $\downarrow$ (Ru-En)}
    \end{subfigure} \\
    \begin{subfigure}[b]{0.48\textwidth}
        \includegraphics[width=\textwidth]{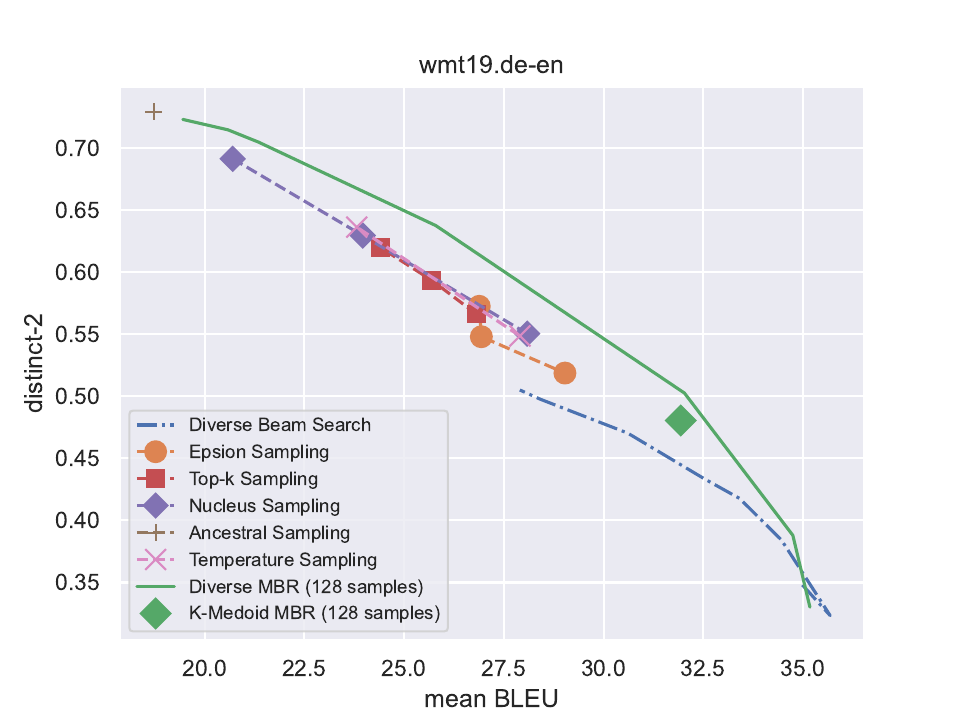}
        \caption{distinct-2 $\uparrow$ (De-En)}
    \end{subfigure}
    \begin{subfigure}[b]{0.48\textwidth}
        \includegraphics[width=\textwidth]{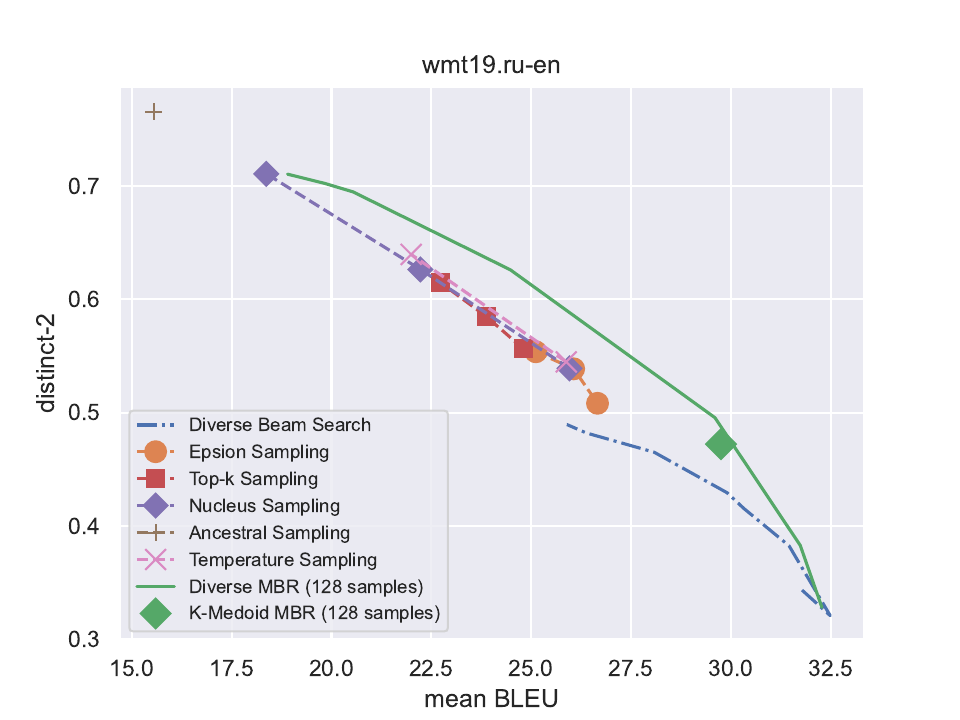}
        \caption{distinct-2 $\uparrow$ (Ru-En)}
    \end{subfigure}
    \caption{Evaluation of P-BLEU and distinct-2 as a function of mean BLEU on WMT'19 De-En and Ru-En. The size of the outputs $k$ is set to 4. $\uparrow$ and $\downarrow$ denote that larger and smaller are better in diversity, respectively.}  
    \label{fig:wmt}        
    \end{minipage}\hspace{16pt}
    \begin{minipage}{.31\textwidth}
    \centering
    \begin{subfigure}[b]{\textwidth}
        \includegraphics[width=\textwidth]{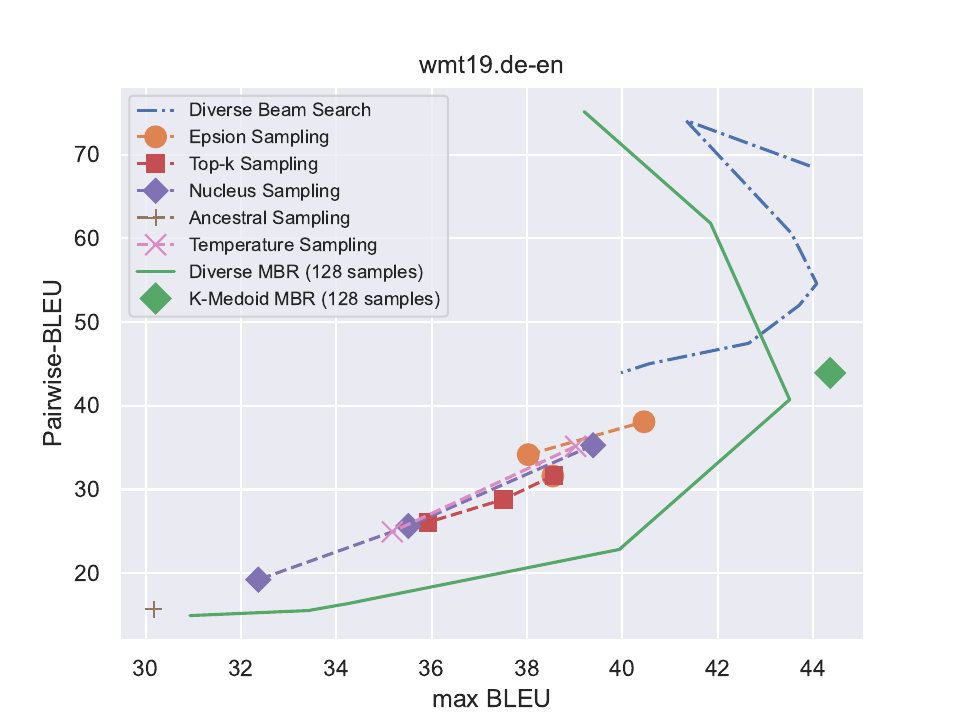}
        \caption{P-BLEU $\downarrow$ (De-En)}
    \end{subfigure} \\
    \begin{subfigure}[b]{\textwidth}
        \includegraphics[width=\textwidth]{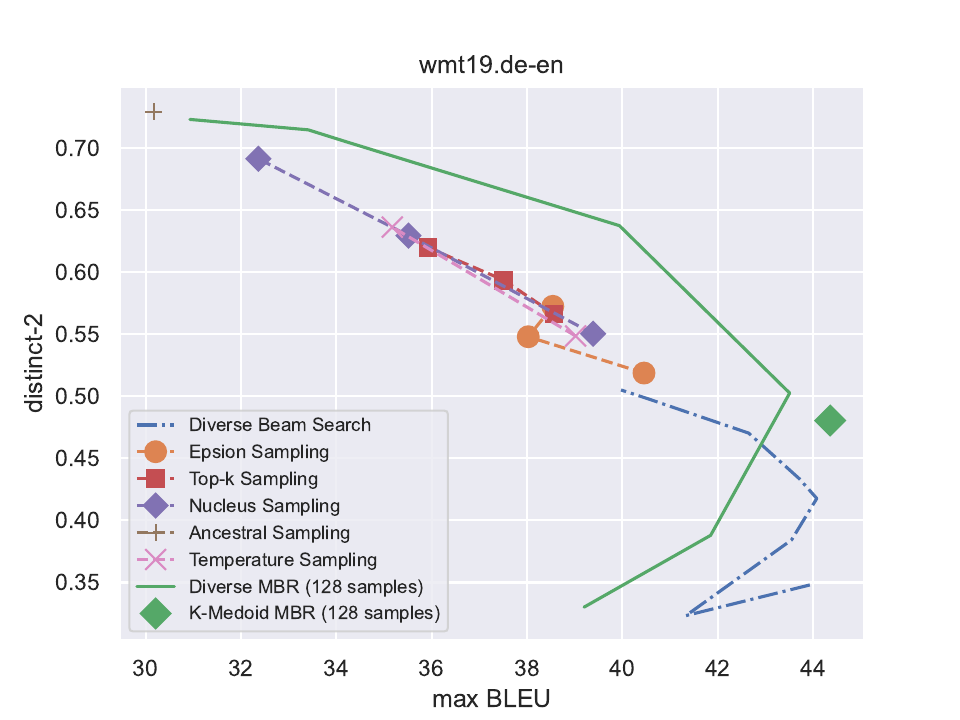}
        \caption{distinct-2 $\uparrow$ (De-En)}
    \end{subfigure}
    \caption{Evaluation of P-BLEU, distinct-2 as a function of max BLEU (Oracle score) on WMT'19 De-En. The size of the outputs $k$ is set to 4.
    }  
    \label{fig:oracle}
    \end{minipage}
\end{figure*}

\section{Experiments}
\label{sec:experiments}

We evaluate DMBR and KMBR in machine translation, image captioning, question generation, generative common sense reasoning, and text summarization.
All experiments use BERTScore \cite{bert-score} as the utility function of the MBR.
We compare the performance of the sampling algorithms, beam search, diverse beam search (DBS) \cite{vijayakumar2018diverse}, MBR, diverse MBR (DMBR), and $k$-Medoids MBR (KMBR).

We evaluate the quality and the diversity of the generated texts.
As a quality metric, we report the mean, max, and min quality over each set of $k$ sentences measured by BLEU, ROUGE-L, or METEOR \cite{papineni-etal-2002-bleu,lin-och-2004-automatic,banerjee-lavie-2005-meteor}. 
We use \textbf{distinct-n} \cite{li-etal-2016-diversity} and \textbf{pairwise-BLEU (P-BLEU)} \cite{shen2019mixture} as diversity metrics.

Due to limitations in computational resources, we run experiments using the first 1000 entries of the dataset for all the experiments in this paper.
We use Huggingface's Transformers library for running all the experiments \cite{wolf-etal-2020-transformers}.
We initialize the clusters for PAM used in KMBR by k-medoids++ with maximum iterations set to $300$.
For reproducibility, all the experiments are conducted using publicly available pretrained models and datasets.
We use sacreBLEU system \cite{post-2018-call} to compute BLEU scores.

Examples of generations are shown in Appendix~\ref{sec:sample}.
See Appendix~\ref{sec:results} for additional figures and the summary of the experimental results in tables.

\begin{figure}[tb]
    \centering
    \begin{subfigure}[b]{0.9\columnwidth}
        \includegraphics[width=\textwidth]{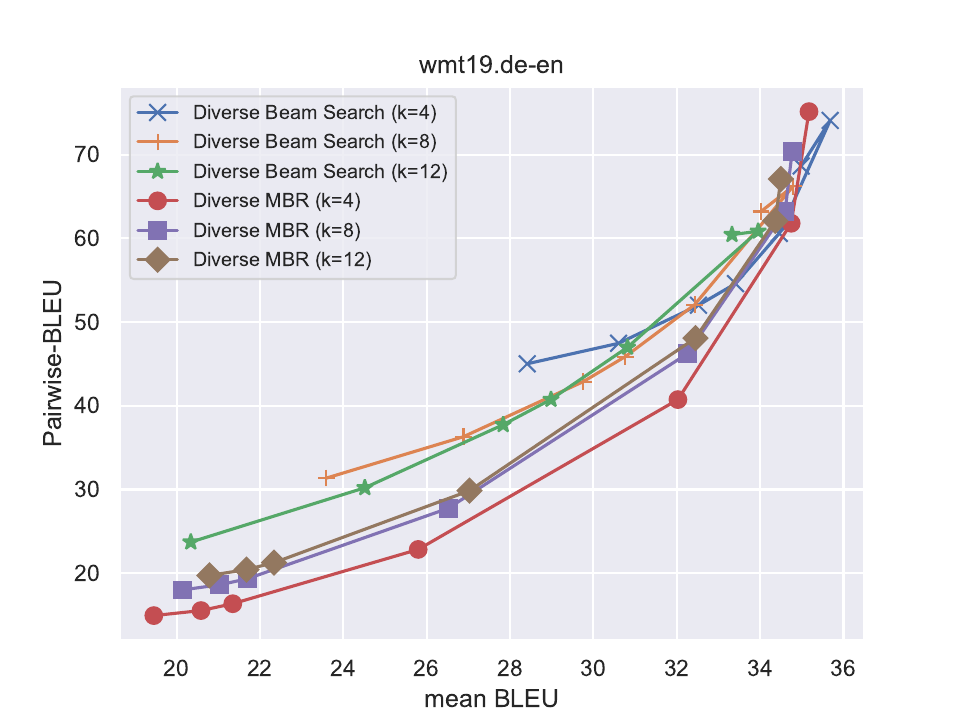}
        \caption{P-BLEU $\downarrow$ with varying \#outputs}
        \label{fig:outputs-pbleu}
    \end{subfigure}
    \begin{subfigure}[b]{0.9\columnwidth}
        \includegraphics[width=\textwidth]{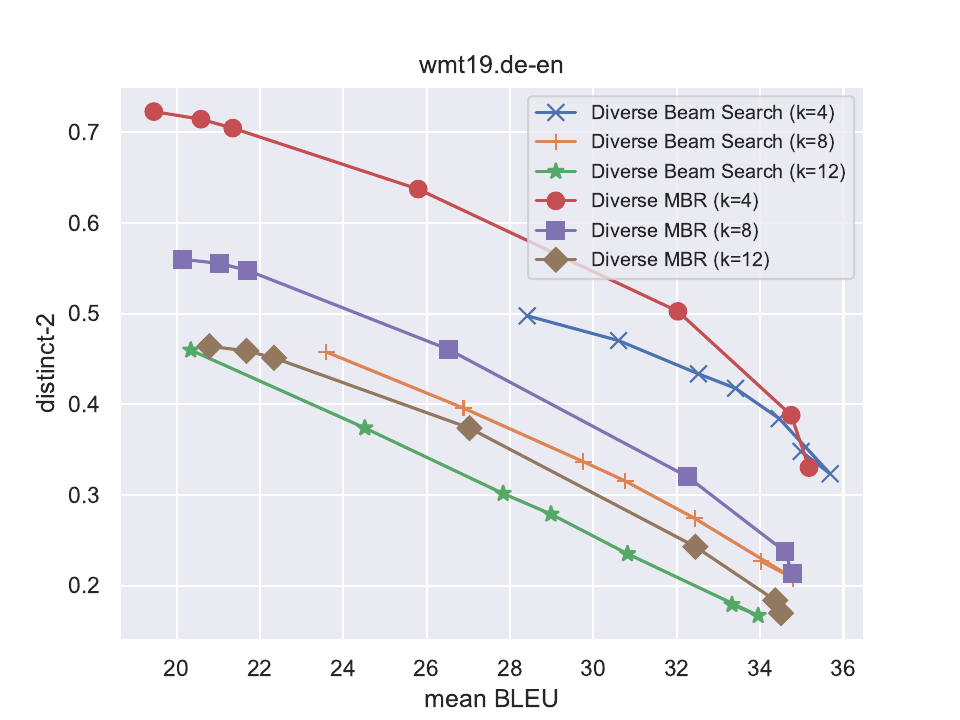}
        \caption{distinct-2 $\uparrow$ with varying \#outputs}
        \label{fig:outputs-dn}
    \end{subfigure}
    \caption{Evaluation of DMBR and KMBR with varying number of outputs ($k \in \{4, 8, 12\}$).
    Mean BLEU, P-BLEU, and distinct-2 on WMT'19 De-En are reported.}  
    \label{fig:noutputs}
\end{figure}

\begin{figure}[tb]
    \centering
    \begin{subfigure}[b]{0.9\columnwidth}
        \includegraphics[width=\textwidth]{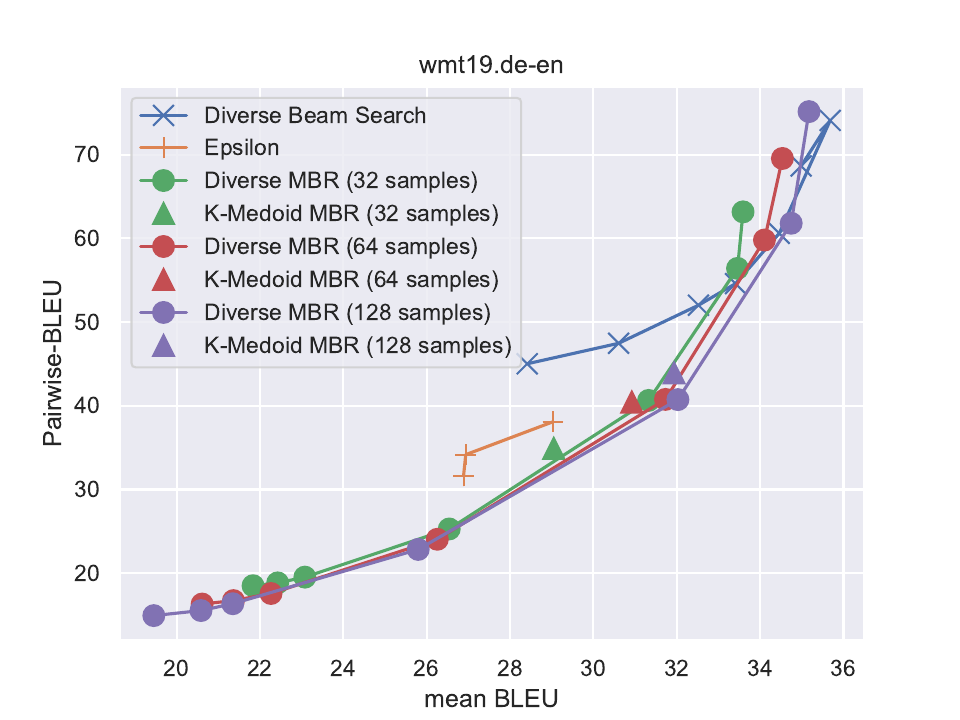}
        \caption{P-BLEU $\downarrow$ with varying \#samples}
        \label{fig:samples-pbleu}
    \end{subfigure}
    \begin{subfigure}[b]{0.9\columnwidth}
        \includegraphics[width=\textwidth]{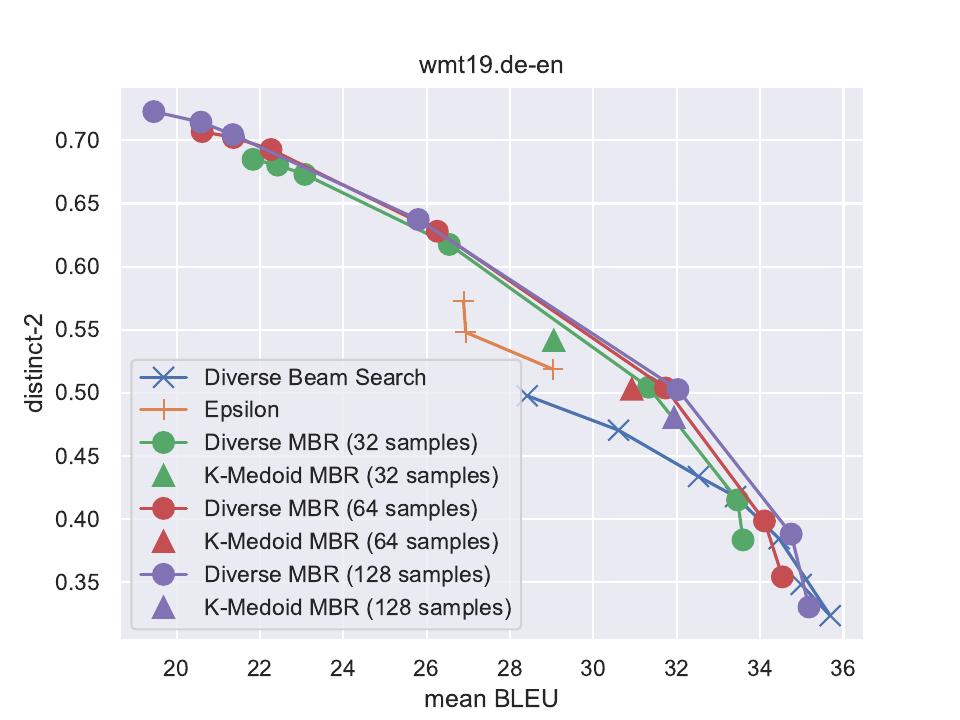}
        \caption{distinct-2 $\uparrow$ with varying \#samples}
        \label{fig:samples-dn}
    \end{subfigure}
    \caption{
    Evaluation of DMBR and KMBR with varying number of samples ($N \in \{32, 64, 128\}$). Mean BLEU, P-BLEU, and distinct-2 on WMT'19 De-En are reported.}  
    \label{fig:nsamples}
\end{figure}

\begin{figure}[t]
    \centering
    \begin{subfigure}[b]{0.95\columnwidth}
        \includegraphics[width=\textwidth]{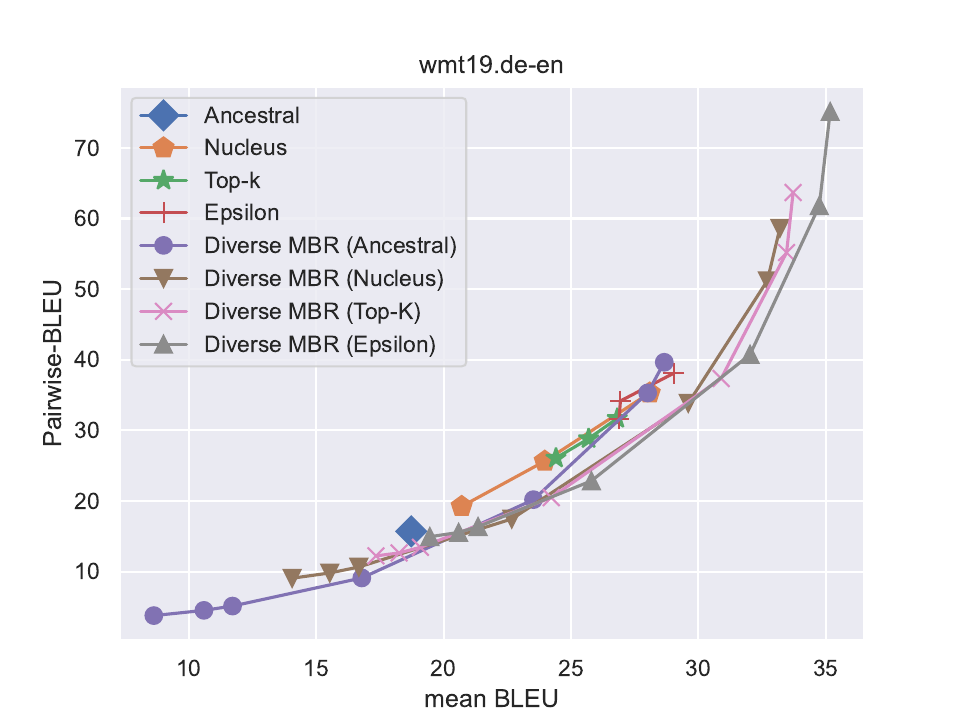}
        \caption{P-BLEU $\downarrow$ with varying sampling}
        \label{fig:samplings-pbleu}
    \end{subfigure}
    \begin{subfigure}[b]{0.95\columnwidth}
        \includegraphics[width=\textwidth]{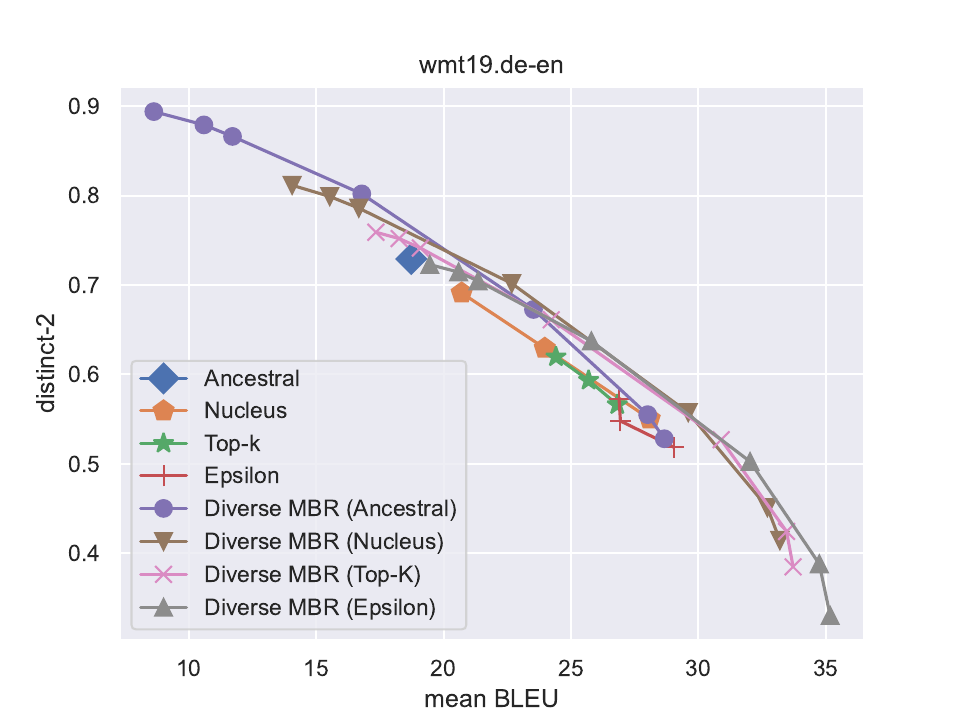}
        \caption{distinct-2 $\uparrow$ with varying sampling}
        \label{fig:samplings-dn}
    \end{subfigure}
    \caption{Evaluation of DMBR and KMBR using varying sampling algorithms: ancestral sampling, nucleus sampling, top-$k$ sampling, and epsilon sampling.. Mean BLEU, P-BLEU, and distinct-2 on WMT'19 De-En are reported.}  
    \label{fig:samplings}
\end{figure}

\subsection{Machine Translation}
\label{sec:mt}
We use the WMT'19 dataset \cite{barrault-etal-2019-findings} to evaluate the performance on machine translation tasks.
WMT'19 dataset examines translation between English and other languages in the news domain.
We run experiments on two language pairs: German $\rightarrow$ English (De$\rightarrow$En) and Russian $\rightarrow$ English (Ru$\rightarrow$En) using the pretrained models of each language pair provided by fairseq \cite{ng-etal-2019-facebook}.

We set the number of outputs to be $k \in \{4, 8, 12\}$.
We compare the performance of sampling algorithms, diverse beam search, and the proposed methods.
For sampling algorithms, we evaluate ancestral sampling, nucleus sampling with $p \in \{0.8, 0.9, 0.95\}$, top-$k$ sampling with $k_{\mathrm{top}} \in \{5, 10, 50\}$, epsilon sampling with $\epsilon \in \{0.01, 0.02, 0.04\}$, and temperature sampling with $T \in \{0.8, 0.9\}$ \cite{Holtzman2020The,fan-etal-2018-hierarchical,hewitt-etal-2022-truncation}.
For DBS \cite{vijayakumar2018diverse}, we set the number of groups to be equal to the beam width ($k$). The strength of diverse penalty $\lambda$ for DBS is set to $\{0.0 \:\text{(beam search)}, 0.2, 0.5, 1.0, 2.0, 5.0,$ $10.0, 20.0\}$.
For MBR-based decoding methods, we set the sample size $N = 128$ per source sentence (see Eq.~\ref{eq:empirical-p}) and use epsilon sampling with $\epsilon=0.02$ \cite{hewitt-etal-2022-truncation,freitag2023epsilon}. The diversity penalty $\lambda$ is set $\{0.0 \:\text{(vanilla MBR)}, 0.1, 0.3, 0.5, 1.0, 2.0\}$. 

Due to space limitations, we discuss the essential results here and show the detailed results in Tables~\ref{tab:wmt}, \ref{tab:wmt-deen-k}, and \ref{tab:wmt-ruen-k} in Appendix \ref{sec:results}.

\paragraph{DMBR achieves higher diversity than baselines.}
Figure~\ref{fig:wmt} shows the P-BLEU and distinct-2 as a function of mean BLEU score. The results show that DMBR achieves higher diversity (lower P-BLEU and higher distinct-2) than DBS and sampling algorithms with the same mean BLEU score. 


\paragraph{DMBR achieves more flexibility than DBS on the quality-diversity trade-off.}
We observe that DBS does not increase the diversity by increasing the diversity penalty larger than $10.0$ (Table \ref{tab:wmt} in Appendix). On the other hand, the diversity of DMBR continue to increase with larger $\lambda$, achieving higher maximum diversity than DBS.
DMBR also achieves higher diversity than the sampling algorithms. 


\paragraph{DMBR achieves higher oracle score than vanilla MBR.}
The Oracle score indicates the score of the highest-scoring output in a set of $k$ outputs \cite{vijayakumar2018diverse}. It is intended to evaluate how well the obtained diversity benefits in producing output close to a particular correct answer.
Figure~\ref{fig:oracle} shows the P-BLEU and distinct-2 as a function of max BLEU score (i.e., Oracle score). We observe that DMBR achieves a slightly lower max BLEU score than DBS, yet with higher diversity.
Interestingly, we observe that DMBR and KMBR achieve higher max scores than vanilla MBR (Table \ref{tab:wmt}). This shows the potential of DMBR to further improve the quality of the output of MBR. This is analogous to DBS achieving a higher Oracle score than beam search \cite{Vijayakumar2016}.
KMBR achieves the highest max BLEU score over all algorithms compared. 


\paragraph{DMBR outperforms DBS with varying output sizes.}
Figure~\ref{fig:noutputs} shows the quality and diversity trade-off with varying output size $k \in \{4, 8, 12\}$.
DBS shows no degradation of P-BLEU but shows a lower distinct-2 score with a larger output size. Overall, we observe that DMBR outperforms DBS in both diversity metrics with all the output sizes we evaluated.

\paragraph{DMBR improves with a larger number of samples.} Figure~\ref{fig:nsamples} shows the performance of DMBR and KMBR with varying numbers of samples $N \in \{32, 64, 128\}$. We observe that the quality-diversity trade-off is improved with a larger sample size, yet it does not significantly improve by doubling the sample size.
The results indicate that one can improve both quality and diversity by increasing the sample size, but it comes at the cost of inference time. Note that the computational complexity of DMBR is quadratic to the number of samples.

\paragraph{Choice of the underlying sampling algorithm matters.} Figure~\ref{fig:samplings} shows the comparison of DMBR with varying sampling algorithms. We evaluate ancestral sampling, nucleus sampling with $p \in \{0.8, 0.9, 0.95\}$, top-$k$ sampling with $k_{\mathrm{top}} \in \{5, 10, 50\}$, and epsilon sampling with $\epsilon \in \{0.01, 0.02, 0.04\}$ \cite{Holtzman2020The,fan-etal-2018-hierarchical,hewitt-etal-2022-truncation}. The temperature is set to 1 for all the runs.
We observe that the choice of the sampling algorithm changes the Pareto front of the DMBR. Using less biased sampling algorithms such as ancestral sampling, DMBR improves the diversity, and using more focused sampling algorithms such as epsilon sampling, DMBR improves the mean quality of the outputs.

\begin{figure*}[t]
    \centering
    \begin{subfigure}[b]{0.325\textwidth}
        \includegraphics[width=\textwidth]{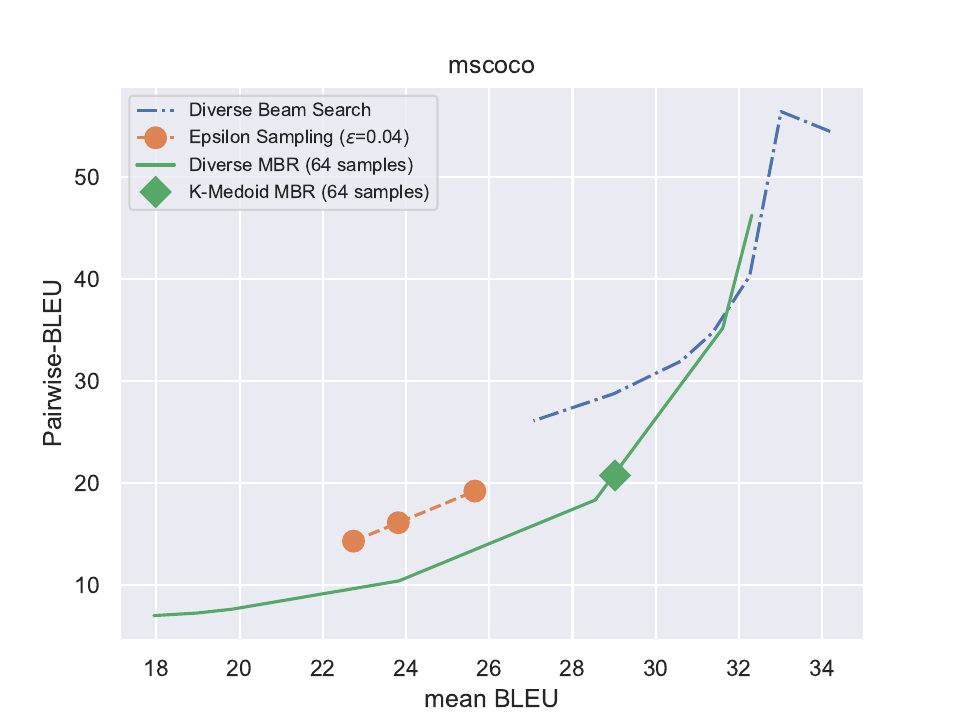}
        \caption{P-BLEU $\downarrow$ (MS COCO)}
        \label{fig:pbleu-mscoco}
    \end{subfigure}
    \begin{subfigure}[b]{0.325\textwidth}
        \includegraphics[width=\textwidth]{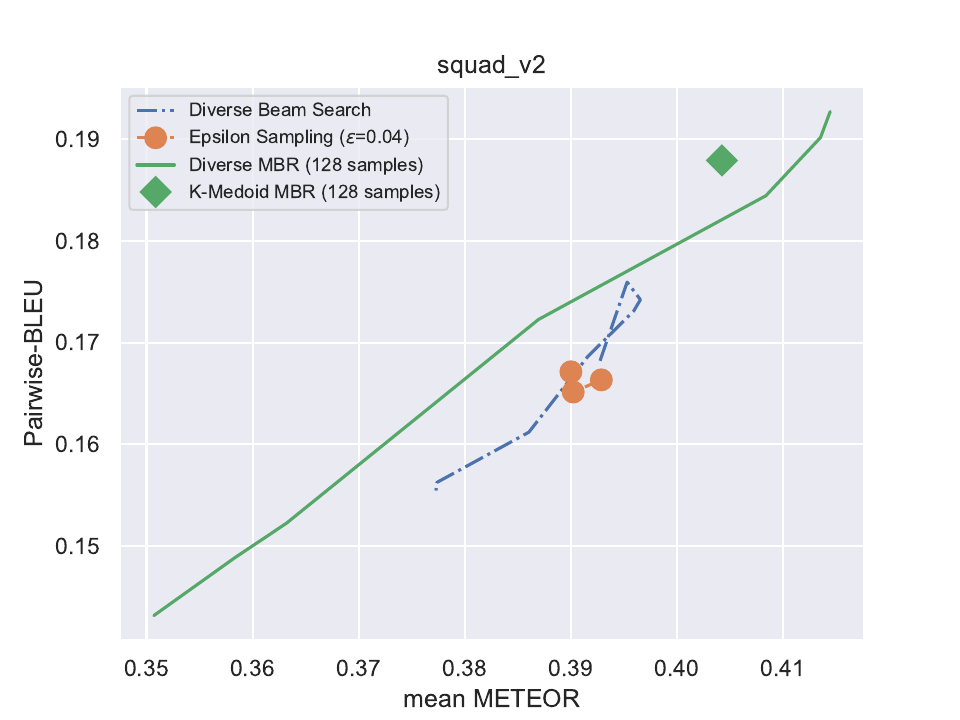}
        \caption{P-BLEU $\downarrow$ (SQuADv2)}
        \label{fig:pbleu-squad}
    \end{subfigure}
    \begin{subfigure}[b]{0.325\textwidth}
        \includegraphics[width=\textwidth]{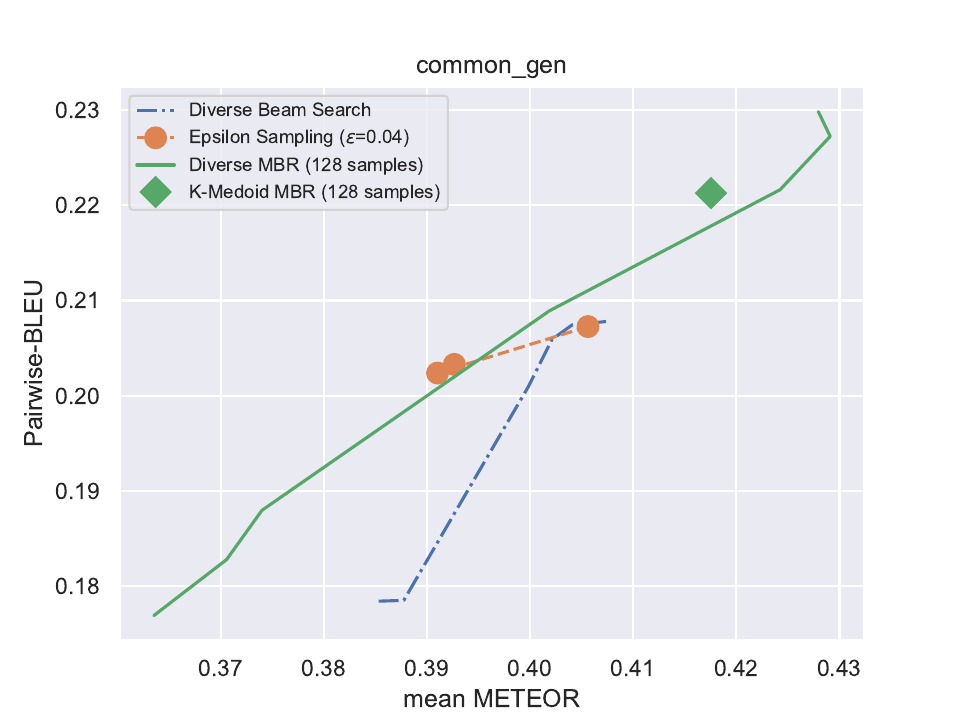}
        \caption{P-BLEU $\downarrow$ (CommonGen)}
        \label{fig:pbleu-common}
    \end{subfigure} \\
    \begin{subfigure}[b]{0.325\textwidth}
        \includegraphics[width=\textwidth]{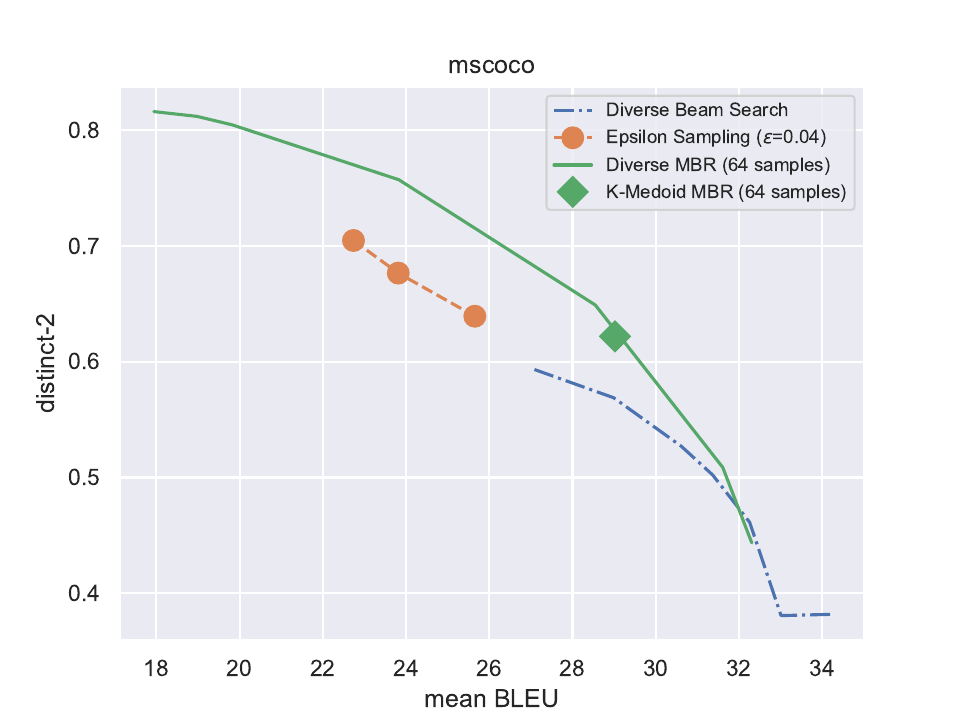}
        \caption{distinct-2 $\uparrow$ (MS COCO)}
        \label{fig:dist2-mscoco}
    \end{subfigure}
    \begin{subfigure}[b]{0.325\textwidth}
        \includegraphics[width=\textwidth]{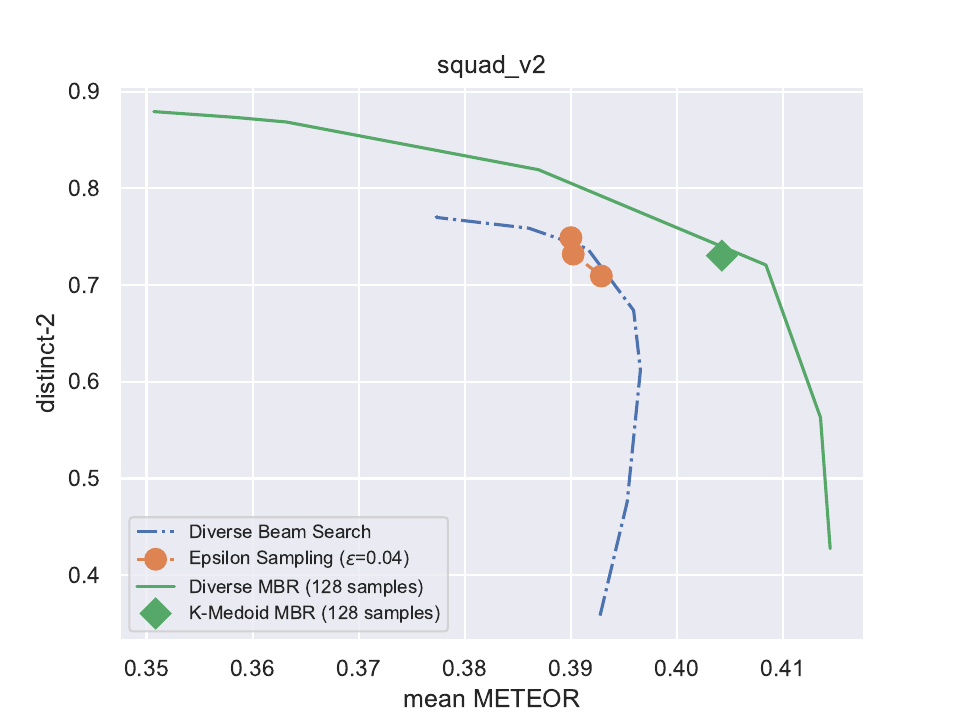}
        \caption{distinct-2 $\uparrow$ (SQuADv2)}
        \label{fig:dist2-squad}
    \end{subfigure}
    \begin{subfigure}[b]{0.325\textwidth}
        \includegraphics[width=\textwidth]{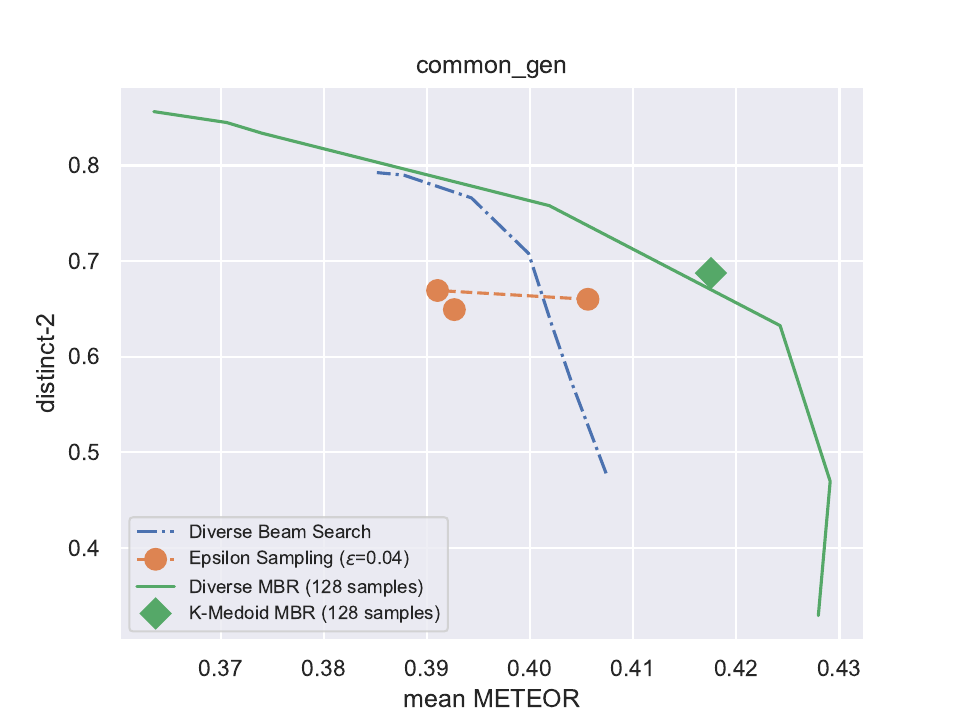}
        \caption{distinct-2 $\uparrow$ (CommonGen) }
        \label{fig:dist2-common}
    \end{subfigure} \\
    \begin{subfigure}[b]{0.325\textwidth}
        \includegraphics[width=\textwidth]{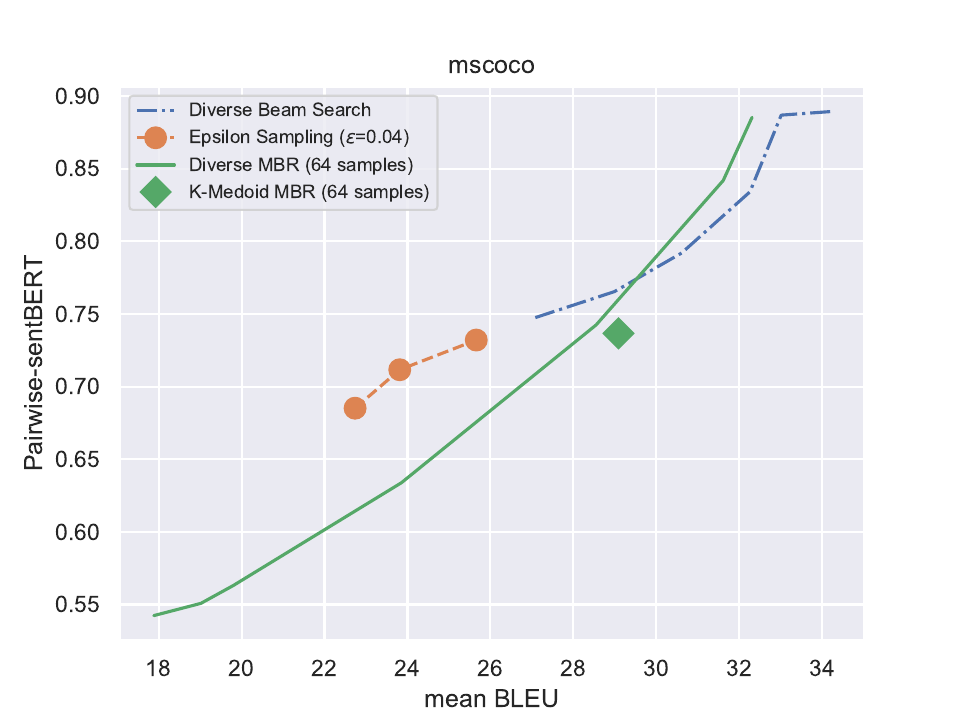}
        \caption{P-SentBERT $\downarrow$ (MS COCO)}
        \label{fig:sentbert-mscoco}
    \end{subfigure}
    \begin{subfigure}[b]{0.325\textwidth}
        \includegraphics[width=\textwidth]{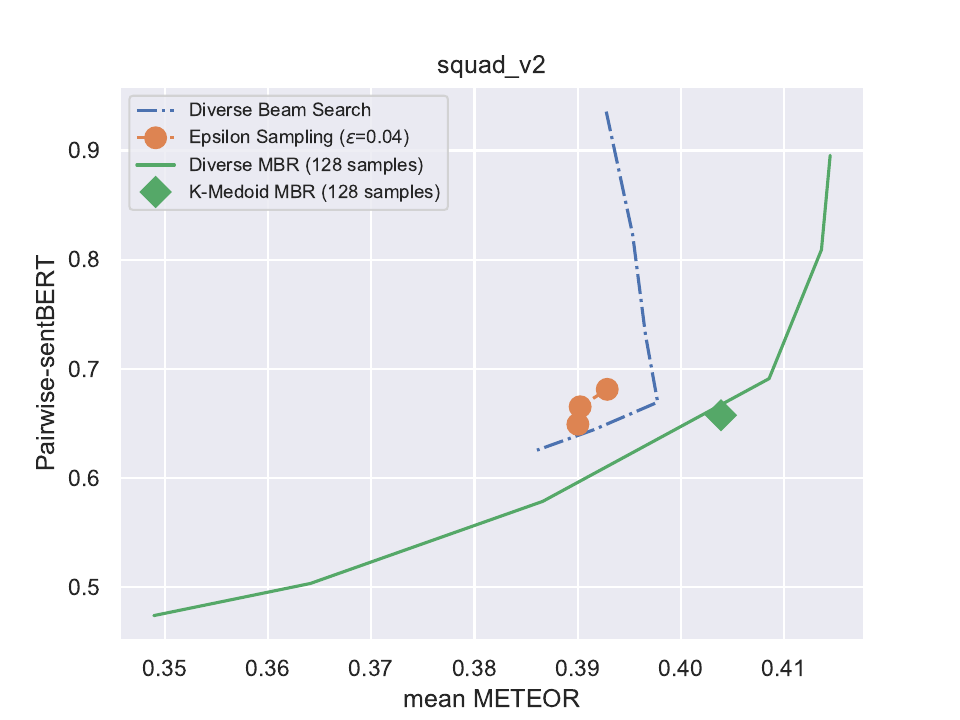}
        \caption{P-SentBERT $\downarrow$ (SQuADv2) }
        \label{fig:sentbert-squad}
    \end{subfigure}
    \begin{subfigure}[b]{0.325\textwidth}
        \includegraphics[width=\textwidth]{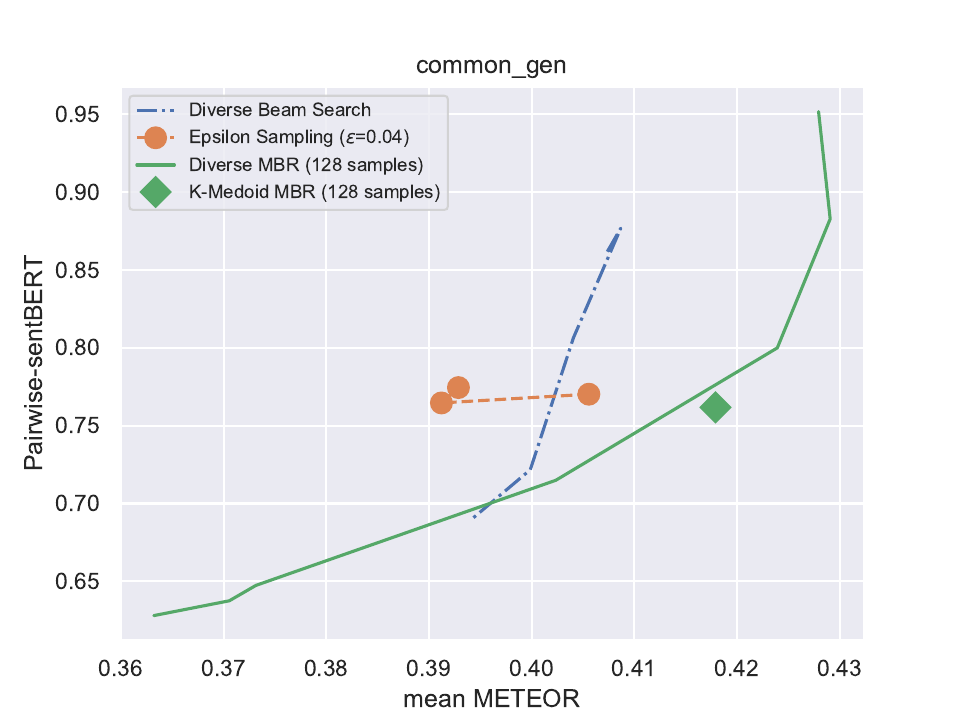}
        \caption{P-SentBERT $\downarrow$ (CommonGen)}
        \label{fig:sentbert-common}
    \end{subfigure}
    
    \caption{Evaluation of P-BLEU, distinct-2, and P-SentBERT as a function of mean BLEU (MS COCO) and METEOR (SQuADv2 and CommonGen). The number of the outputs $k$ is $4$.}
    \label{fig:others}
\end{figure*}

\subsection{Image Captioning using BLIP-2}
\label{sec:captioning}


We evaluate the performance of the proposed methods on image captioning using MS COCO dataset \cite{lin2014microsoft}.
We use BLIP-2 \cite{pmlr-v202-li23q} with Flan T5-xl \cite{chung2022scaling} fine-tuned for MS COCO. We load the model in 8-bit to reduce the VRAM consumption. 
The number of outputs $k$ is set to $\{4, 8, 12\}$.
We evaluate the performance of epsilon sampling with $\epsilon \in \{0.01, 0.02, 0.04\}$.
The diversity penalty for DBS is set to $\{0.0, 0.2, 0.5, 1.0, 2.0, 5.0\}$.
For DMBR, we generate $N=64$ samples using epsilon sampling with $\epsilon=0.02$.
The results on $k=4$ are shown in Figure~\ref{fig:others} (\subref{fig:pbleu-mscoco}, \subref{fig:dist2-mscoco},  \subref{fig:sentbert-mscoco}). DMBR achieves lower P-BLEU and higher distinct-2 than DBS and epsilon sampling.

\paragraph{Evaluation of Semantic Diversity.} While the machine translation task requires generating a text semantically the same to the input text, the image captioning task allows more diversity in the contents of the output \cite{WangC19}. To evaluate the semantic diversity beyond the surface diversity, we employ sentence BERT \cite{reimers-gurevych-2019-sentence}.
We compute the embedding of each output using the sentence BERT and compute the cosine similarity of each pair of outputs. 
The cosine similarity over a pair of sentences is shown to have a high correlation to human preference \cite{tevet-berant-2021-evaluating}.
We evaluate the \textbf{pairwise sentence BERT (P-SentBERT)}, the average cosine similarity of the sentence embeddings over a set of pairs of outputs.
We use \textsc{all-mpnet-base-v2} model. 
The model is based on MPNet \cite{song2020mpnet} and has shown to be effective for a variety of sentence embedding tasks. 
The result in Figure~\ref{fig:sentbert-mscoco} shows that DMBR achieves better (lower) P-SentBERT than DBS and epsilon sampling.

\subsection{Question Generation using Language Model}
\label{sec:squad}

The goal of question generation is to generate a question on a topic in natural language from a paragraph of text \cite{question-generation2023}.
We use a Stanford Question Answering Dataset (SQuADv2) to evaluate the decoding algorithms for question generation \cite{rajpurkar-etal-2016-squad,rajpurkar-etal-2018-know}. SQuADv2 is a reading comprehension dataset consisting of questions and answers on Wikipedia articles.
We use a language model Zephyr-7B $\beta$ \cite{tunstall2023zephyr} with prompting as a text generation model. We use the following prompt: \\
{\it 
    \indent Given a paragraph provided by the user, generate a very short question one can answer by a word to test the understanding of the paragraph. Make sure that the question is very short. Do NOT include the answer.
}
\\
We generate $k \in \{4, 8, 12\}$ outputs. For DBS, we set the diversity penalty to $\{0.0, 0.5, 1.0, 2.0, 5.0\}$.
For MBR we generate $128$ samples with epsilon sampling with $\epsilon = 0.01$. The diversity penalty $\lambda$ for DMBR is set to $\{0.0, 0.1, 0.3, 0.5, 1.0, 2.0\}$.
The mean METEOR and the diversity metrics are shown in Figure~\ref{fig:others} (\subref{fig:pbleu-squad}, \subref{fig:dist2-squad}, \subref{fig:sentbert-squad}). Compared with the same METEOR score, DMBR has better distinct-2 and P-SentBERT than DBS.
The P-BLEU and P-sentBERT of DMBR are slightly worse than DBS and epsilon sampling compared with the same mean METEOR score. 

We speculate that DMBR underperforms DBS for SQuADv2 and CommonGen (Section~\ref{sec:commongen}) because DBS is more likely to generate a set of sequences of varying lengths. Table~\ref{tab:length} shows the average standard deviation of the sequences generated by DMBR and DBS with varying diversity penalties. DBS tends to generate a set of diverse lengths of sentences than DMBR in these domains. Because P-BLEU is sensitive to the difference in length whereas distinct-n and P-SentBERT are less sensitive, P-BLEU is a favorable metric for DBS. As such, DMBR underperforms DBS for SQuADv2 and CommonGen in P-BLEU but not in distinct-n and P-SentBERT. It implies that if a practitioner benefits from varying sequence lengths, DBS may be preferred over DMBR, and if not, DMBR may be preferred over DBS. 
\begin{table}
    \centering
    \begin{tabular}{lcccc}
    \toprule
    & \multicolumn{4}{c}{DMBR} \\
    $\lambda$ & 0.1 & 0.5 & 1.0 & 2.0 \\
    \cmidrule(l){2-5}
SQuADv2 & 3.73 & 4.29 & 4.30 & 5.38 \\
CommonGen & 3.36 & 3.69 & 4.41 & 6.86 \\\midrule
    & \multicolumn{4}{c}{DBS} \\
    $\lambda$& 0.5 & 1.0 & 2.0 & 5.0 \\
    \cmidrule(l){2-5}
SQuADv2 & 2.70 & 4.72 & 6.37 & 5.96 \\
CommonGen & 1.92 & 5.46 & 7.50 & 7.45 \\
\bottomrule
    \end{tabular}
    \caption{The standard deviation of the sequence lengths averaged over the inputs using DMBR and DBS. Note that the diversity strength $\lambda$ of the two algorithms are used differently.}
    \label{tab:length}
\end{table}

\subsection{Generative Common Sense Reasoning using Language Model}
\label{sec:commongen}
CommonGen is a constrained text generation task to evaluate the ability of common sense reasoning of the system \cite{lin-etal-2020-commongen}. Given a set of common concepts, the task is to generate a coherent sentence describing an everyday scenario using these concepts. 
We use Zephyr-7B $\beta$ with prompting as a text generation model.
We use the following prompt: \\
{\it 
    \indent Generate a short and interesting sentence using all the words provided from the user as is. Make sure that it is short and all the words are included without rewording.
}
\\
The other experimental setting is the same as in Section~\ref{sec:squad}.
The mean METEOR and the diversity metrics are shown in Figure~\ref{fig:others} (\subref{fig:pbleu-common}, \subref{fig:dist2-common}, \subref{fig:sentbert-common}). DMBR has better distinct-2 than DBS and epsilon sampling but slightly worse P-BLEU. 
We observe that the coverage of the input concepts is low ($<18\%$) in all settings in our experiments. See Appendix~\ref{sec:coverage} for the analysis of the coverage.

\begin{figure}
    \centering
    \begin{subfigure}[b]{0.66\columnwidth}
        \includegraphics[width=\textwidth]{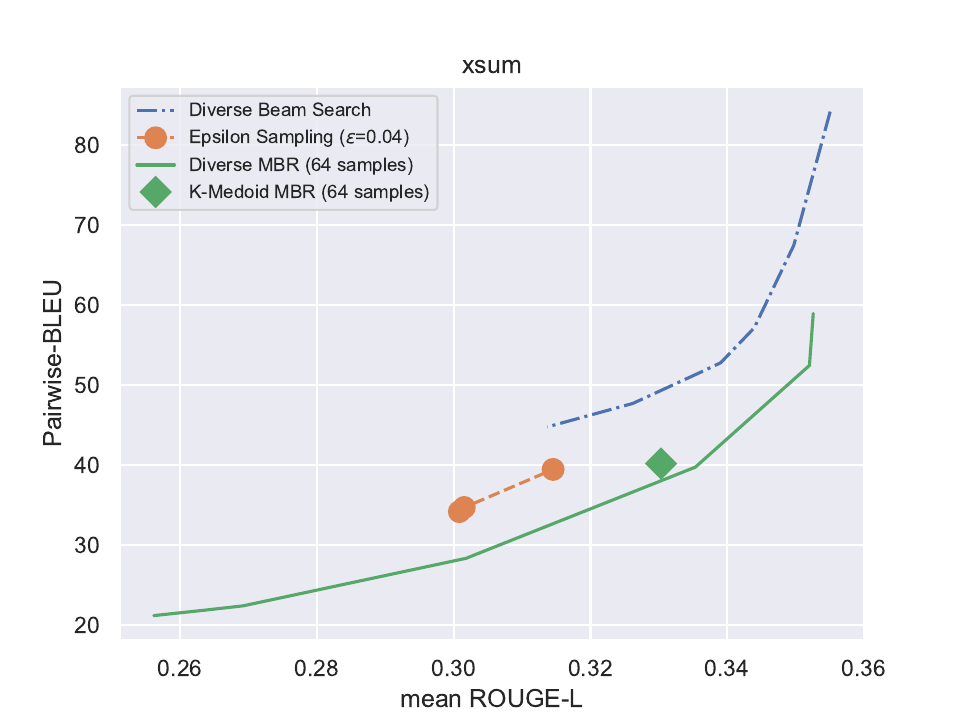}
        \caption{P-BLEU $\downarrow$ (XSum)}
    \end{subfigure}
    \begin{subfigure}[b]{0.66\columnwidth}
        \includegraphics[width=\textwidth]{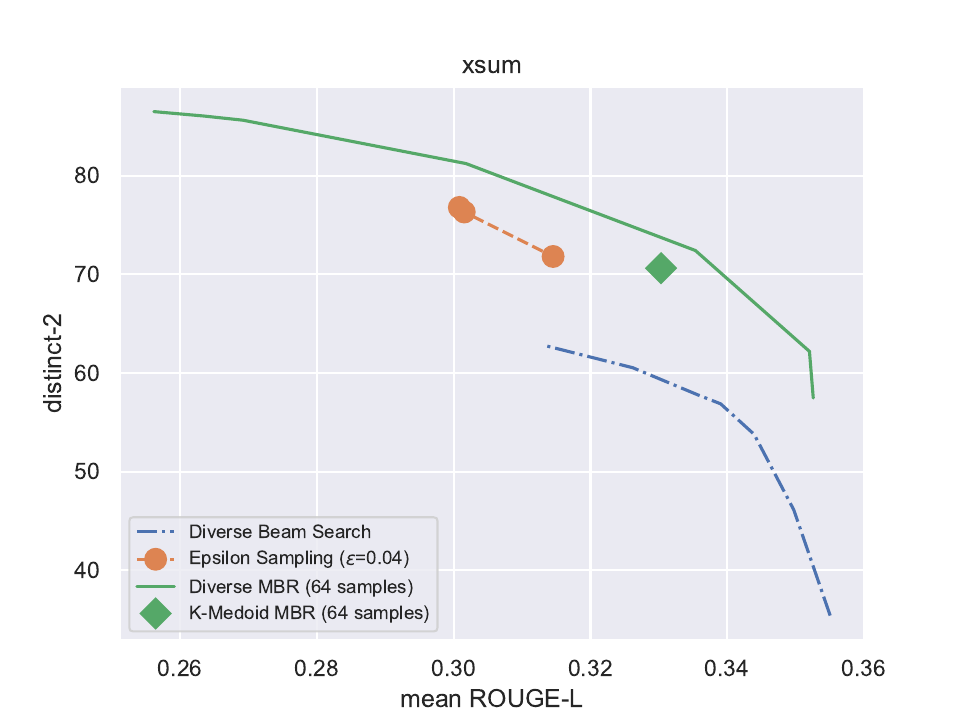}
        \caption{distinct-2 $\uparrow$ (XSum) }
    \end{subfigure}
    \caption{Evaluation of the P-BLEU and distinct-2 as a function of mean ROUGE-L. The number of the outputs $k$ is $4$.}
    \label{fig:xsum}
\end{figure}

\subsection{Text Summarization}
\label{sec:sum}
We use the XSum dataset as a benchmark for an abstractive single-document summarization \cite{narayan-etal-2018-dont}.
We use a BART model pretrained on XSum dataset \cite{lewis-etal-2020-bart}. 
We evaluate the quality of the outputs by ROUGE-L using HuggingFace's evaluate library \cite{lin-2004-rouge}.
We generate $k=4$ outputs per each input document.
For DBS, we set the diversity penalty to $\{0.0, 0.5, 1.0, 2.0, 5.0\}$.
For MBR we generate 64 samples with epsilon sampling with $\epsilon=0.02$. The diversity penalty $\lambda$ for DMBR is set to $\{0.0, 0.1, 0.3, 0.5, 1.0, 2.0\}$.
The results are shown in Figure~\ref{fig:xsum}. DMBR achieves better diversity than DBS measured by P-BLEU and distinct-n.

\section{Conclusions}

We study the problem of generating a set of texts with high quality and diversity.
Our approach is to promote diversity to MBR which is shown to generate high quality texts so that it can generate high quality and diverse outputs.
We extend MBR and propose DMBR and KMBR that seek to optimize both the diversity and the quality when selecting a set of outputs.
Because both algorithms are too expensive to compute exactly, we devise approximate algorithms to make it feasible.
We evaluate DMBR and KMBR on machine translation, image captioning, question generation, generative common sense reasoning, and text summarization tasks and show that overall they achieve better quality-diversity trade-off than DBS.
We also observe that both methods, especially KMBR, achieve a higher max BLEU score than MBR, analogous to DBS achieving a higher BLEU score than beam search.


\section{Limitations}

Our experiments are focused on directed text generation tasks.
Open-ended and directed text generation tasks are different tasks for text generation algorithms. This distinction separates what kind of text generation algorithms are suitable for each task. While beam search variants tend to perform better in directed text generation tasks, stochastic decoding algorithms tend to do better in open-ended text generation tasks \cite{Holtzman2020The,basu2021mirostat,hewitt-etal-2022-truncation,xu-etal-2023-look}.
The evaluation of the methods in open-ended text generation tasks is future work.

We rely on automatic evaluations to evaluate the quality and the diversity of the generated texts. Human evaluation is desirable, especially for evaluating diversity. 
Although the automatic evaluation metrics for diversity used in this paper (e.g., P-BLEU, distinct-n, P-SentBERT) are shown to correlate with human evaluation, there is still a clear gap between automatic metrics and humans \cite{tevet-berant-2021-evaluating}.

DMBR and KMBR are much slower than DBS as they require the computation of the MBR objective which needs a computation of the utility function for $N^2$ times. Although recent work has shown that the computation of the MBR objective can be significantly reduced \cite{cheng-vlachos-2023-faster,jinnai2024hyperparameterfree,deguchi2024centroidbased,vamvas2024lineartime}, it is not directly applicable to DMBR and KMBR. Reducing the inference time of these algorithms will be future work.

We use a simple greedy algorithm to compute the DMBR objective. More sophisticated approximation algorithms may improve the performance of DMBR \cite{pmlr-v65-feldman17b,sakaue2020guarantees}.

We consider the Monte Carlo estimate (Eq.~\ref{eq:empirical-p}) as the target quality objective for simplicity. Exploring other quality objective functions such as model-based estimate \cite{jinnai2023modelbased} is future work.


\section*{Acknowledgements}

We thank all the reviewers for their constructive comments throughout the manuscript review.
We thank Naoto Ohsaka for providing insights on approximation algorithms for a non-monotonic submodular function maximization problem.



\appendix

\section{Proof of Submodularity}
\label{sec:proof}

We show that Eq.~\eqref{eq:dmbr} with the pairwise similarity objective (Eq.~\ref{eq:pairwise}) is a submodular function maximization problem.
Let $f$ be the objective function:
\begin{align*}
    f(H) = & \sum_{\vh \in H}\left( \frac{1}{N} \sum_{\vy \in \RefH} u(\vh, \vy) \right) \\
    &- \sum_{\vh \in H} \sum_{\vh' \in H \setminus \{h\}} \frac{\lambda}{|H|} u(\vh, \vh').    
\end{align*}
Then, 
\begin{align*}
    f&(H) + f(H') \\
    &= \sum_{\vh \in H}\left( \frac{1}{N} \sum_{\vy \in \RefH} u(\vh, \vy) \right) \\
    &+ \sum_{\vh \in H'}\left( \frac{1}{N} \sum_{\vy \in \RefH} u(\vh, \vy) \right) \\
    &- \sum_{\vh \in H} \sum_{\vh' \in H \setminus \{h\}} \frac{\lambda}{|H|} u(\vh, \vh') \\
    &- \sum_{\vh \in H'} \sum_{\vh' \in H' \setminus \{h\}} \frac{\lambda}{|H'|} u(\vh, \vh') \\
    &\geq \sum_{\vh \in H \cup H'}\left( \frac{1}{N} \sum_{\vy \in \RefH} u(\vh, \vy) \right) \\
    &+ \sum_{\vh \in H \cap H'}\left( \frac{1}{N} \sum_{\vy \in \RefH} u(\vh, \vy) \right) \\
    &- \sum_{\vh \in H \cup H'} \sum_{\vh' \in H \cup H' \setminus \{h\}} \frac{\lambda}{|H \cup H'|} u(\vh, \vh') \\
    &- \sum_{\vh \in H \cap H'} \sum_{\vh' \in H \cap H' \setminus \{h\}} \frac{\lambda}{|H \cap H'|} u(\vh, \vh') \\
    &= f(H \cup H') + f(H \cap H').
\end{align*}
Thus, $f$ is a submodular function.

\section{Evaluation of the Coverage for CommonGen}
\label{sec:coverage}
The task encourages the generated sentence to contain as many input concepts as possible, so we also consider the coverage of the input concepts. The definition of coverage is the number of captured concepts divided by the number of input concepts.
To compute the captured concepts, we use a Porter stemmer \cite{porter1980algorithm} to extract the stems of the input concepts and the generated sequences and compute the number of overlaps. 
Overall, the coverage is low ($<18\%$) for all the decoding algorithms. For $k=4$, the average coverage is $15.56, 15.42, 16.03, 15.12,$ and $15.93$ for beam search, DBS-1.0, MBR, DMBR-1.0, and KMBR, respectively.
We speculate this is because the text generation model (Zephyr-7b $\beta$) is not trained to follow the instructions on the lexical constraints without few-shot prompting.

\section{Examples of Generations}
\label{sec:sample}

We show outputs generated by the decoding algorithms.
Tables \ref{tab:deen-output}, \ref{tab:ruen-output}, \ref{tab:mscoco-output}, \ref{tab:squad-output}, \ref{tab:commongen-output}, and \ref{tab:xsum-output} are the examples of the generations with various sampling algorithms for each domain evaluated in Section~\ref{sec:experiments}.
The examples show the generations of the first input source of each dataset. 

\begin{table*}
    \centering
    \begin{adjustbox}{max width=0.95\textwidth}
    \begin{tabular}{c|l}
        \toprule
        \multicolumn{2}{c}{WMT'19 De-En ($k=4$)}  \\
        \midrule\midrule
        \makecell[c]{Epsilon\\($\epsilon=0.02$)} & \makecell[l]{Beautiful Munich Woman 2018: Beautiful Munchausen in Hvar: Nine dates \\
Beauty and the Beast in Hvar, 2018: Nine Dates \\
The Best Munich Women 2018 in Hvar: Nine Dates \\
Beautiful Munich 2018 in Hvar: nine dates} \\
        \midrule\midrule
        Beam & \makecell[l]{Beautiful Munich Woman 2018: Beautiful Municipal Woman 2018 in Hvar: Nine Dates \\
Beautiful Munich 2018: Beautiful Munch 2018 in Hvar: Nine Dates \\
Beautiful Munich Woman 2018: Beautiful Munch 2018 in Hvar: Nine Dates \\
Beautiful Munich Woman 2018: Beautiful Municipal Woman 2018 in Hvar: Nine dates} \\
        \midrule
        DBS-0.5 & \makecell[l]{Beautiful Munich 2018: Beautiful Munchwoman 2018 in Hvar: Nine Dates \\
Beautiful Munich Woman 2018: Beautiful Municipal Woman 2018 in Hvar: Nine Dates \\
Beautiful Munich 2018: Beautiful Munch 2018 in Hvar: Nine Dates \\
Beautiful Munich 2018: Beautiful Hvar 2018: Nine Dates} \\
        \midrule
        DBS-1.0 & \makecell[l]{Beautiful Munich 2018: Beautiful Munchwoman 2018 in Hvar: Nine Dates \\
Beautiful Munich Woman 2018: Beautiful Municipal Woman 2018 in Hvar: Nine Dates\\
Beauty in Hvar 2018: Nine Dates\\
Beautiful Munich 2018: Beautiful Hvar in 2018: Nine Dates} \\
        \midrule
        DBS-2.0 & \makecell[l]{Nice Munich 2018: Beautiful Munich 2018 in Hvar: Nine Dates \\
Beautiful Munich 2018: Beautiful Munchwoman 2018 in Hvar: Nine Dates \\
Beautiful Munich Woman 2018: Beautiful Municipal Woman 2018 in Hvar: Nine Dates \\
Beauty in Hvar 2018: Nine Dates} \\
        \midrule
        DBS-5.0 & \makecell[l]{Nice Munich 2018: Beautiful Munich 2018 in Hvar: Nine Dates \\
Beautiful Munich 2018: Beautiful Munchwoman 2018 in Hvar: Nine Dates \\
Pretty Woman 2018: Pretties in Hvar: Nine Dates \\
Beauty in Hvar 2018: Nine Dates} \\
        \midrule\midrule
        MBR & \makecell[l]{Beautiful Munich 2018: Beautiful Munch 2018 in Hvar: Nine Dates \\
Beautiful Munich 2018: Beautiful Munch 2018 in Hvar: Nine Dates \\
Beautiful Munich 2018: Beautiful Munch 2018 in Hvar: Nine Dates \\
Beautiful Munich 2018: Beautiful Munch 2018 in Hvar: Nine Dates} \\
        \midrule
        DMBR-0.5 & \makecell[l]{Beautiful Munich 2018 in Hvar: nine dates \\
Beautiful Munich 2018: Beautiful Munch 2018 in Hvar: Nine Dates \\
Pretty Little Liars 2018: Prettiest Girls in Hvar: Nine Dates \\
Beautiful Munich Woman 2018: Beautiful Munic 2018 in Hvar: Nine Dates} \\
        \midrule
        DMBR-1.0 & \makecell[l]{2018 Pretty Little Liars in Hvar: Nine Dates \\
Pretty Muenchen 2018: Pretties Muencher 2018 in Hvar: Nine Dates \\
Beauty and the City 2018: The Best of the City in 2018: Nine Dates \\
2018 Beautiful Munich: Beautiful Hvar in Hvars - nine dates} \\
        \midrule
        DMBR-2.0 & \makecell[l]{2018 Pretty Little Liars in Hvar: Nine Dates \\
Pretty Muenchen 2018: Pretties Muencher 2018 in Hvar: Nine Dates \\
Beauty and the City 2018: The Best of the City in 2018: Nine Dates \\
2018 Beautiful Munich: Beautiful Hvar in Hvars - nine dates} \\
        \midrule
        KMBR & \makecell[l]{Nice Munich 2018: Beautiful Munich 2018 in Hvar: Nine dates \\
Beautiful Munich 2018: Beautiful Munch 2018 in Hvar: Nine Dates \\
Beautiful Munich Girl in 2018 in Hvar: Nine Dates \\
Pretty Little Liars 2018: Pretties in Hvar: Nine Dates} \\
        \bottomrule
    \end{tabular}
    \end{adjustbox}
    \caption{Examples of generations on WMT'19 De-En dataset. The number of outputs $k$ is set to $4$.}
    \label{tab:deen-output}
\end{table*}

\begin{table*}
    \centering
    \begin{adjustbox}{max width=\textwidth}
    \begin{tabular}{c|l}
        \toprule
        \multicolumn{2}{c}{WMT'19 Ru-En ($k=4$)}  \\
        \midrule\midrule
        \makecell[c]{Epsilon\\($\epsilon=0.02$)} & \makecell[l]{Named number of Ukraine conscripts preparing for departure to Donbas \\
The number of recruits from Ukraine about to be sent to Donbas is revealed \\
The number of Ukraine's new recruits to be sent to Donbas is announced \\
The number of new recruits from Ukraine preparing to be sent to Donbass has been announced} \\
         \midrule\midrule
         Beam & \makecell[l]{The number of recruits from Ukraine preparing to be sent to Donbass has been announced \\
The number of recruits from Ukraine preparing to be sent to Donbas has been announced \\
The number of new recruits from Ukraine preparing to be sent to Donbass has been announced \\
The number of recruits from Ukraine who are preparing to be sent to Donbass has been announced} \\
         \midrule
         DBS-0.5 & \makecell[l]{The number of recruits from Ukraine preparing to be sent to Donbass has been announced \\
The number of new recruits from Ukraine preparing to be sent to Donbass has been announced \\
Number of recruits from Ukraine preparing to be sent to Donbass is announced \\
The number of recruits from Ukraine preparing to be sent to Donbas has been announced} \\
         \midrule
         DBS-1.0 & \makecell[l]{The number of recruits from Ukraine preparing to be sent to Donbass has been announced \\
The number of new recruits from Ukraine preparing to be sent to Donbass has been announced \\
Number of recruits from Ukraine preparing to be sent to Donbass is announced \\
The number of Ukrainian recruits being prepared for deployment to Donbas has been announced} \\
         \midrule
         DBS-2.0 & \makecell[l]{The number of recruits from Ukraine preparing to be sent to Donbass has been announced \\
Number of recruits from Ukraine preparing to be sent to Donbass is announced \\
The number of new recruits from Ukraine preparing to be sent to Donbass has been announced \\
Named the number of recruits from Ukraine preparing for deployment in the Donbass region} \\
         \midrule
         DBS-5.0 & \makecell[l]{The number of recruits from Ukraine preparing to be sent to Donbass has been announced \\
Number of recruits from Ukraine preparing to be sent to Donbass is announced \\
Named the number of recruits from Ukraine preparing for deployment to Donbass \\
Announces Number of Ukrainian Recruits Ready for Deployment to Donbass} \\
         \midrule\midrule
         MBR & \makecell[l]{The number of recruits from Ukraine preparing to be sent to Donbass is announced \\
The number of recruits from Ukraine preparing to be sent to Donbas is announced \\
The number of recruits from Ukraine preparing to be sent to Donbass was announced. \\
The number of recruits from Ukraine preparing to be sent to Donbas has been announced} \\
         \midrule
         DMBR-0.5 & \makecell[l]{Number of new recruits from Ukraine being prepared for deployment to Donbass \\
The number of Ukrainian recruits preparing to be sent to Donbas has been announced \\
The number of Ukraine recruits to be sent to Donbass was announced \\
The number of recruits from Ukraine who are preparing to go to Donbass is announced} \\
         \midrule
         DMBR-1.0 & \makecell[l]{The number of new recruits being sent from Ukraine to Donbas has been named \\
Number of Ukrainian recruits in preparation for deployment to the Donbass \\
Announces the Number of Recruiters from Ukraine to Be Prepared for Designating in Donbas \\
The number of Ukraine's recruits preparing to be sent to Donetsk has been announced} \\
         \midrule
         DMBR-2.0 & \makecell[l]{The number of new recruits being sent from Ukraine to Donbas has been named \\
Number of Ukrainian recruits in preparation for deployment to the Donbass \\
Announces the Number of Recruiters from Ukraine to Be Prepared for Designating in Donbas \\
The number of Ukraine's recruits preparing to be sent to Donetsk has been announced} \\
         \midrule
         KMBR & \makecell[l]{The number of Ukrainian recruits preparing to be sent to Donbass has been announced \\
The number of recruits from Ukraine preparing to be sent to Donbass is announced \\
The number of new recruits from Ukraine preparing to be sent to Donbass has been announced \\
The number of recruits from Ukraine preparing to be sent to Donbass} \\
        \bottomrule
    \end{tabular}
    \end{adjustbox}
    \caption{Examples of generations on WMT'19 Ru-En dataset. The number of outputs $k$ is set to $4$.}
    \label{tab:ruen-output}
\end{table*}

\begin{table*}
    \centering
    \begin{adjustbox}{max width=0.7\textwidth}
    \begin{tabular}{c|l}
        \toprule
        \multicolumn{2}{c}{MS COCO ($k=4$)}  \\
        \midrule\midrule
        \makecell[c]{Epsilon\\($\epsilon=0.02$)} & \makecell[l]{a person riding a motorcycle down a dirt road \\
a man riding on his motorcycle on the dirt road \\
a man on a motorcycle sitting on the top of a dirt path \\
a person is sitting on the back of a motorcycle} \\
        \midrule \midrule
        Beam & \makecell[l]{a man riding a motorcycle on a dirt road \\
a man on a motorcycle on a dirt road \\
a man is riding a motorcycle on a dirt road \\
a man is riding a motorcycle down a dirt road} \\
        \midrule
        DBS-0.5 & \makecell[l]{a man riding a motorcycle on a dirt road \\
a man on a dirt bike on a dirt road \\
a man is riding a motorcycle down a dirt road \\
a man riding a motorcycle on a dirt road} \\
        \midrule
        DBS-1.0 & \makecell[l]{a man riding a motorcycle on a dirt road \\
the man is riding his motorcycle down a dirt road \\
a man on a dirt bike on a dirt road \\
a person is riding a motorcycle down a dirt road} \\
        \midrule
        DBS-2.0 & \makecell[l]{a man riding a motorcycle on a dirt road \\
the man is sitting on the back of the motorcycle \\
man riding a motorcycle on a dirt road \\
a person on a dirt bike on a dirt road} \\
        \midrule
        DBS-5.0 & \makecell[l]{a man riding a motorcycle on a dirt road \\
the man is sitting on the back of the motorcycle \\
man riding a motorcycle on a dirt road \\
there is someone on the back of a dirt bike on a dirt road} \\
        \midrule\midrule
        MBR & \makecell[l]{a man sitting on a motorcycle on a dirt road \\
a man riding a motorcycle down a dirt road \\
a man riding a motorcycle on a dirt path \\
a man on a dirt motorcycle riding on a mountain road} \\
        \midrule
        DMBR-0.5 & \makecell[l]{a person on a dirt bike sits on the back of a dirt road \\
a man is riding a motor bike down a dirt path \\
a man on a motorcycle near mountains \\
a man riding on the back of a motorcycle on a dirt road} \\
        \midrule
        DMBR-1.0 & \makecell[l]{a person on a dirt bike is riding down a dirt road \\
a motorcyclist sitting on the back of his motorcycle \\
a man on a motorcycle near mountains \\
a man is riding down a dirt road with a mountain in the distance} \\
        \midrule
        DMBR-2.0 & \makecell[l]{a person on a dirt bike is riding down a dirt road \\
a motorcyclist sitting on the back of his motorcycle \\
a man on a motorcycle near mountains \\
a man is riding down a dirt road with a mountain in the distance} \\
        \midrule
        KMBR & \makecell[l]{a man sitting on a motorcycle on a dirt road \\
a man riding a dirt bike near the mountains \\
a man riding a motorcycle down a dirt road \\
a man sits on the back of a dirt bike on a dirt road} \\
        \bottomrule
    \end{tabular}
    \end{adjustbox}
    \caption{Examples of generations on MS COCO dataset. The number of outputs $k$ is set to $4$.}
    \label{tab:mscoco-output}
\end{table*}

\begin{table*}
    \centering
    \begin{adjustbox}{width=1\textwidth}
    \begin{tabular}{c|l}
        \toprule
        \multicolumn{2}{c}{SQuADv2 ($k=4$)}  \\
        \midrule\midrule
        \makecell[c]{Epsilon\\($\epsilon=0.01$)} & \makecell[l]{What group of people gave their name to the region of Normandy in France? (Short answer: Normans) \\
What group of people in the paragraph are commonly associated with the region of Normandy in France? \\
What group of people in the paragraph are referred to by the term "Normans"? \\
What nationality did the Normals originate from in the tenth and eleventh centuries?} \\
        \midrule\midrule
        Beam & \makecell[l]{What people are referred to as Normans in the paragraph? \\
What people are referred to as "Normans" in the paragraph? \\
What people are referred to as "Normans" in the paragraph and where did they originate from? \\
What people are referred to as Normans in the given paragraph? (Note: The answer is "Normans" itself.)} \\
        \midrule
        DBS-0.5 & \makecell[l]{What people are referred to as Normans in the paragraph? \\
What group of people in the paragraph are commonly referred to as "Normans"? \\
What people gave their name to Normancy in the tenth and eleventh centuries? \\
What people in the paragraph are known for giving their nameto a region and being descended from Norse raiders?} \\
        \midrule
        DBS-1.0 & \makecell[l]{What group of people in the paragraph are commonly referred to as "Normans"? \\
What people are referred to as Normans in the paragraph? \\
What people gave their name to Normancy in the tenth and eleventh centuries? \\
What ethnic group gave their name to Normancy in the tenth and eleventh centuries?} \\
        \midrule
        DBS-2.0 & \makecell[l]{What group of people in the paragraph are commonly referred to as "Normans"? \\
What people gave their name to Normancy in the tenth and eleventh centuries? \\
Who gave their name to Normancy in the paragraph provided? \\
What ethnic group gave their name to Normancy in the tenth and eleventh centuries?} \\
        \midrule
        DBS-5.0 & \makecell[l]{What people gave their name to Normancy in the tenth and eleventh centuries? \\
Who gave their name to Normancy in the paragraph provided? \\
Question: What people in the paragraph are known as Normans, and where did they originate from? (Answer: Normans originated from NorSE raiders in Denmark, Norway, and \\
What group of people gave their name to Normancy in the paragraph provided? (Answer: Normans)} \\
        \midrule\midrule
        MBR & \makecell[l]{What ethnic group gave their name to Normancy in the Middle Ages? \\
What group of people in the paragraph gave their Name to Normancy in France? \\
What ethnic group in the Middle Ages gave their name to the region of Normandy in France? \\
What ethnic group gave their name to the region of Normandy in France?} \\
        \midrule
        DMBR-0.5 & \makecell[l]{What people gave their name to Normancy in the tenth and eleventh centuries? \\
What ethnic group gave their name to Normany? (Answer: Normans) \\
What people were the Normals named after in the region of Normandy in France? \\
What group of people in the paragraph are known as the Normands?} \\
        \midrule
        DMBR-1.0 & \makecell[l]{What people gave their name to Normamy in the10t and ith centuries? \\
What group of people in the paragraph are known as Normans? (Note: The answer is simply "Normans" as the question asks to answer with a single word.) \\
What people from Denmark-, Icelander- and Norway-descent were known as in France in the tenth and eleventh centuries? \\
What ethnically distinct group gave their name to the region of Normandy in France and how did they come to do so? (2 parts)} \\
        \midrule
        DMBR-2.0 & \makecell[l]{What nationality were the Normals originally before giving their name to Normandy? \\
What people gave their name to Normamy in the10t and ith centuries? \\
What group of people in the paragraph are known as Normans? (Note: The answer is simply "Normans" as the question asks to answer with a single word.) \\
What people from Denmark-, Icelander- and Norway-descent were known as in France in the tenth and eleventh centuries?} \\
        \midrule
        KMBR & \makecell[l]{What people gave their name to Normancy in the tenth and eleventh centuries? \\
What ethnic group gave their name to the region of Normandy in France? \\
What nationality did the Normals originate from in the tenth and eleventh centuries? \\
What group of people in the paragraph are known as Normans?} \\
        \bottomrule
    \end{tabular}
    \end{adjustbox}
    \caption{Examples of generations on SQuADv2. The number of outputs $k$ is set to $4$.}
    \label{tab:squad-output}
\end{table*}

\begin{table*}
    \centering
    \begin{adjustbox}{width=1\textwidth}
    \begin{tabular}{c|l}
        \toprule
        \multicolumn{2}{c}{CommonGen ($k=4$)}  \\
        \midrule\midrule
        \makecell[c]{Epsilon\\($\epsilon=0.01$)} & \makecell[l]{The standing crop in the fields catches the eye with its verdant hue. \\
The standing crops in the agricultural field appear majestic as the observer looks on. \\
The standing crops in the scenic field are currently being observed by onlookers. \\
In the field, a statue appears to be standing despite not actually looking like it's standing due to its stationary position.} \\
        \midrule\midrule
        Beam & \makecell[l]{In the field, a look stands out. \\
Standing in the open field, one's gaze wanders afar. \\
Standing in the open field, one's gaze is fixed on the horizon. \\
In the countryside, a stationary telescope gazes out at the landscape.} \\
        \midrule
        DBS-0.5 & \makecell[l]{In the countryside, a stationary telescope gazes out at the landscape. \\
In the field, a look stands out. \\
The farmer's eyes scanned the horizon, waiting for his crops to come to their feet. \\
In the countryside, a stationary view meets the eye.} \\
        \midrule
        DBS-1.0 & \makecell[l]{In the countryside, a stationary telescope gazes out at the landscape. \\
Standing in the open field, one's gaze wanders afar. \\
The farmer's eyes scanned the horizon, waiting for his crops to come to their feet. \\
In the field, a look stands out.} \\
        \midrule
        DBS-2.0 & \makecell[l]{In the countryside, a stationary telescope gazes out at the landscape. \\
Standing in the open field, one's gaze wanders afar. \\
The farmer's eyes scanned the horizon, waiting for his crops to come to their feet. \\
"In the countryside, a farmer stops to survey the landscape with a keen gaze." (Field, look, stand)} \\
        \midrule
        DBS-5.0 & \makecell[l]{In the countryside, a stationary telescope gazes out at the landscape. \\
Standing in the open field, one's gaze wanders afar. \\
"In the countryside, a farmer stops to survey the landscape with a keen gaze." (Field, look, stand)\textbackslash{}n Alternatively: "In a rural area, an agricultural worker paus \\
The farmer's eyes scanned the horizon, waiting for his crops to come to their feet.\textbackslash{}nSentence: In the farmer' s field, the look stood. \textbackslash{}nExplanation} \\
        \midrule\midrule
        MBR & \makecell[l]{The standing figure in the open field catches the eye with its striking appearance. \\
The standing figure in the open field catches the eye with its striking appearance. \\
In the field, a stationary scene catches the eye. \\
In the field, a statue appears to be standing and looking in a specific direction.} \\
        \midrule
        DMBR-0.5 & \makecell[l]{In the field, a stationary gaze is observed. \\
The standing figure in the open field catches the eye with its captivating appearance. \\
In the field, a statue appears to be standing and looking out into the distance. \\
The standing crops in the agricultural field undergo visual inspections, known as field looks.} \\
        \midrule
        DMBR-1.0 & \makecell[l]{The standing crops in the agricultural field undergo visual inspections, known as field looks. \\
The field appeared still and stationary in the standstill of the stand.' \\
In the field, a look stands out. \\
The standing figure gazes out into the open expanse of greenery in the distance.} \\
        \midrule
        DMBR-2.0 & \makecell[l]{In the field, a stationary observation is called a field look or stand. \\
The standing crops in the open field catch the eye with their vivid green hues on a sunny day. \\
The field appeared still and stationary in the standstill of the stand.' \\
In the field, a sculpture appears to be standing and gazing outwards, as if looking in a particular direction.} \\
        \midrule
        KMBR & \makecell[l]{In the field, a statue appears to be standing and looking in a specific direction. \\
The standing crops in the open field catch the eye of passersby. \\
The standing figure gazes out into the open expanse of the field. \\
In the field, a stationary view can be observed.} \\
        \bottomrule
    \end{tabular}
    \end{adjustbox}
    \caption{Examples of generations on CommonGen. The number of outputs $k$ is set to $4$.}
    \label{tab:commongen-output}
\end{table*}

\begin{table*}
    \centering
    \begin{adjustbox}{width=\textwidth}
    \begin{tabular}{c|l}
        \toprule
        \multicolumn{2}{c}{XSum ($k=4$)}  \\
        \midrule\midrule
        \makecell[c]{Epsilon\\($\epsilon=0.02$)} & \makecell[l]{A charity which helps prison leavers is claiming there is a "desperate need" for more housing for those released after prison. \\
There is a "desperate need" for housing for former inmates who are left homeless after leaving prison, a charity has said. \\
The number of referrals to a homelessness charity by people released from prison in Wales is at an all-time high, it has been claimed. \\
A charity has called for more homes for men who are left homeless after leaving prison in Wales.} \\
        \midrule\midrule
        Beam & \makecell[l]{There is a "desperate need" for housing for men and women released from prison in Wales, a charity has said. \\
There is a "desperate need" for housing for men and women released from prison, a charity has said. \\
There is a "desperate need" for housing for former prisoners in Wales, a charity has said. \\
There is a "desperate need" for housing for men and women released from prison in Wales, a charity has claimed.} \\
        \midrule
        DBS-0.5 & \makecell[l]{A charity which helps former prisoners find housing has said there is a "desperate need" for more one-bedroom flats. \\
More needs to be done to help former prisoners find accommodation after they leave prison, a charity has said. \\
The number of people referred to a charity for help finding housing after leaving prison has more than doubled in the past year, it has been revealed. \\
The number of people being referred to a charity for help finding housing after leaving prison has risen by more than a third in the past year, figures show.} \\
        \midrule
        DBS-1.0 & \makecell[l]{More needs to be done to help former prisoners find accommodation after they leave prison, a charity has said. \\
A charity which helps former prisoners find housing has said there is a "desperate need" for accommodation for them after they leave prison. \\
The number of people being referred to a charity for help finding housing after leaving prison has risen by more than a third in the past year, figures show. \\
The number of homeless people in Wales has risen by more than 50\% in the past year, a charity has said.} \\
        \midrule
        DBS-2.0 & \makecell[l]{There is a "desperate need" for more housing for men and women who leave prison, a charity has said. \\
More needs to be done to help former prisoners find accommodation after they leave prison, a charity has said. \\
A charity which helps former prisoners find housing has said there is a "desperate need" for accommodation for them after they leave prison. \\
The number of people being referred to a charity for help finding housing after leaving prison has risen by more than a third in the past year, figures show.} \\
        \midrule
        DBS-5.0 & \makecell[l]{There is a "desperate need" for more housing for men and women who leave prison, a charity has said. \\
More needs to be done to help former prisoners find accommodation after they leave prison, a charity has said. \\
A charity which helps former prisoners find housing has said there is a "desperate need" for accommodation for them after they leave prison. \\
The number of people being referred to a charity for help finding housing after leaving prison has risen by more than a third in the past year, figures show.} \\
        \midrule\midrule
        MBR & \makecell[l]{The number of former prisoners needing help to find housing is on the rise in Wales, a charity has said. \\
There is a "desperate" need for better housing for people leaving prison, a charity has said. \\
The number of homeless people in Wales is rising because of a "desperate" need for help to find accommodation after they leave prison, a charity has said. \\
The number of people needing help to find accommodation after being released from prison in Wales is rising, a charity has said.} \\
        \midrule
        DMBR-0.5 & \makecell[l]{A charity which helps prisoners find housing has said there is a "desperate need" for more support. \\
The number of former prisoners in Wales being referred for help to find housing is at its highest level for seven years, new figures have shown. \\
The number of people being put on the streets by homelessness services in Wales after being released from prison has doubled this year, according to a charity. \\
More affordable homes should be built for people leaving prison, a charity has said.} \\
        \midrule
        DMBR-1.0 & \makecell[l]{Welsh jailing for homeless people could save money for the public purse, according to a homeless charity. \\
The number of people being referred to Wales' prisons because of problems getting housing has risen by 30\%. \\
There is a "desperate" need for better support for prison leavers, with more than one a day being referred to an accommodation charity, a charity has said. \\
A charity which helps ex-prisoners find housing after being released has said it is "desperately" in need of more one-bedroom flats.} \\
        \midrule
        DMBR-2.0 & \makecell[l]{More one-bedroom flats could be built in Wales to help ease a "desperate" need for accommodation for former prison leavers, a charity has said. \\
Welsh jailing for homeless people could save money for the public purse, according to a homeless charity. \\
The number of people being referred to Wales' prisons because of problems getting housing has risen by 30\%. \\
A charity which helps men and women leave prison says there are "desperate" problems for them after they leave.} \\
        \midrule
        KMBR & \makecell[l]{More affordable homes should be built for people leaving prison, a charity has said. \\
The number of former prisoners needing help to find housing is on the rise in Wales, a charity has said. \\
There is a "desperate" need for better housing for people leaving prison, a charity has said. \\
A charity that helps men and women find housing after leaving prison has claimed there is a "desperate need" for more flats.} \\
        \bottomrule
    \end{tabular}
    \end{adjustbox}
    \caption{Examples of generations on XSum. The number of outputs $k$ is set to $4$.}
    \label{tab:xsum-output}
\end{table*}

\clearpage

\section{Evaluation of Oversampling Strategy}
\label{sec:oversampling}
We additionally evaluate the performance of an oversampling strategy as a baseline.
Oversampling strategy generates $N (> k)$ samples and then selects $k$ samples out of the $N$, maximizing the objective in Eq.~\eqref{eq:setdiv}:
\begin{equation}
    H^* = \argmax_{H \subseteq \mathcal{Y}} \sum_{\vh \in H} P_{\mathrm{human}}(\vh | \vx) + d_{\mathrm{human}}(H).    
\end{equation}
Because $P_{\mathrm{human}}$ and $d_{\mathrm{human}}$ are inaccessible, we approximate them using a model and Eq.~\eqref{eq:pairwise}:
\begin{align}
    H^* = \argmax_{H \subseteq \mathcal{Y}}& \sum_{\vh \in H} P(\vh | \vx) - \nonumber\\
    &\sum_{\vh \in H} \sum_{\vh' \in H \setminus \{h\}} \frac{\lambda}{|H|} u(\vh, \vh').
\end{align}
The results with $N=128$ and $k=4$ are shown in Table~\ref{tab:oversampling}. Overall, we observe that it performs slightly worse than DMBR, and the hyperparameter $\lambda$ is dependent on the generation probability of the sentences which in turn depends on sequence length (Table \ref{tab:wmt}). 
The result indicates that the improvement of DMBR comes from the use of the utility function to select high-quality samples in addition to the oversampling. 

\begin{table*}
    \centering

    \adjustbox{max width=\textwidth}{
    \begin{tabular}{lccccccc}
    \toprule
     & \multicolumn{3}{c}{Quality} & \multicolumn{4}{c}{Diversity} \\
    \cmidrule(lr){2-4}\cmidrule(lr){5-8}
    Decoder & min BLEU $\uparrow$ & mean BLEU $\uparrow$ & max BLEU $\uparrow$ & Pairwise-BLEU $\downarrow$ & distinct-1 $\uparrow$ & distinct-2 $\uparrow$ & distinct-3 $\uparrow$ \\
    \midrule\midrule
    & \multicolumn{6}{c}{WMT'19 De-En ($k=4$)} \\
    \midrule
    OS-0.1 & 30.96 & 35.09 & 39.44 & 75.92 & 0.28 & 0.33 & 0.34 \\
    OS-0.3 & 30.92 & 35.09 & 39.46 & 75.81 & 0.28 & 0.33 & 0.34 \\
    OS-0.5 & 30.92 & 35.09 & 39.46 & 75.74 & 0.28 & 0.33 & 0.34 \\
    OS-1.0 & 30.85 & 35.06 & 39.46 & 75.56 & 0.28 & 0.33 & 0.34 \\
    OS-2.0 & 30.83 & 35.05 & 39.50 & 75.37 & 0.28 & 0.33 & 0.34 \\\midrule
    MBR & 31.25 & 35.17 & 39.20 & 75.14 & 0.28 & 0.33 & 0.35 \\
    DMBR-0.1 & 27.80 & 34.74 & 41.85 & 61.80 & 0.31 & 0.39 & 0.42 \\
    DMBR-0.3 & 21.21 & 32.03 & 43.51 & 40.74 & 0.37 & 0.50 & 0.55 \\
    DMBR-0.5 & 13.59 & 25.80 & 39.94 & 22.84 & 0.45 & 0.64 & 0.69 \\
    DMBR-1.0 & 10.76 & 20.59 & 33.41 & 15.54 & 0.51 & 0.71 & 0.75 \\
    DMBR-2.0 & 10.47 & 19.45 & 30.93 & 14.94 & 0.51 & 0.72 & 0.76 \\
    \midrule\midrule
    & \multicolumn{6}{c}{WMT'19 Ru-En ($k=4$)} \\
    \midrule
    OS-0.1 & 27.59 & 31.59 & 35.62 & 75.68 & 0.27 & 0.33 & 0.34 \\
    OS-0.3 & 27.58 & 31.59 & 35.65 & 75.62 & 0.27 & 0.33 & 0.34 \\
    OS-0.5 & 27.58 & 31.59 & 35.65 & 75.62 & 0.27 & 0.33 & 0.34 \\
    OS-1.0 & 27.56 & 31.60 & 35.65 & 75.51 & 0.27 & 0.33 & 0.34 \\
    OS-2.0 & 27.50 & 31.57 & 35.66 & 75.38 & 0.27 & 0.33 & 0.34 \\\midrule
    MBR & 28.08 & 32.28 & 36.44 & 75.12 & 0.27 & 0.33 & 0.35 \\
    DMBR-0.1 & 24.91 & 31.73 & 38.91 & 61.97 & 0.30 & 0.38 & 0.42 \\
    DMBR-0.3 & 19.74 & 29.60 & 40.33 & 41.30 & 0.35 & 0.50 & 0.55 \\
    DMBR-0.5 & 13.47 & 24.48 & 37.51 & 23.58 & 0.43 & 0.63 & 0.68 \\
    DMBR-1.0 & 10.49 & 19.84 & 31.51 & 16.23 & 0.48 & 0.70 & 0.75 \\
    DMBR-2.0 & 10.06 & 18.90 & 30.03 & 15.51 & 0.49 & 0.71 & 0.76 \\
    \bottomrule
    \end{tabular}
    }
    \caption{Evaluation of the quality and diversity using the oversampling strategy (OS-$\lambda$) on WMT'19 De-En and Ru-En dataset (Appendix \ref{sec:oversampling}). The size of the output $k$ is set to 4.}
    \label{tab:oversampling}
\end{table*}

\section{Additional Figures and Tables}
\label{sec:results}

\subsection{Additional Figures}
\begin{figure*}
    \centering
    \begin{subfigure}[b]{0.32\textwidth}
        \includegraphics[width=\textwidth]{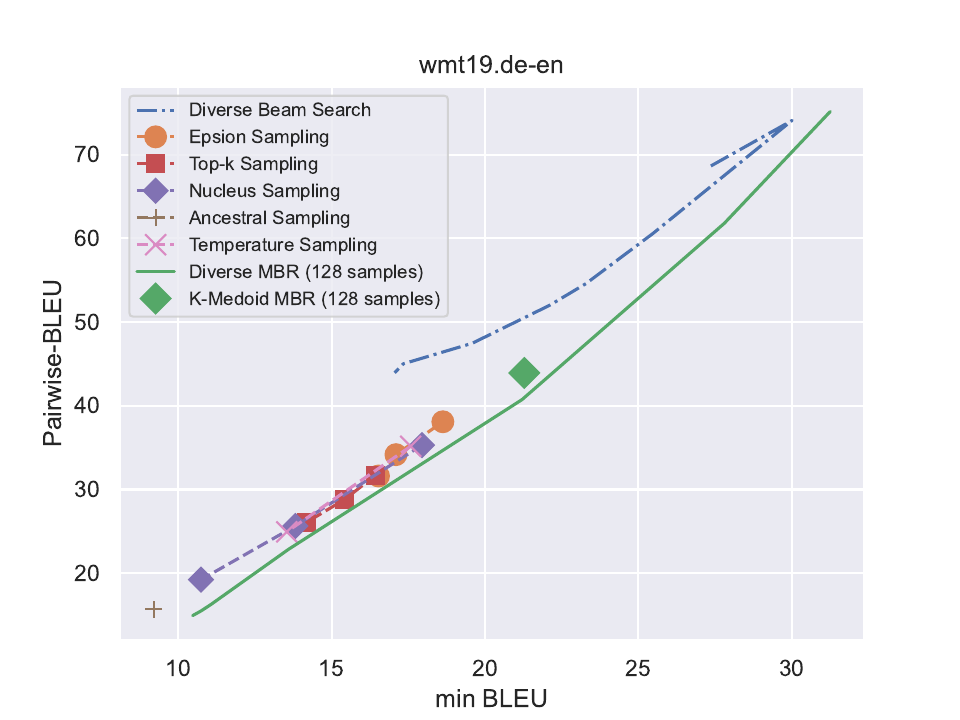}
        \caption{P-BLEU $\downarrow$ (De-En)}
    \end{subfigure}
    \begin{subfigure}[b]{0.32\textwidth}
        \includegraphics[width=\textwidth]{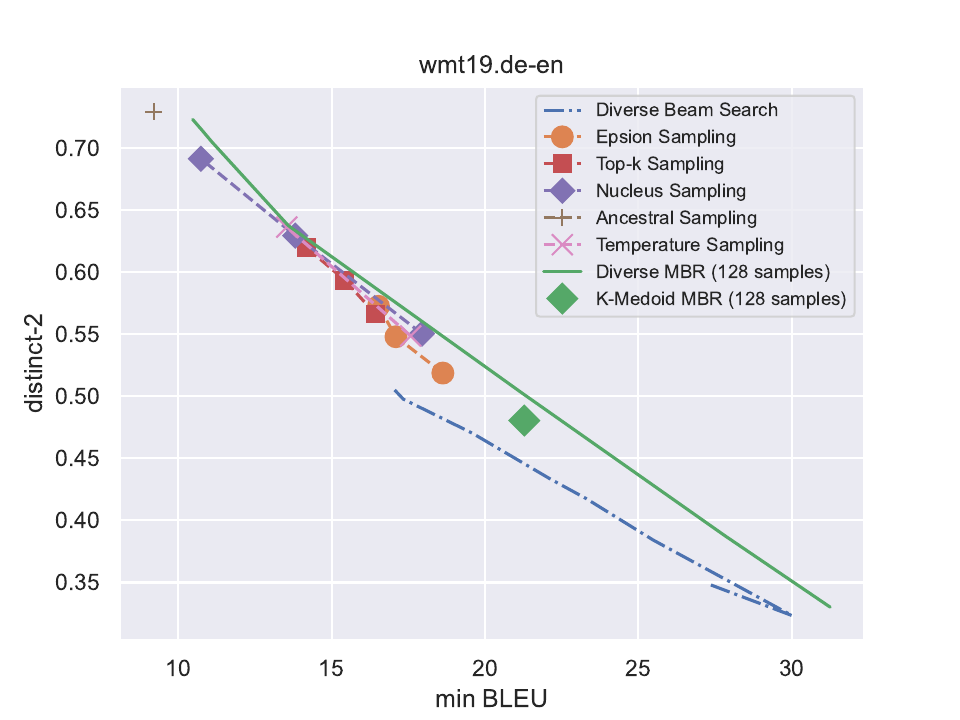}
        \caption{distinct-2 $\uparrow$ (De-En)}
    \end{subfigure}
    \begin{subfigure}[b]{0.32\textwidth}
        \includegraphics[width=\textwidth]{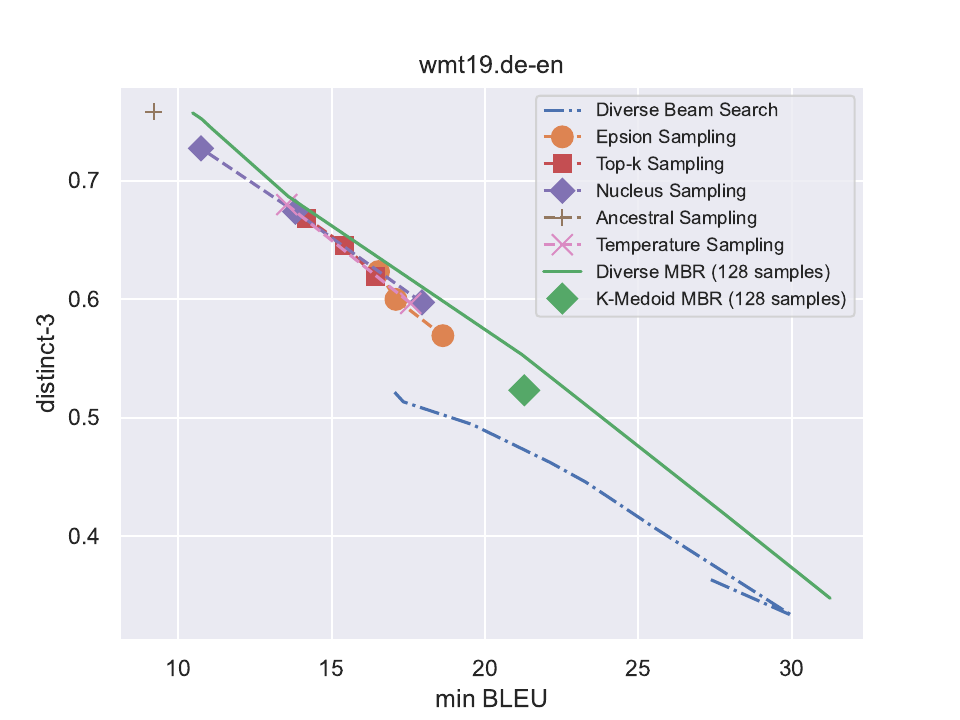}
        \caption{distinct-3 $\uparrow$ (De-En)}
    \end{subfigure}
    \begin{subfigure}[b]{0.32\textwidth}
        \includegraphics[width=\textwidth]{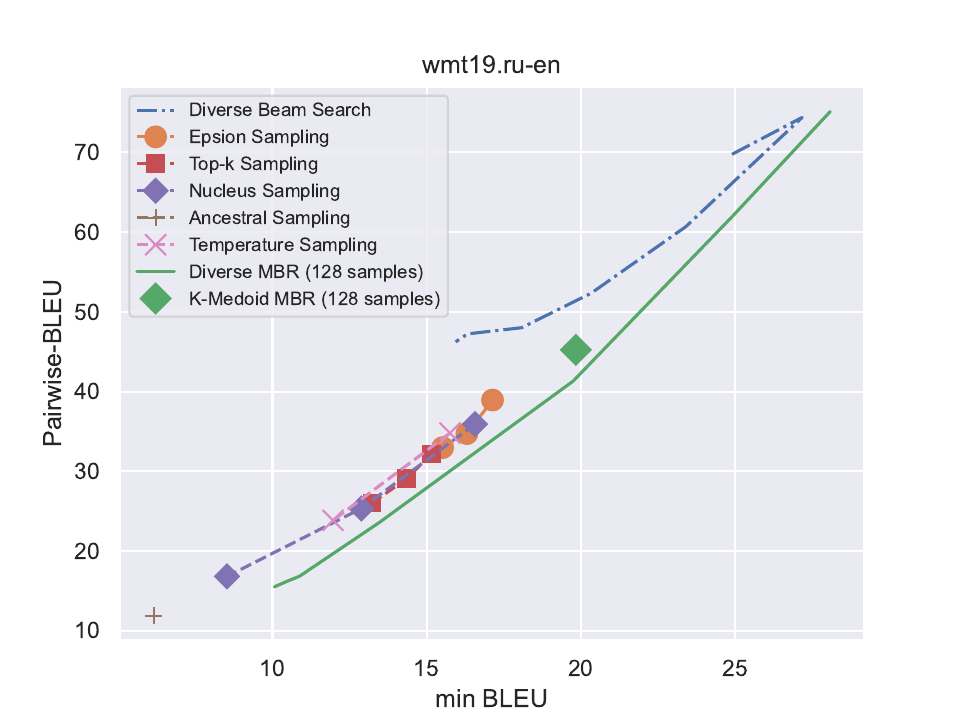}
        \caption{P-BLEU $\downarrow$ (Ru-En)}
    \end{subfigure}
    \begin{subfigure}[b]{0.32\textwidth}
        \includegraphics[width=\textwidth]{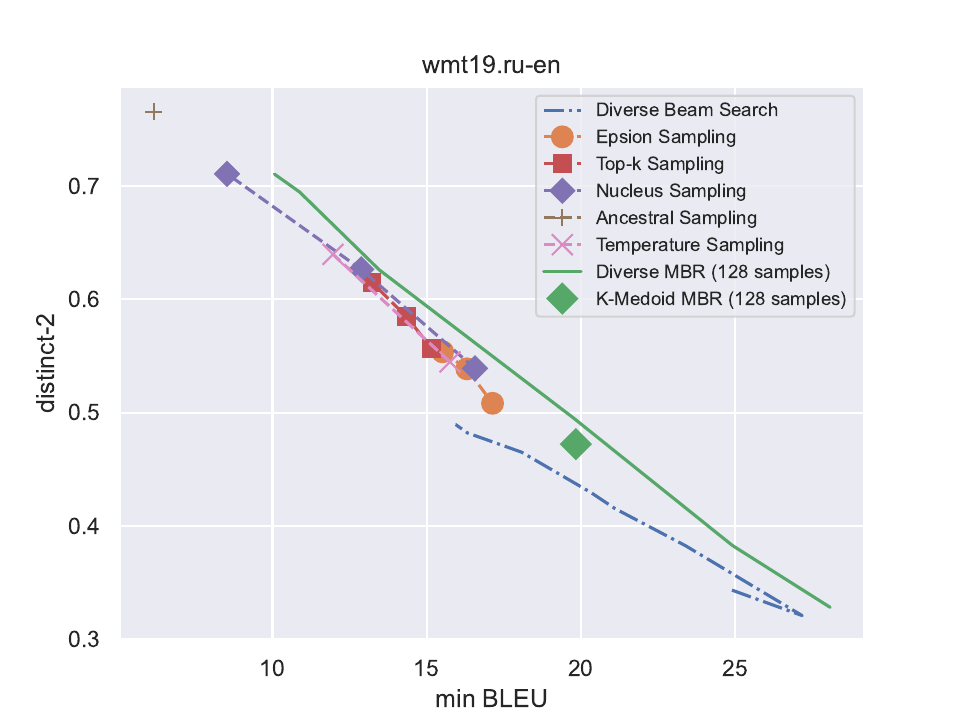}
        \caption{distinct-2 $\uparrow$ (Ru-En)}
    \end{subfigure}
    \begin{subfigure}[b]{0.32\textwidth}
        \includegraphics[width=\textwidth]{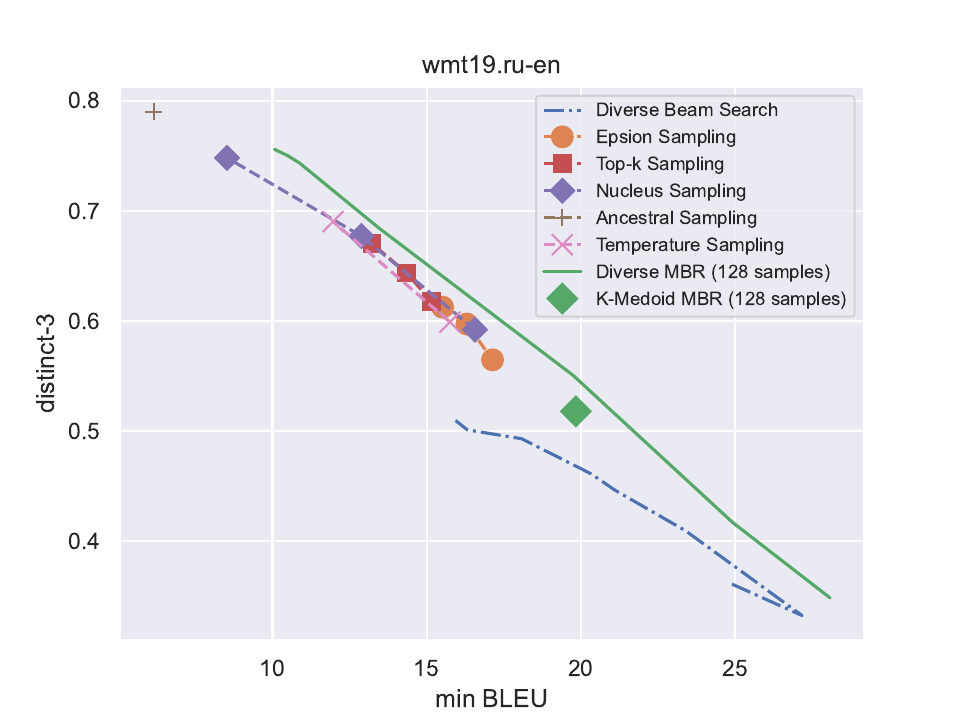}
        \caption{distinct-3 $\uparrow$ (Ru-En)}
    \end{subfigure}

    \caption{Min BLEU, P-BLEU, distinct-n on WMT'19 De-En and Ru-En. The size of the output $k$ is set to 4.}  
    \label{fig:min}
\end{figure*}

The evaluation of the diversity as a function of min BLEU over the outputs on WMT'19 datasets is present in Figure~\ref{fig:min}. DMBR achieves a better trade-off than DBS and sampling algorithms with the same min BLEU score.

\begin{figure*}
    \centering
    \begin{subfigure}[b]{0.32\textwidth}
        \includegraphics[width=\textwidth]{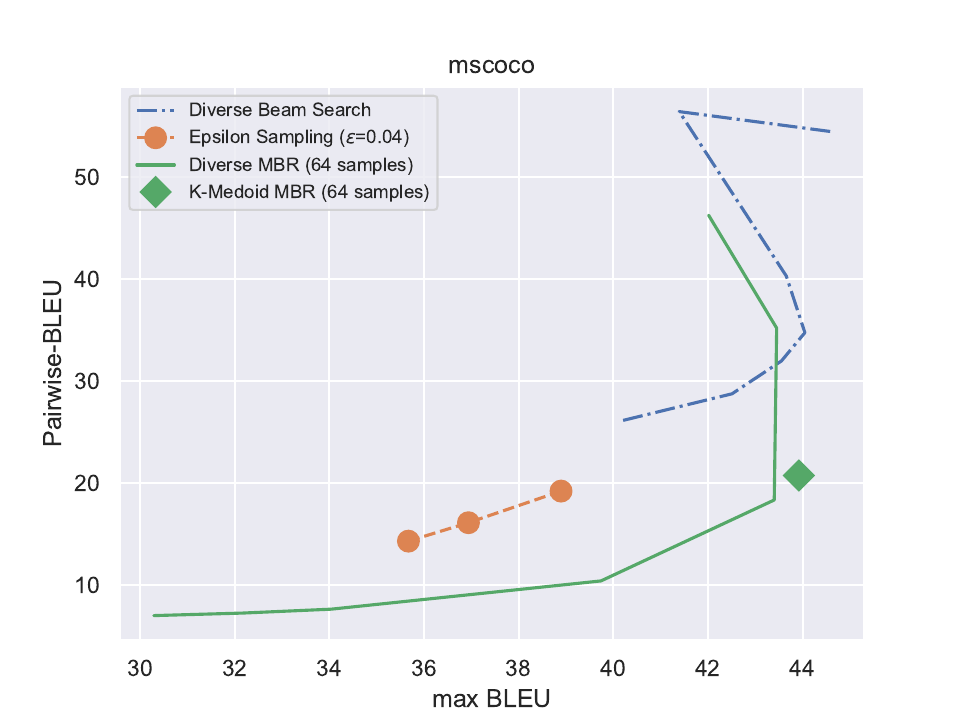}
        \caption{P-BLEU $\downarrow$ (MS COCO)}
    \end{subfigure}
    \begin{subfigure}[b]{0.32\textwidth}
        \includegraphics[width=\textwidth]{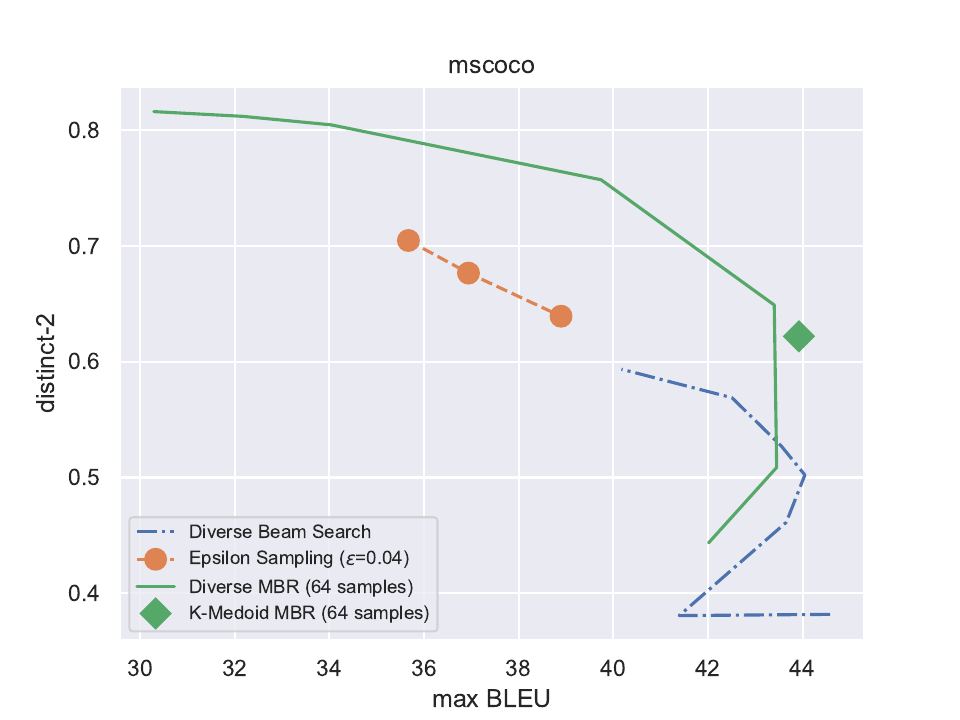}
        \caption{distinct-2 $\uparrow$ (MS COCO)}
    \end{subfigure}
    \begin{subfigure}[b]{0.32\textwidth}
        \includegraphics[width=\textwidth]{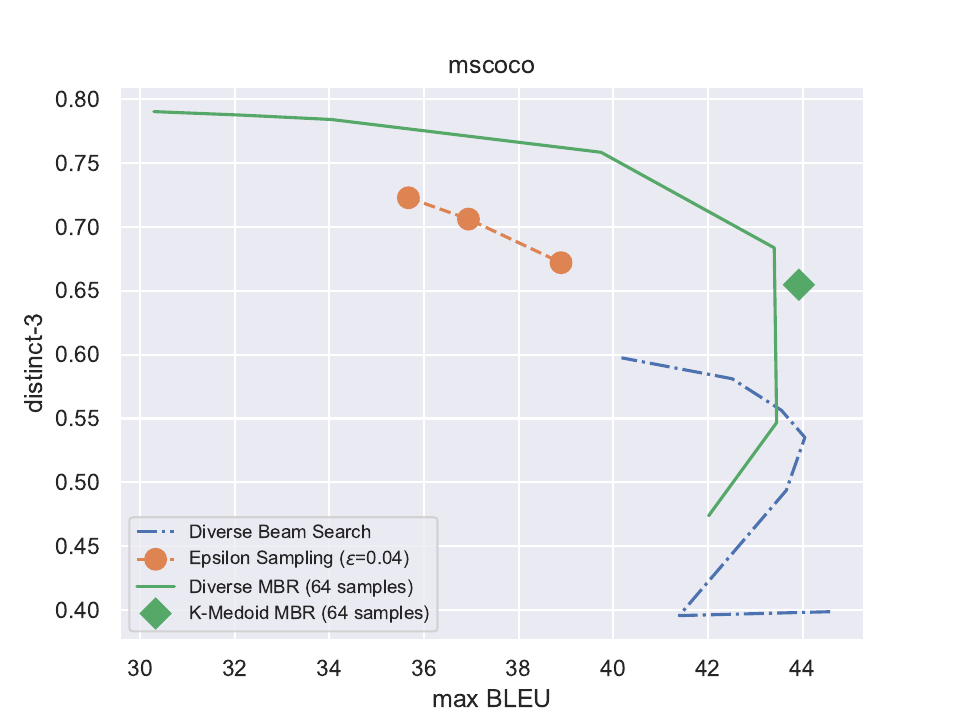}
        \caption{distinct-3 $\uparrow$ (MS COCO)}
    \end{subfigure}

    \begin{subfigure}[b]{0.32\textwidth}
        \includegraphics[width=\textwidth]{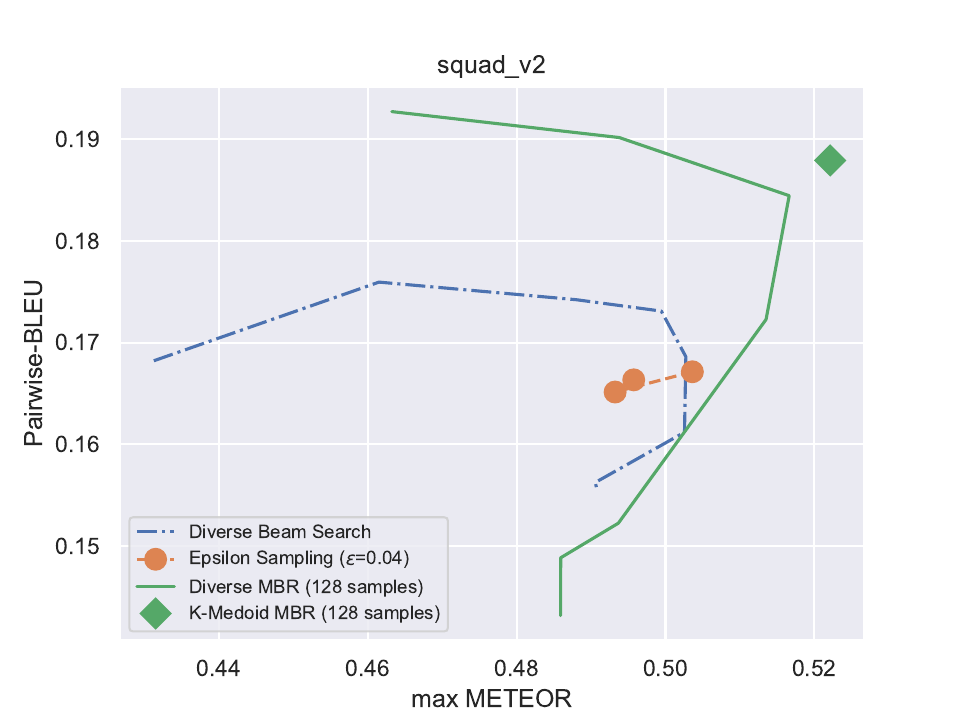}
        \caption{P-BLEU $\downarrow$ (SQuADv2)}
    \end{subfigure}
    \begin{subfigure}[b]{0.32\textwidth}
        \includegraphics[width=\textwidth]{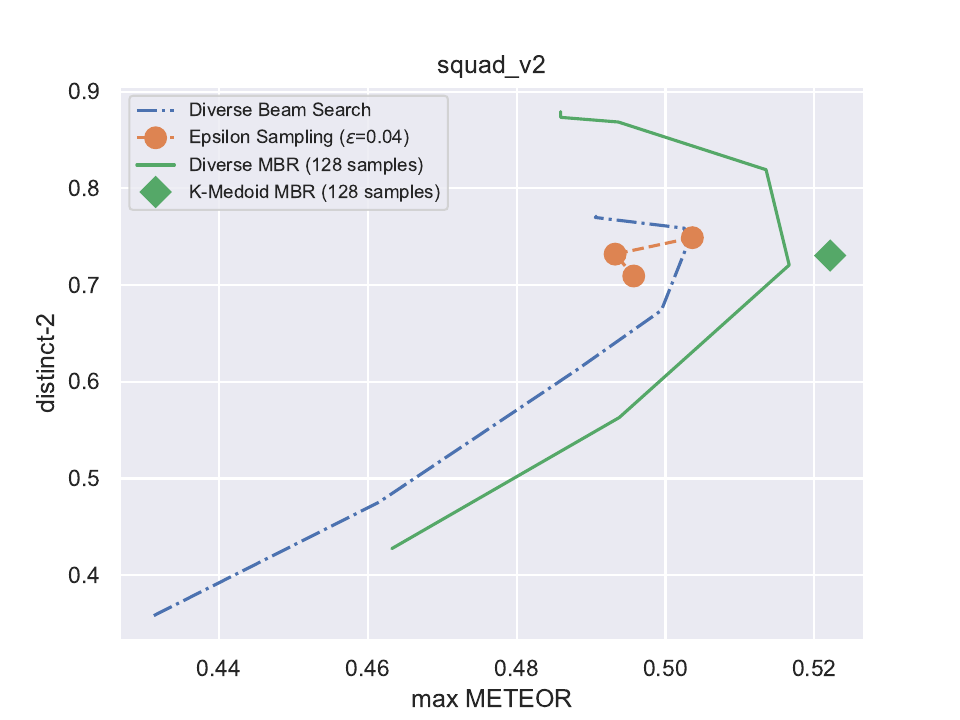}
        \caption{distinct-2 $\uparrow$ (SQuADv2)}
    \end{subfigure}
    \begin{subfigure}[b]{0.32\textwidth}
        \includegraphics[width=\textwidth]{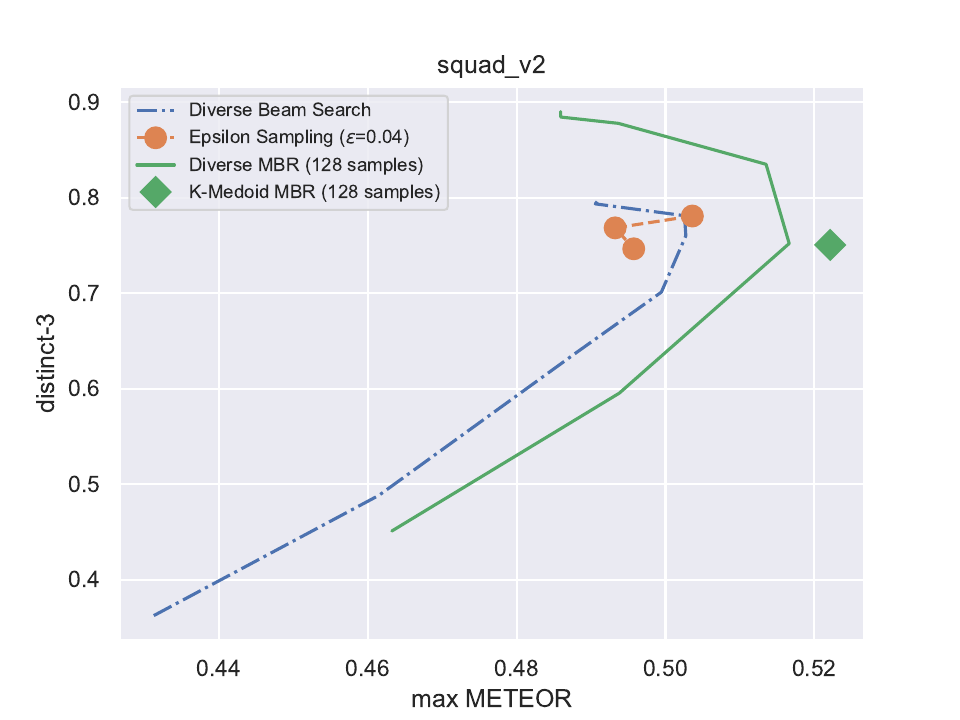}
        \caption{distinct-3 $\uparrow$ (SQuADv2)}
    \end{subfigure}

    \begin{subfigure}[b]{0.32\textwidth}
        \includegraphics[width=\textwidth]{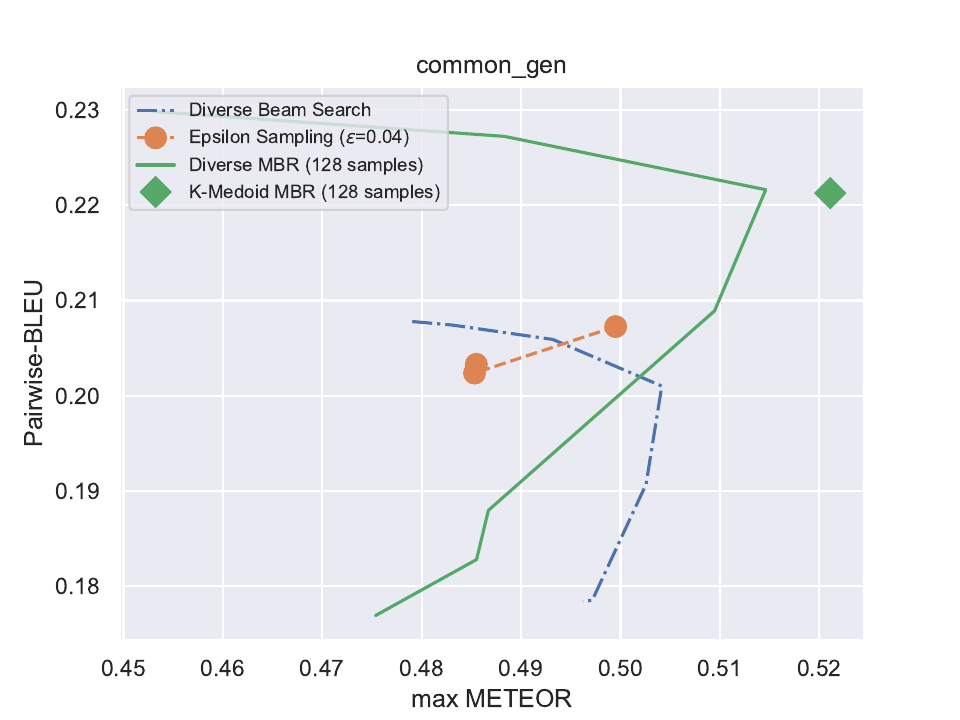}
        \caption{P-BLEU $\downarrow$ (CommonGen)}
    \end{subfigure}
    \begin{subfigure}[b]{0.32\textwidth}
        \includegraphics[width=\textwidth]{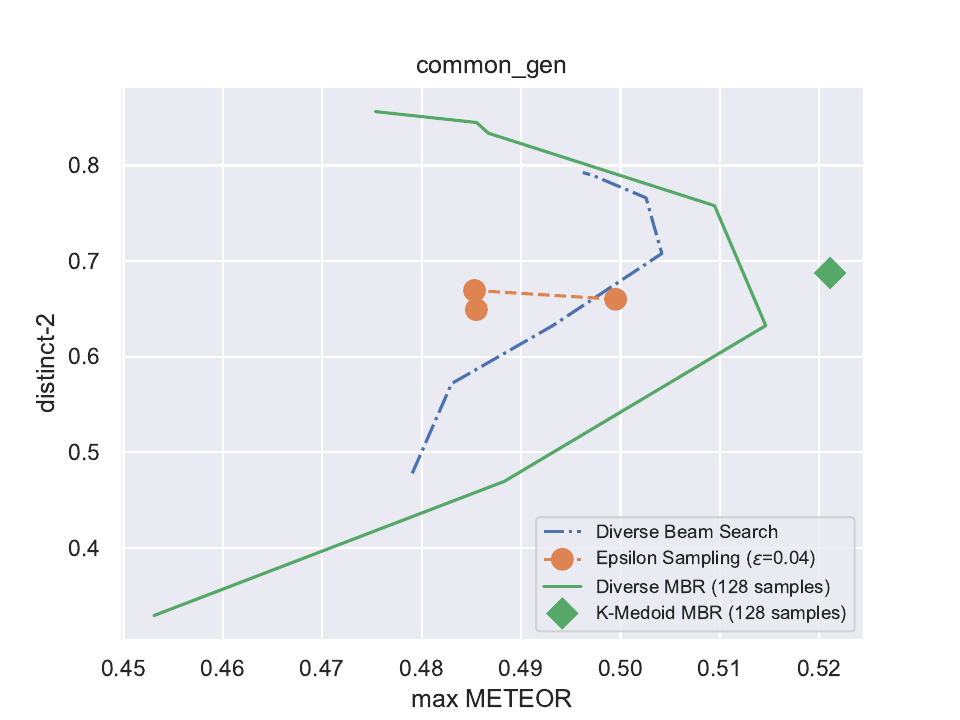}
        \caption{distinct-2 $\uparrow$ (CommonGen) }
    \end{subfigure}
    \begin{subfigure}[b]{0.32\textwidth}
        \includegraphics[width=\textwidth]{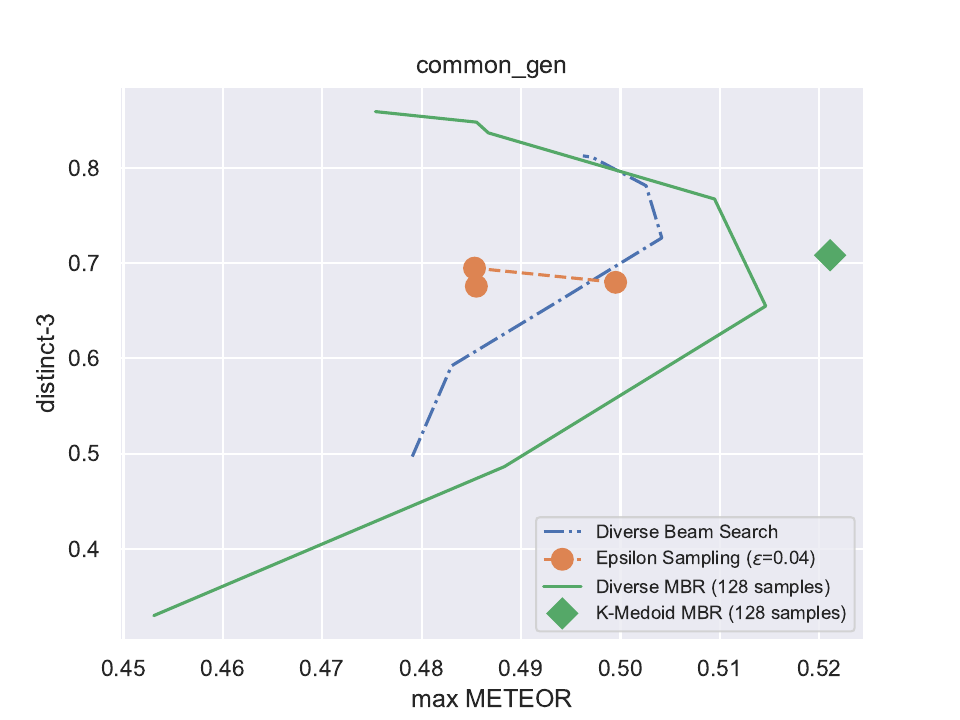}
        \caption{distinct-3 $\uparrow$ (CommonGen)}
    \end{subfigure}

    \begin{subfigure}[b]{0.32\textwidth}
        \includegraphics[width=\textwidth]{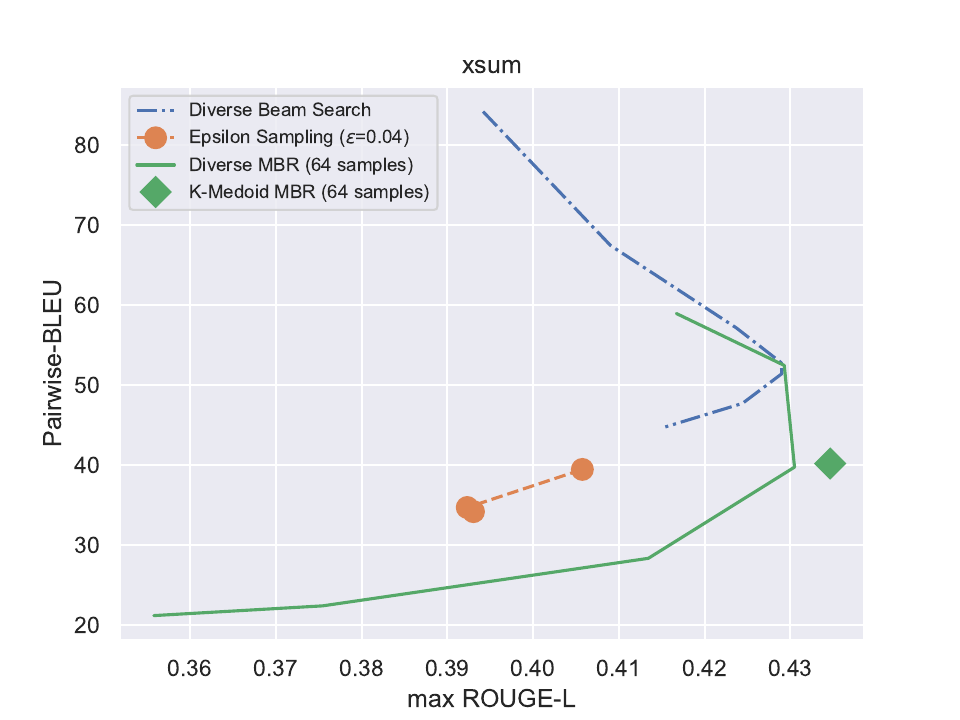}
        \caption{P-BLEU $\downarrow$ (XSum)}
    \end{subfigure}
    \begin{subfigure}[b]{0.32\textwidth}
        \includegraphics[width=\textwidth]{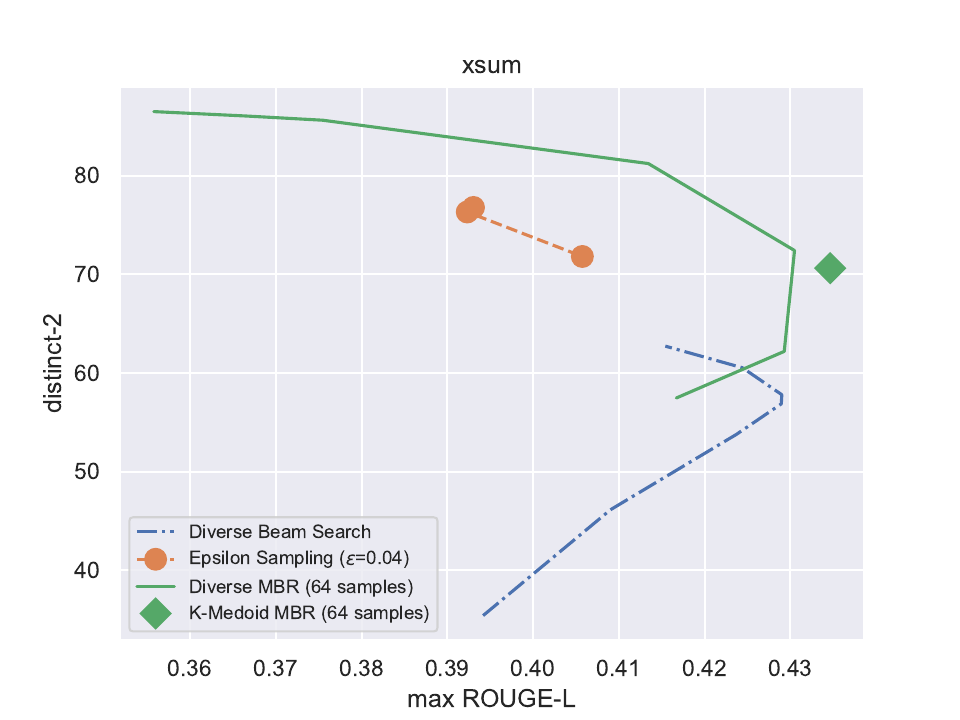}
        \caption{distinct-2 $\uparrow$ (XSum) }
    \end{subfigure}
    \begin{subfigure}[b]{0.32\textwidth}
        \includegraphics[width=\textwidth]{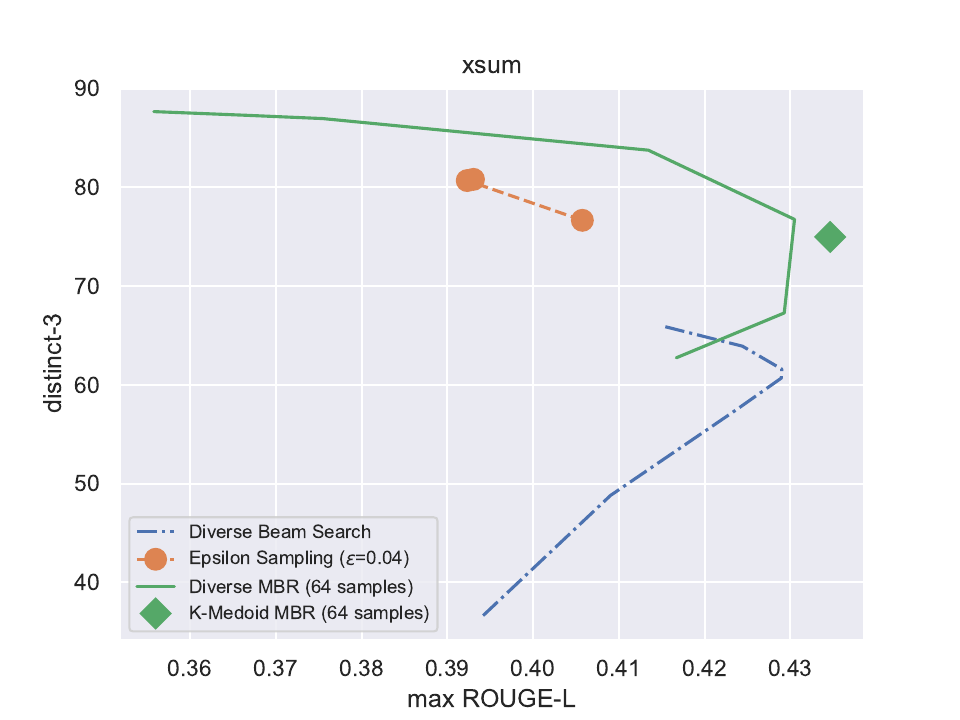}
        \caption{distinct-3 $\uparrow$ (XSum)}
    \end{subfigure}

    \caption{Evaluation of the P-BLEU and distinct-2, 3 as a function of max BLEU (MS COCO), METEOR (SQuADv2, CommonGen), and ROUGE-L (XSum). The size of the output $k$ is set to 4.}
    \label{fig:othersmax}
\end{figure*}

\begin{figure*}
    \centering
    \begin{subfigure}[b]{0.32\textwidth}
        \includegraphics[width=\textwidth]{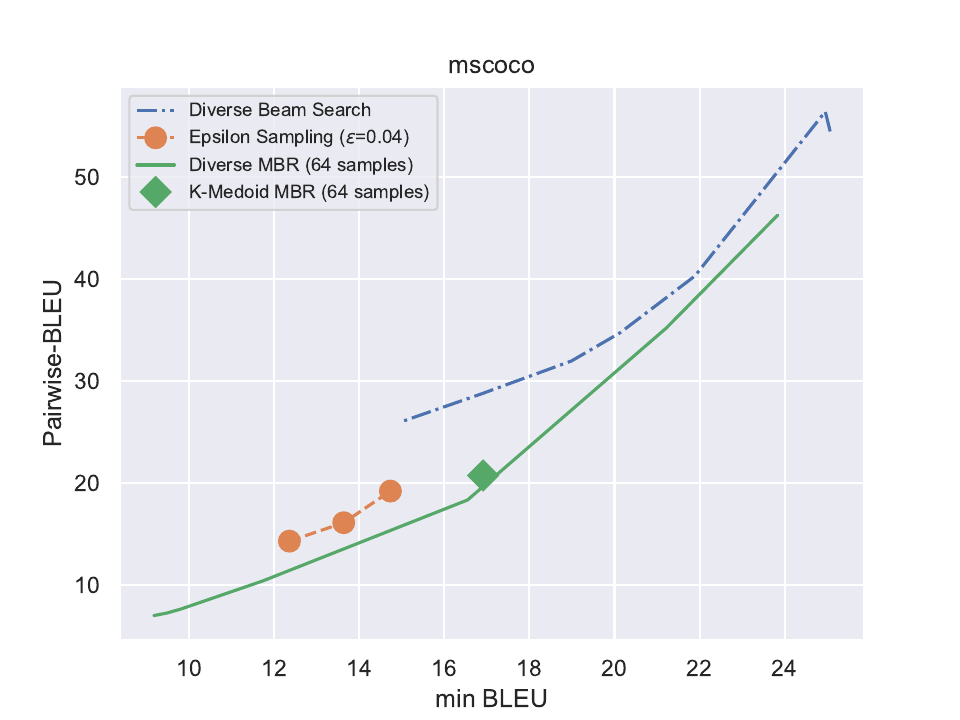}
        \caption{P-BLEU $\downarrow$ (MS COCO)}
    \end{subfigure}
    \begin{subfigure}[b]{0.32\textwidth}
        \includegraphics[width=\textwidth]{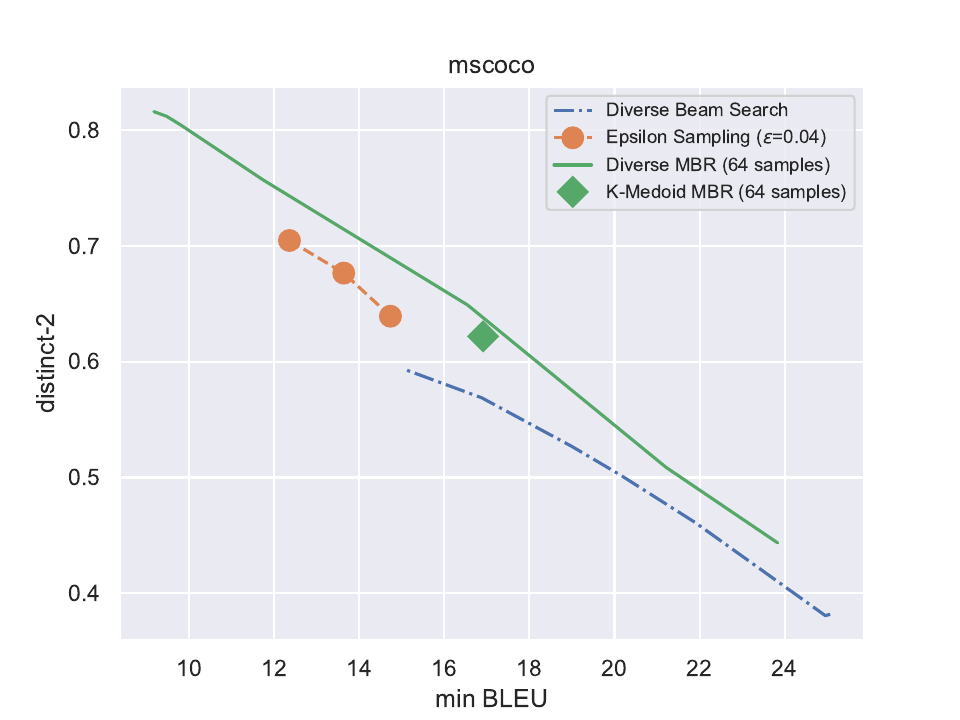}
        \caption{distinct-2 $\uparrow$ (MS COCO)}
    \end{subfigure}
    \begin{subfigure}[b]{0.32\textwidth}
        \includegraphics[width=\textwidth]{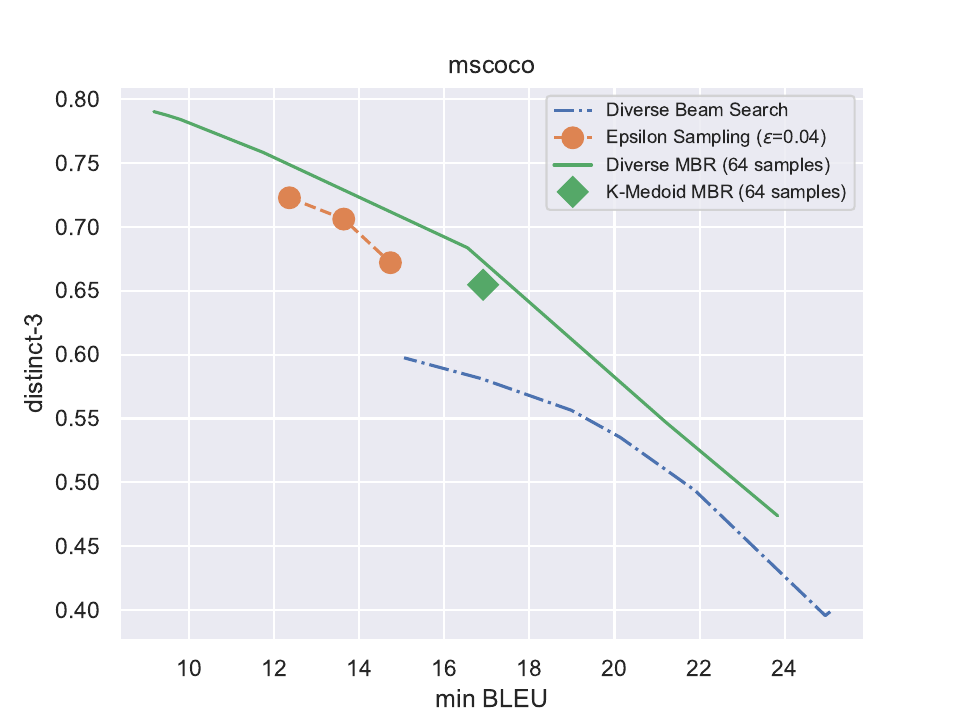}
        \caption{distinct-3 $\uparrow$ (MS COCO)}
    \end{subfigure}

    \begin{subfigure}[b]{0.32\textwidth}
        \includegraphics[width=\textwidth]{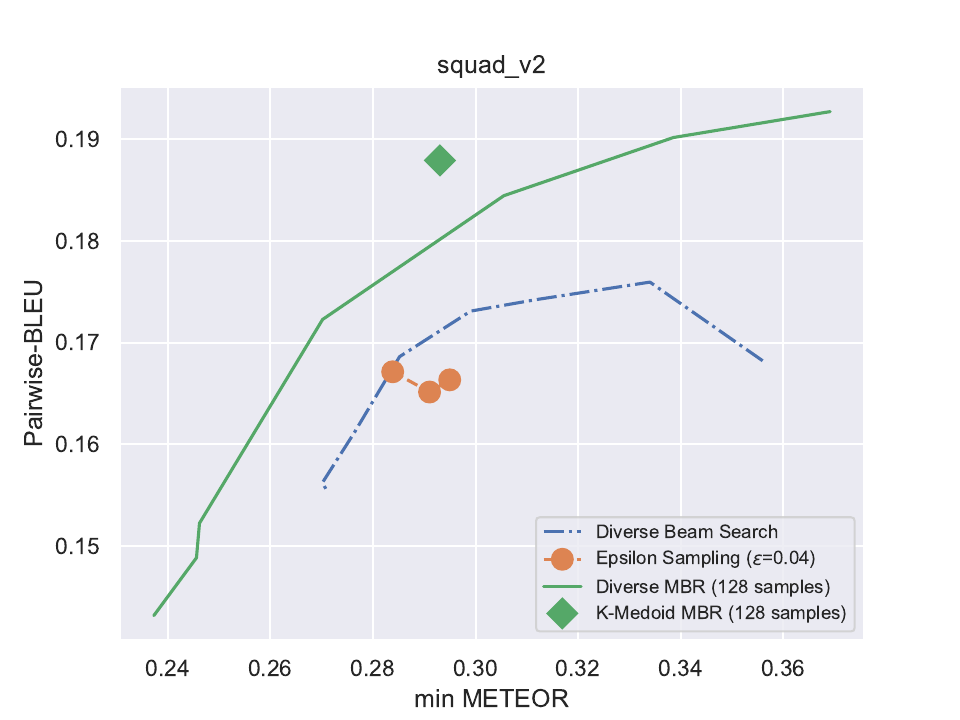}
        \caption{P-BLEU $\downarrow$ (SQuADv2)}
    \end{subfigure}
    \begin{subfigure}[b]{0.32\textwidth}
        \includegraphics[width=\textwidth]{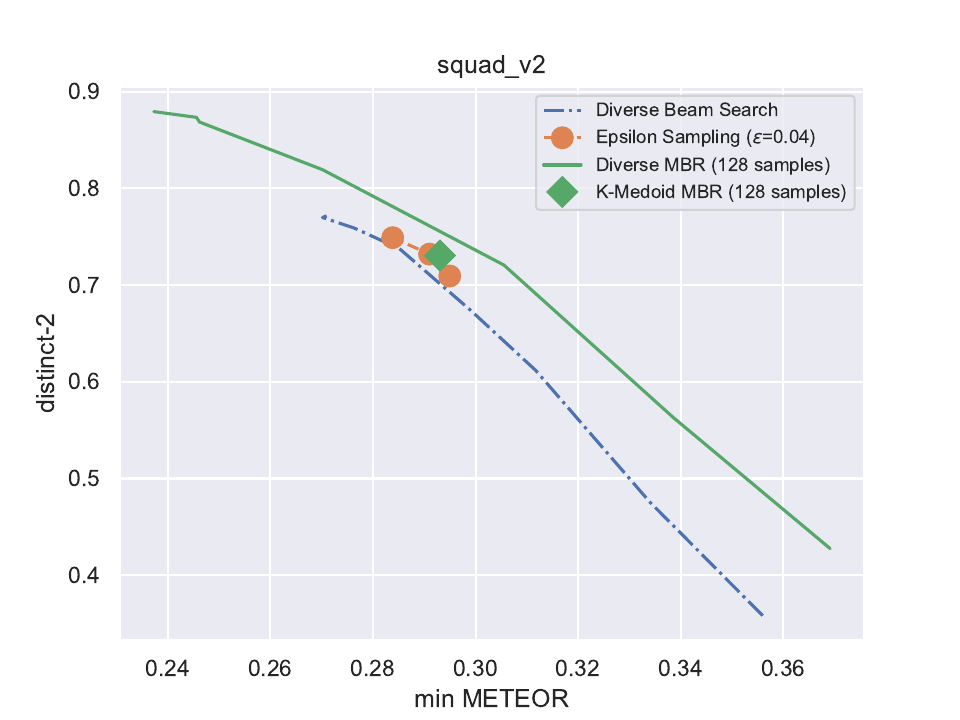}
        \caption{distinct-2 $\uparrow$ (SQuADv2)}
    \end{subfigure}
    \begin{subfigure}[b]{0.32\textwidth}
        \includegraphics[width=\textwidth]{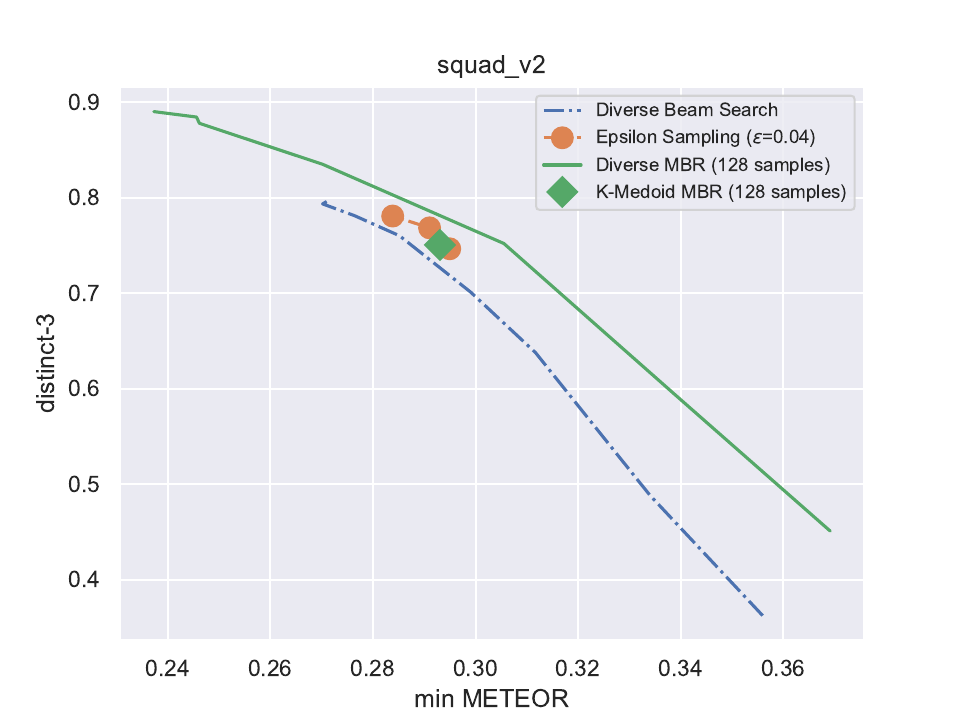}
        \caption{distinct-3 $\uparrow$ (SQuADv2)}
    \end{subfigure}

    \begin{subfigure}[b]{0.32\textwidth}
        \includegraphics[width=\textwidth]{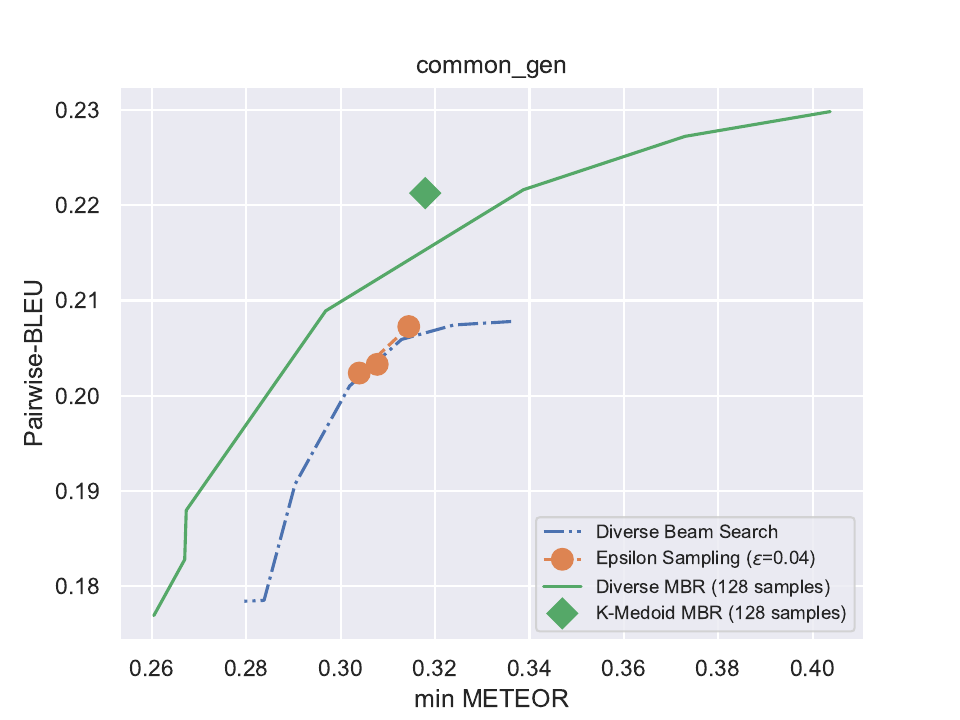}
        \caption{P-BLEU $\downarrow$ (CommonGen)}
    \end{subfigure}
    \begin{subfigure}[b]{0.32\textwidth}
        \includegraphics[width=\textwidth]{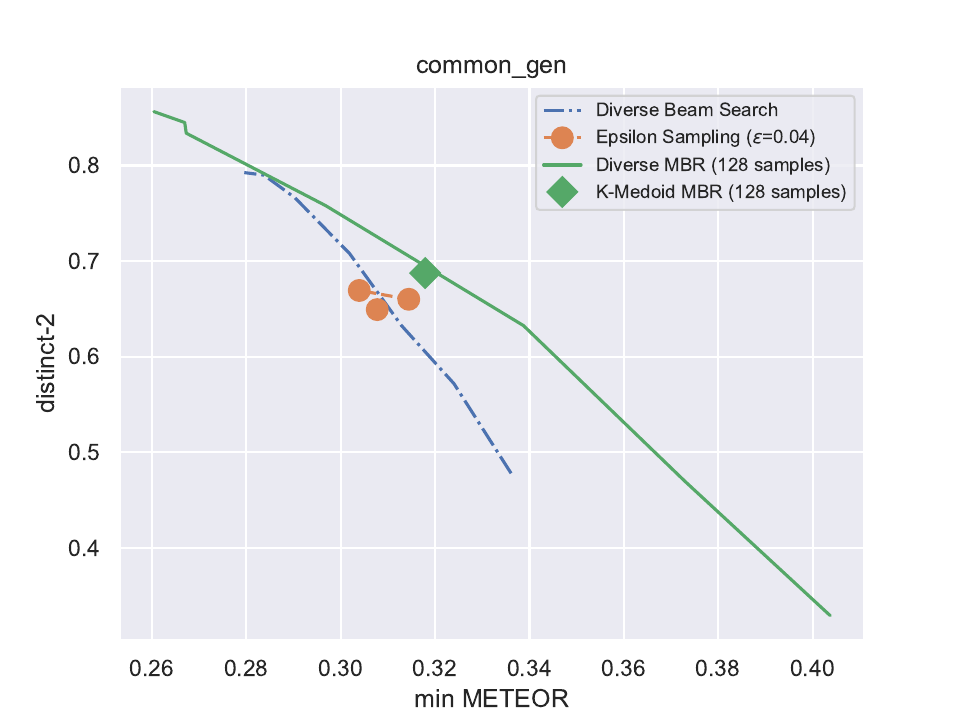}
        \caption{distinct-2 $\uparrow$ (CommonGen) }
    \end{subfigure}
    \begin{subfigure}[b]{0.32\textwidth}
        \includegraphics[width=\textwidth]{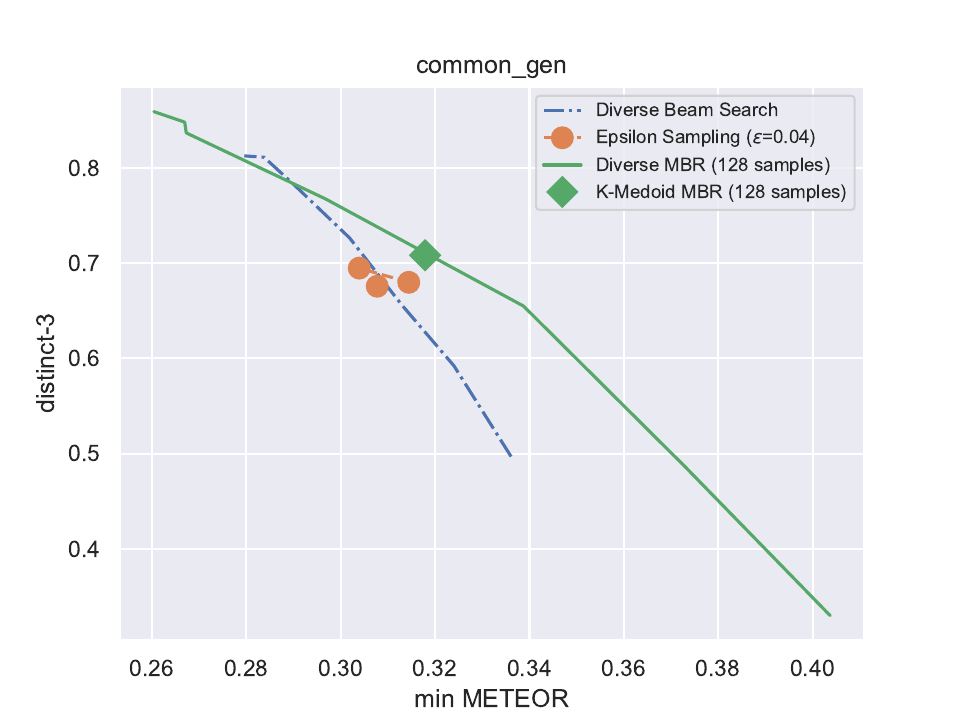}
        \caption{distinct-3 $\uparrow$ (CommonGen)}
    \end{subfigure}

    \begin{subfigure}[b]{0.32\textwidth}
        \includegraphics[width=\textwidth]{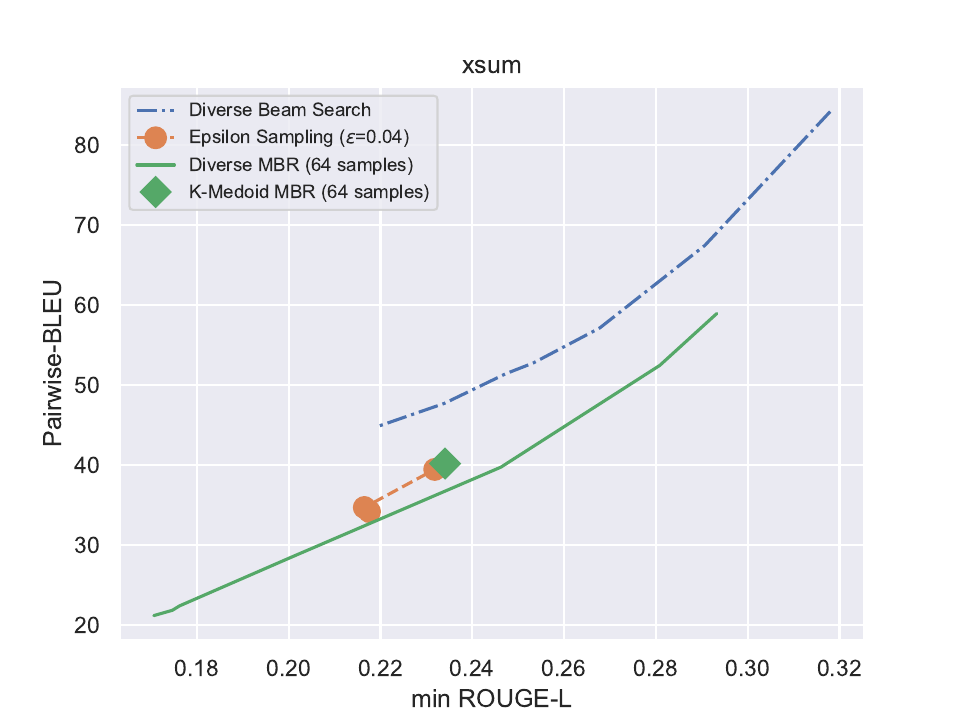}
        \caption{P-BLEU $\downarrow$ (XSum)}
    \end{subfigure}
    \begin{subfigure}[b]{0.32\textwidth}
        \includegraphics[width=\textwidth]{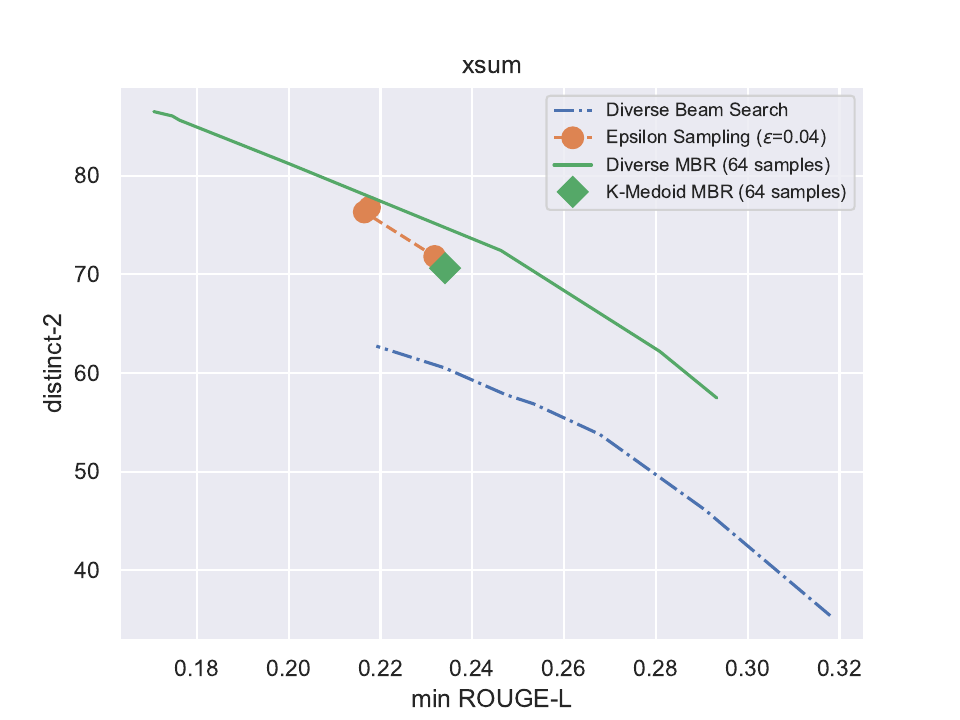}
        \caption{distinct-2 $\uparrow$ (XSum) }
    \end{subfigure}
    \begin{subfigure}[b]{0.32\textwidth}
        \includegraphics[width=\textwidth]{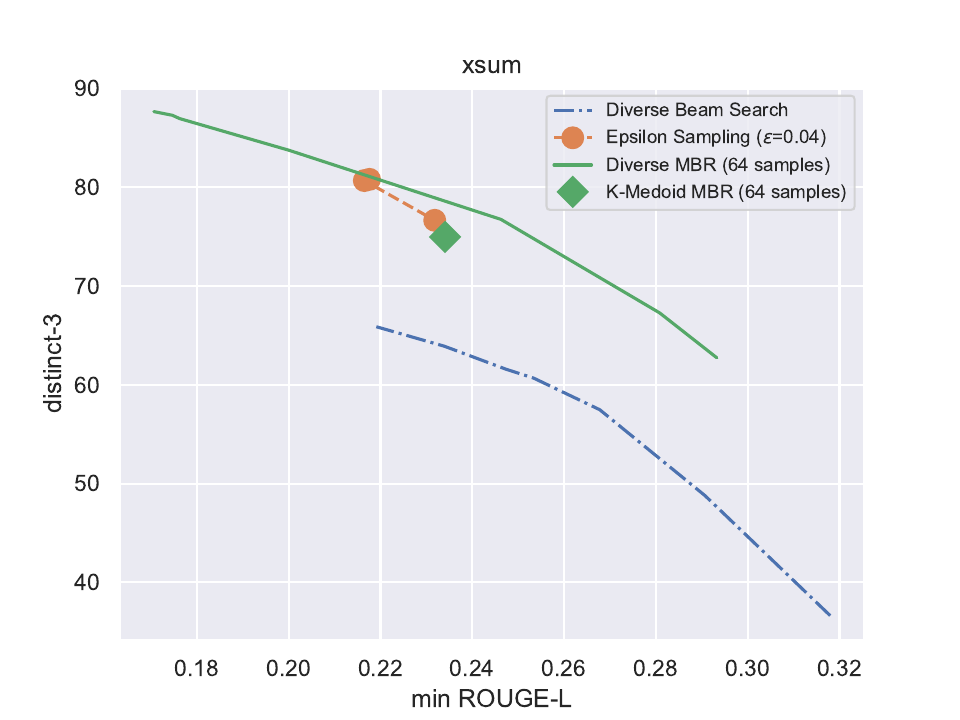}
        \caption{distinct-3 $\uparrow$ (XSum)}
    \end{subfigure}

    \caption{Evaluation of the P-BLEU and distinct-2, 3 as a function of min BLEU (MS COCO), METEOR (SQuADv2, CommonGen), and ROUGE-L (XSum). The size of the output $k$ is set to 4.}
    \label{fig:othersmin}
\end{figure*}

The Oracle (max) and the min quality scores on MS COCO, SQuADv2, CommonGen, and XSum with an output size of 4 are present in Figures~\ref{fig:othersmax} and \ref{fig:othersmin}. We observe similar trends as in machine translation tasks.

\begin{figure*}
    \centering
    \begin{subfigure}[b]{0.32\textwidth}
        \includegraphics[width=\textwidth]{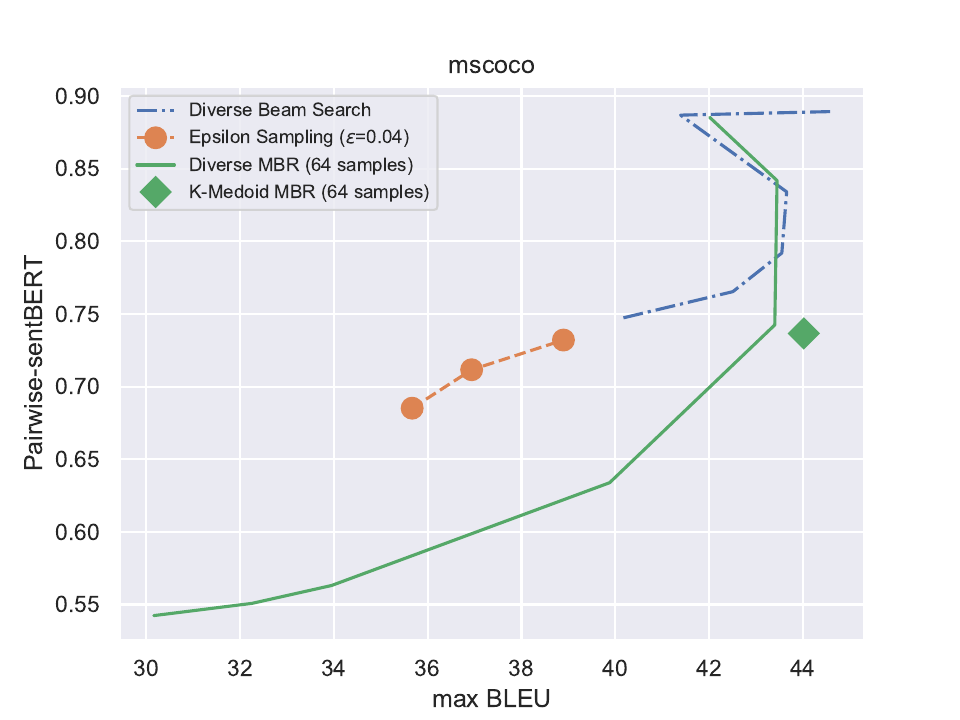}
        \caption{P-SentBERT $\downarrow$ (MS COCO)}
        \label{fig:sentbert-mscoco-max}
    \end{subfigure}
    \begin{subfigure}[b]{0.32\textwidth}
        \includegraphics[width=\textwidth]{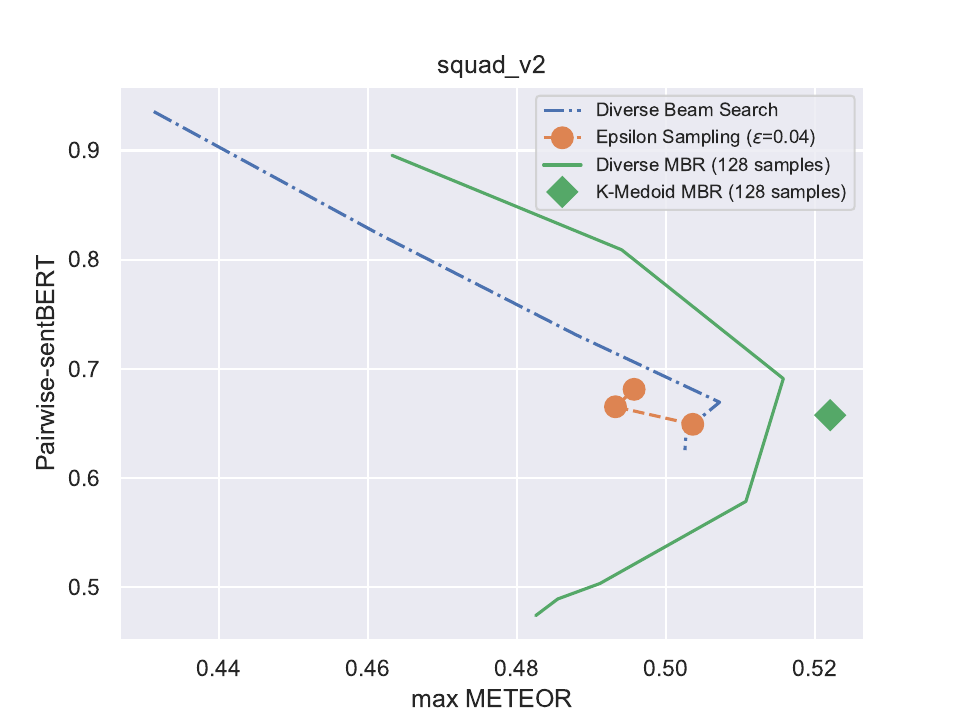}
        \caption{P-SentBERT $\downarrow$ (SQuADv2) }
    \end{subfigure}
    \begin{subfigure}[b]{0.32\textwidth}
        \includegraphics[width=\textwidth]{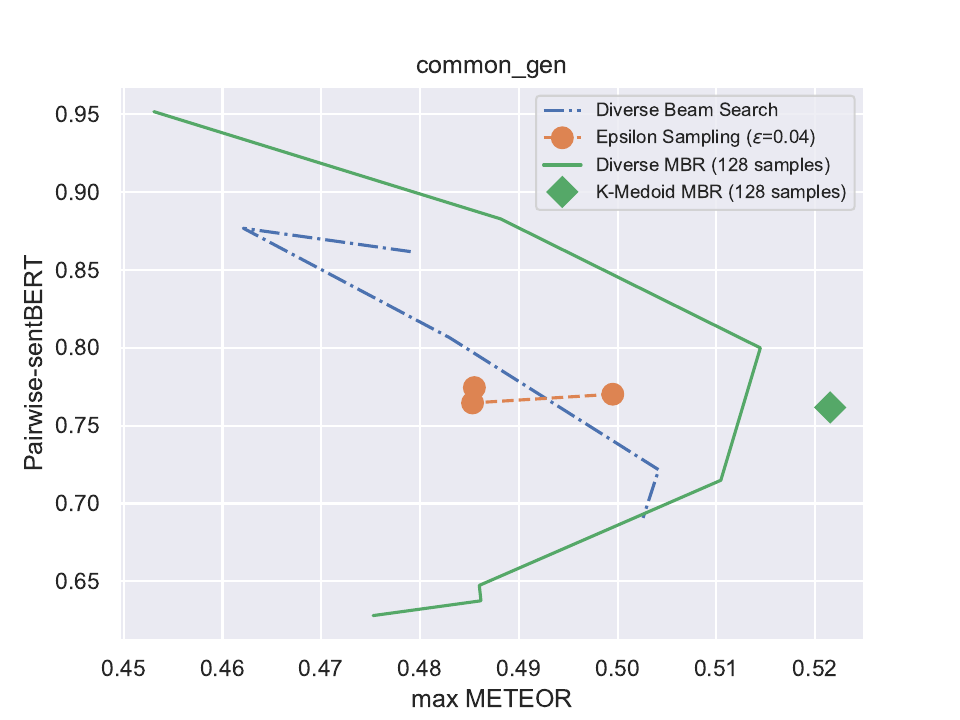}
        \caption{P-SentBERT $\downarrow$ (CommonGen)}
    \end{subfigure}
    \caption{Evaluation of semantic textual similarity using P-SentBERT as a function of the Oracle quality score (max BLEU and max METEOR). The number of outputs $k$ is $4$.}
    \label{fig:sentbert-max}
\end{figure*}

\begin{figure*}
    \centering
    \begin{subfigure}[b]{0.32\textwidth}
        \includegraphics[width=\textwidth]{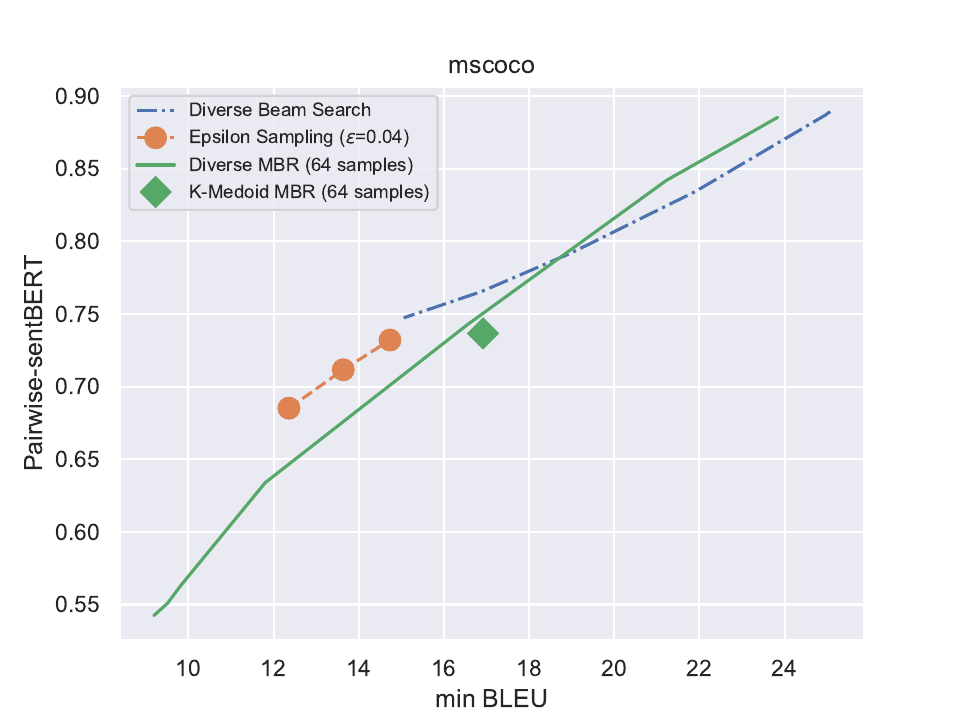}
        \caption{P-SentBERT $\downarrow$ (MS COCO)}
        \label{fig:sentbert-mscoco-min}
    \end{subfigure}
    \begin{subfigure}[b]{0.32\textwidth}
        \includegraphics[width=\textwidth]{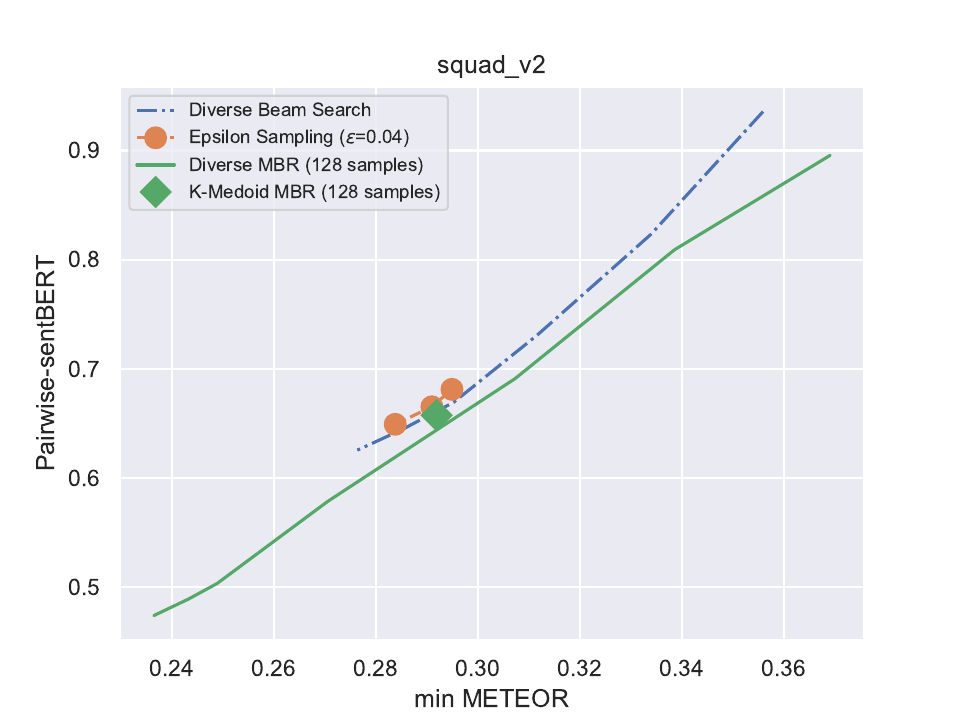}
        \caption{P-SentBERT $\downarrow$ (SQuADv2) }
    \end{subfigure}
    \begin{subfigure}[b]{0.32\textwidth}
        \includegraphics[width=\textwidth]{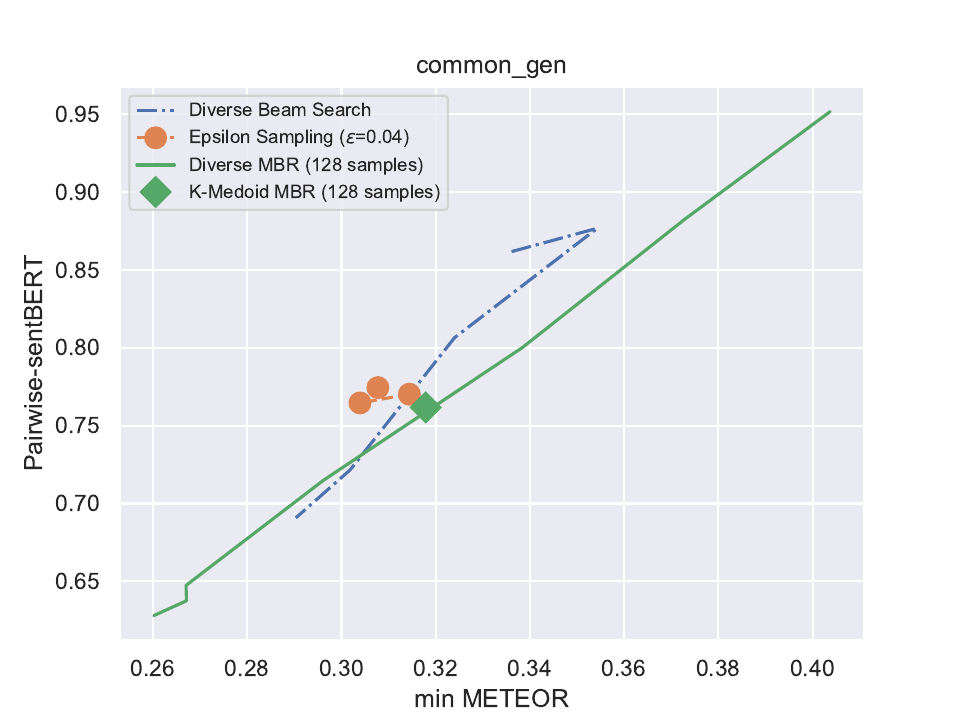}
        \caption{P-SentBERT $\downarrow$ (CommonGen)}
    \end{subfigure}
    \caption{Evaluation of semantic textual similarity using P-SentBERT as a function of the Oracle quality score (min BLEU and min METEOR). The number of outputs $k$ is $4$.}
    \label{fig:sentbert-min}
\end{figure*}

Figures~\ref{fig:sentbert-max} and \ref{fig:sentbert-min} show the P-SentBERT as a function of the max and min BLEU and METEOR scores on MS COCO, SQuADv2, and CommonGen. The result indicates that DMBR and DBS are successfully generating diverse outputs, not only lexically but also semantically.

\subsection{Summary of the Results}
\label{sec:sum-results}
Tables~\ref{tab:wmt}, \ref{tab:wmt-deen-k}, \ref{tab:wmt-ruen-k}, \ref{tab:mscoco}, \ref{tab:squad}, \ref{tab:commongen}, and \ref{tab:xsum} show the summary of the experimental results described in Section~\ref{sec:experiments}.

\begin{table*}
    \centering

    \adjustbox{max width=\textwidth}{
\begin{tabular}{lccccccc}
    \toprule
     & \multicolumn{3}{c}{Quality} & \multicolumn{4}{c}{Diversity} \\
    \cmidrule(lr){2-4}\cmidrule(lr){5-8}
    Decoder & min BLEU $\uparrow$ & mean BLEU $\uparrow$ & max BLEU $\uparrow$ & Pairwise-BLEU $\downarrow$ & distinct-1 $\uparrow$ & distinct-2 $\uparrow$ & distinct-3 $\uparrow$ \\
    \midrule\midrule
    & \multicolumn{6}{c}{WMT'19 De-En ($k=4$)} \\
    \midrule
    Epsilon & 17.09 & 27.11 & 38.06 & 34.13 & 0.40 & 0.55 & 0.60 \\
    Top-k & 15.42 & 25.81 & 37.51 & 28.87 & 0.42 & 0.59 & 0.65 \\
    Nucleus & 13.81 & 24.14 & 35.51 & 25.70 & 0.46 & 0.63 & 0.67 \\
    Ancestral & 9.20 & 18.82 & 30.19 & 15.72 & \textbf{0.55} & \textbf{0.73} & \textbf{0.76} \\
    \midrule
    Beam & 27.36 & \underline{34.98} & 43.93 & 68.64 & 0.29 & 0.35 & 0.36 \\
\bottomrule
    \end{tabular}
    }
    \caption{Evaluation of the quality and diversity using various decoding algorithms on WMT'19 De-En and Ru-En dataset. The size of the output $k$ is set to 4. Hyperparameters of the sampling algorithms are epsilon $\epsilon=0.02$, top-k $k_{\mathrm{top}}=10$, and nucleus $p=0.9$. The best score is bolded and the second best score is underlined.}
    \label{tab:wmt}
\end{table*}

\begin{table*}
    \centering

    \adjustbox{max width=\textwidth}{
    \begin{tabular}{lccccccc}
    \toprule
     & \multicolumn{3}{c}{Quality} & \multicolumn{4}{c}{Diversity} \\
    \cmidrule(lr){2-4}\cmidrule(lr){5-8}
    Decoder & min BLEU $\uparrow$ & mean BLEU $\uparrow$ & max BLEU $\uparrow$ & Pairwise-BLEU $\downarrow$ & distinct-1 $\uparrow$ & distinct-2 $\uparrow$ & distinct-3 $\uparrow$ \\
    \midrule\midrule
    & \multicolumn{6}{c}{WMT'19 De-En ($k=8$)} \\
    \midrule
    Epsilon & 13.94 & 27.32 & 43.54 & 33.94 & 0.25 & 0.40 & 0.48 \\
    Top-k & 12.19 & 25.46 & 42.23 & 28.39 & 0.28 & 0.46 & 0.54 \\
    Nucleus & 10.49 & 23.90 & 40.17 & 25.37 & 0.33 & 0.51 & 0.58 \\
    Ancestral & 5.34 & 17.28 & 34.05 & \textbf{13.23} & \textbf{0.47} & \textbf{0.67} & \textbf{0.72} \\
    \midrule
    Beam & 23.82 & 34.02 & 47.28 & 63.21 & 0.17 & 0.23 & 0.25 \\
\bottomrule
    \end{tabular}
    }
    \caption{Evaluation of the quality and diversity using various decoding algorithms on WMT'19 De-En and Ru-En dataset. The size of the output $k$ is set to 8. Hyperparameters of the sampling algorithms are $\epsilon=0.02$, $k_{\mathrm{top}}=10$, and $p=0.9$ for epsilon sampling, top-k sampling, and nucleus sampling, respectively. The best score is bolded and the second best score is underlined.}
    \label{tab:wmt-deen-k}
\end{table*}

\begin{table*}
    \centering

    \adjustbox{max width=\textwidth}{
    \begin{tabular}{lccccccc}
    \toprule
     & \multicolumn{3}{c}{Quality} & \multicolumn{4}{c}{Diversity} \\
    \cmidrule(lr){2-4}\cmidrule(lr){5-8}
    Decoder & min BLEU $\uparrow$ & mean BLEU $\uparrow$ & max BLEU $\uparrow$ & Pairwise-BLEU $\downarrow$ & distinct-1 $\uparrow$ & distinct-2 $\uparrow$ & distinct-3 $\uparrow$ \\
    \midrule\midrule
    & \multicolumn{6}{c}{WMT'19 De-En ($k=12$)} \\
    \midrule
    Epsilon  & 12.61 & 27.54 & 46.32 & 34.16 & 0.19 & 0.33 & 0.41 \\
    Top-k & 10.93 & 25.56 & 45.03 & 28.68 & 0.22 & 0.39 & 0.48 \\
    Nucleus  & 9.30 & 23.83 & 42.75 & 25.41 & 0.27 & 0.45 & 0.53 \\
    Ancestral  & 4.09 & 16.77 & 36.51 & \textbf{12.61} & \textbf{0.42} & \textbf{0.64} & \textbf{0.70} \\
    \midrule
    Beam & 21.69 & 33.32 & 49.22 & 60.45 & 0.13 & 0.18 & 0.21 \\
\bottomrule
    \end{tabular}
    }
    \caption{Evaluation of the quality and diversity using various decoding algorithms on WMT'19 De-En and Ru-En dataset. The size of the output $k$ is set to 12. Hyperparameters of the sampling algorithms are $\epsilon=0.02$, $k_{\mathrm{top}}=10$, and $p=0.9$ for epsilon sampling, top-k sampling, and nucleus sampling, respectively. The best score is bolded and the second best score is underlined.}
    \label{tab:wmt-ruen-k}
\end{table*}

\begin{table*}
    \centering
    \adjustbox{max width=\textwidth}{
    \begin{tabular}{lccccccc}
    \toprule
     & \multicolumn{3}{c}{Quality} & \multicolumn{4}{c}{Diversity} \\
    \cmidrule(lr){2-4}\cmidrule(lr){5-8}
    Decoder & min BLEU $\uparrow$ & mean BLEU $\uparrow$ & max BLEU $\uparrow$ & Pairwise-BLEU $\downarrow$ & distinct-1 $\uparrow$ & distinct-2 $\uparrow$ & distinct-3 $\uparrow$ \\
    \midrule\midrule
    & \multicolumn{6}{c}{MS COCO ($k=4$)} \\
    \midrule
    Epsilon & 13.64 & 23.83 & 36.94 & 16.13 & 0.48 & 0.68 & 0.71 \\
    \midrule
    Beam & \textbf{25.07} & \textbf{34.19} & \textbf{44.58} & 54.44 & 0.30 & 0.38 & 0.40 \\
    DBS-0.5 & 21.89 & 32.26 & 43.65 & 40.28 & 0.34 & 0.46 & 0.49 \\
    DBS-1.0 & 18.99 & 30.61 & 43.55 & 31.95 & 0.38 & 0.53 & 0.56 \\
    DBS-2.0 & 16.88 & 28.99 & 42.51 & 28.74 & 0.42 & 0.57 & 0.58 \\
\bottomrule
    \end{tabular}
    }
    \caption{Evaluation of the quality and diversity using various decoding algorithms on MS COCO dataset. The size of the output $k$ is set to 4. Epsilon sampling is set $\epsilon=0.02$. The best score is bolded and the second best score is underlined.}
    \label{tab:mscoco}
\end{table*}

\begin{table*}
    \centering
    \adjustbox{max width=\textwidth}{
    \begin{tabular}{lccccccc}
    \toprule
     & \multicolumn{3}{c}{Quality} & \multicolumn{4}{c}{Diversity} \\
    \cmidrule(lr){2-4}\cmidrule(lr){5-8}
    Decoder & min METEOR $\uparrow$ & mean METEOR $\uparrow$ & max METEOR $\uparrow$ & Pairwise-BLEU $\downarrow$ & distinct-1 $\uparrow$ & distinct-2 $\uparrow$ & distinct-3 $\uparrow$ \\
    \midrule\midrule
    & \multicolumn{6}{c}{SQuADv2 ($k=4$)} \\
    \midrule
    Epsilon & 28.38 & 38.99 & 50.36 & 16.72 & 0.55 & 0.75 & 0.78 \\
    \midrule
    Beam & 35.60 & 39.28 & 43.13 & 16.82 & 0.31 & 0.36 & 0.36 \\
    DBS-0.5 & 31.17 & 39.66 & 48.80 & 17.42 & 0.47 & 0.61 & 0.64 \\
    DBS-1.0 & 29.53 & 39.78 & 50.72 & 17.27 & 0.53 & 0.70 & 0.73 \\
    DBS-2.0 & 28.52 & 39.16 & 50.27 & 16.86 & 0.55 & 0.74 & 0.76 \\
\bottomrule
    \end{tabular}
    }
    \caption{Evaluation of the quality and diversity using various decoding algorithms on SQuAD v2 dataset. The size of the output $k$ is set to 4, 8, and 12. Epsilon sampling is set $\epsilon=0.01$. The best score is bolded and the second best score is underlined.}
    \label{tab:squad}
\end{table*}

\begin{table*}
    \centering
    \adjustbox{max width=\textwidth}{
    \begin{tabular}{lccccccc}
    \toprule
     & \multicolumn{3}{c}{Quality} & \multicolumn{4}{c}{Diversity} \\
    \cmidrule(lr){2-4}\cmidrule(lr){5-8}
    Decoder & min METEOR $\uparrow$ & mean METEOR $\uparrow$ & max METEOR $\uparrow$ & Pairwise-BLEU $\downarrow$ & distinct-1 $\uparrow$ & distinct-2 $\uparrow$ & distinct-3 $\uparrow$ \\
    \midrule\midrule
    & \multicolumn{6}{c}{CommonGen ($k=4$)} \\
    \midrule
    Epsilon & 0.31 & 0.41 & 0.50 & 0.21 & 0.49 & 0.66 & 0.68 \\
    \midrule
    Beam & 0.34 & 0.41 & 0.48 & 0.21 & 0.37 & 0.48 & 0.50 \\
    DBS-0.5 & 0.32 & 0.40 & 0.48 & 0.21 & 0.43 & 0.57 & 0.59 \\
    DBS-1.0 & 0.31 & 0.40 & 0.50 & 0.20 & 0.48 & 0.66 & 0.68 \\
    DBS-2.0 & 0.30 & 0.40 & 0.50 & 0.20 & 0.51 & 0.71 & 0.73 \\
\bottomrule
    \end{tabular}
    }
    \caption{Evaluation of the quality and diversity using various decoding algorithms on CommonGen dataset. The size of the output $k$ is set to 4, 8, and 12. Epsilon sampling is set $\epsilon=0.01$. The best score is bolded and the second best score is underlined.}
    \label{tab:commongen}
\end{table*}

\begin{table*}
    \centering
    \adjustbox{max width=\textwidth}{
    \begin{tabular}{lccccccc}
    \toprule
     & \multicolumn{3}{c}{Quality} & \multicolumn{4}{c}{Diversity} \\
    \cmidrule(lr){2-4}\cmidrule(lr){5-8}
    Decoder & min ROUGE-L $\uparrow$ & mean ROUGE-L $\uparrow$ & max ROUGE-L $\uparrow$ & Pairwise-BLEU $\downarrow$ & distinct-1 $\uparrow$ & distinct-2 $\uparrow$ & distinct-3 $\uparrow$ \\
    \midrule\midrule
    & \multicolumn{6}{c}{XSum ($k=4$)} \\
    \midrule
    Epsilon  & 21.65 & 30.15 & 39.23 & 34.73 & 0.55 & 0.76 & 0.81 \\
    \midrule
    Beam & \textbf{31.80} & \textbf{35.51} & 39.42 & 84.21 & 0.30 & 0.35 & 0.37 \\
    DBS-0.5 & 26.78 & 34.39 & 42.37 & 57.17 & 0.41 & 0.54 & 0.57 \\
    DBS-1.0 & 24.73 & 33.56 & 42.90 & 51.45 & 0.43 & 0.58 & 0.62 \\
    DBS-2.0 & 23.39 & 32.62 & 42.44 & 47.74 & 0.45 & 0.61 & 0.64 \\
    DBS-5.0 & 21.91 & 31.37 & 41.54 & 44.80 & 0.47 & 0.63 & 0.66 \\
    \midrule
    MBR & 29.33 & 35.26 & 41.67 & 58.98 & 0.43 & 0.57 & 0.63 \\
    DMBR-0.1 & 28.08 & 35.20 & 42.93 & 52.48 & 0.47 & 0.62 & 0.67 \\
    DMBR-0.3 & 24.62 & 33.53 & 43.05 & 39.77 & 0.54 & 0.72 & 0.77 \\
    DMBR-0.5 & 20.00 & 30.18 & 41.34 & 28.37 & 0.62 & 0.81 & 0.84 \\
    DMBR-1.0 & 17.46 & 26.34 & 36.65 & 21.87 & 0.66 & 0.86 & 0.87 \\
    DMBR-2.0 & 17.06 & 25.61 & 35.58 & \textbf{21.21} & \textbf{0.66} & \textbf{0.86} & \textbf{0.88} \\
    KMBR & 23.41 & 33.03 & \textbf{43.46} & 40.22 & 0.53 & 0.71 & 0.75 \\
    \bottomrule
    \end{tabular}
    }
    \caption{Evaluation of the quality and diversity using various decoding algorithms on XSum dataset. The size of the output $k$ is set to 4. Epsilon sampling is set $\epsilon=0.02$. The best score is bolded and the second best score is underlined.}
    \label{tab:xsum}
\end{table*}

    



\section{Pretrained Models and Codes used in the Experiments}

We list the pretrained models and codes we used in the experiments in Table~\ref{tab:models}.

\begin{table*}
    \centering
    \adjustbox{max width=\textwidth}{
    \begin{tabular}{cl}
    \toprule
        \multicolumn{2}{c}{Text Generation Models} \\\midrule
        WMT'19 (Section \ref{sec:mt}) & \citet{ng-etal-2019-facebook} \url{https://github.com/facebookresearch/fairseq/blob/main/examples/wmt19/README.md} \\
        MS COCO (Section \ref{sec:captioning}) & \citet{pmlr-v202-li23q} \url{https://huggingface.co/Salesforce/blip2-flan-t5-xl-coco} \\        
        SQuADv2 (Section \ref{sec:squad}) & \citet{tunstall2023zephyr} \url{https://huggingface.co/HuggingFaceH4/zephyr-7b-beta} \\
        CommonGen (Section \ref{sec:commongen}) & \citet{tunstall2023zephyr} \url{https://huggingface.co/HuggingFaceH4/zephyr-7b-beta} \\
        XSum (Section \ref{sec:sum}) & \citet{lewis-etal-2020-bart} \url{https://huggingface.co/facebook/bart-large-xsum} \\\midrule\midrule
        \multicolumn{2}{c}{Models for Evaluation} \\ \midrule
        WMT'19 (Section \ref{sec:mt}) & sacreBLEU: \citet{post-2018-call} \url{https://github.com/mjpost/sacrebleu} \\
        MS COCO (Section \ref{sec:captioning}) & Sentence BERT: \citet{song2020mpnet} \url{https://huggingface.co/sentence-transformers/all-mpnet-base-v2} \\
        SQuADv2 (Section \ref{sec:squad}) & Sentence BERT: \citet{song2020mpnet} \url{https://huggingface.co/sentence-transformers/all-mpnet-base-v2} \\
        CommonGen (Section \ref{sec:commongen}) & Sentence BERT: \citet{song2020mpnet} \url{https://huggingface.co/sentence-transformers/all-mpnet-base-v2} \\
        CommonGen (Section \ref{sec:commongen}) & Porter stemmer: \citet{porter1980algorithm} \url{https://www.nltk.org/_modules/nltk/stem/porter.html} \\
    \bottomrule
    \end{tabular}
    }
    \caption{List of pretrained models we used in the experiments.}
    \label{tab:models}
\end{table*}

\end{document}